\pdfoutput=1
\documentclass{article}
\usepackage{iclr2027_conference}
\iclrfinalcopy
\usepackage{times}
\usepackage{latexsym}
\usepackage[T1]{fontenc}
\usepackage[utf8]{inputenc}
\usepackage{inconsolata}

\usepackage{microtype}
\usepackage{hyperref}
\usepackage{url}
\usepackage{booktabs}
\usepackage{graphicx}
\usepackage{amsmath}
\usepackage{amssymb}
\usepackage{multirow}
\usepackage[table]{xcolor}
\usepackage{caption}
\usepackage{float}
\usepackage{subcaption}
\renewcommand{\topfraction}{0.9}
\renewcommand{\bottomfraction}{0.9}
\renewcommand{\textfraction}{0.1}
\renewcommand{\floatpagefraction}{0.8}
\usepackage{listings}
\usepackage[most]{tcolorbox}
\definecolor{tablerow}{gray}{0.93}
\definecolor{codebg}{HTML}{F6F8FA}
\definecolor{codeborder}{HTML}{D0D7DE}
\definecolor{codetext}{HTML}{24292F}
\definecolor{jsonkey}{HTML}{0550AE}
\definecolor{jsonstr}{HTML}{0A3069}
\definecolor{steered}{HTML}{1A7F37}
\definecolor{original}{HTML}{6E7781}

\lstdefinestyle{jsonoutput}{
  basicstyle=\ttfamily\footnotesize\color{codetext},
  backgroundcolor=\color{codebg},
  frame=single,
  rulecolor=\color{codeborder},
  framesep=6pt,
  xleftmargin=8pt,
  xrightmargin=8pt,
  breaklines=true,
  columns=fullflexible,
  keepspaces=true,
  aboveskip=4pt,
  belowskip=4pt,
}

\newtcolorbox{genbox}[1][]{
  colback=codebg,
  colframe=codeborder,
  boxrule=0.5pt,
  arc=3pt,
  left=3pt, right=3pt, top=2pt, bottom=2pt,
  fontupper=\ttfamily\scriptsize\color{codetext},
  before upper={\hfuzz=200pt},
  #1
}

\definecolor{darkblue}{rgb}{0, 0, 0.5}
\hypersetup{colorlinks=true, citecolor=darkblue, linkcolor=darkblue, urlcolor=darkblue}

\title{Tool Calling is Linearly Readable and Steerable in Language Models}

\author{
  \textbf{Zekun Wu\textsuperscript{1,2}},
  \textbf{Ze Wang\textsuperscript{1,2}},
  \textbf{Seonglae Cho\textsuperscript{2}},
  \textbf{Yufei Yang\textsuperscript{3}},
\\
  \textbf{Adriano Koshiyama\textsuperscript{1,2}},
  \textbf{Sahan Bulathwela\textsuperscript{1}},
  \textbf{Maria Perez-Ortiz\textsuperscript{1}}
\\
\\
  \textsuperscript{1}University College London,
  \textsuperscript{2}Holistic AI,
  \textsuperscript{3}Imperial College London
}

\newcommand{\direction}{\mathbf{d}}
\newcommand{\resid}{\mathbf{h}}

\begin{document}
\raggedbottom

\maketitle
\lhead{Preprint}

\begin{abstract}
Language-model agents can take real actions by calling tools, so choosing the wrong tool can cause errors that are difficult to undo. Most evaluations only observe the tool choice after the model generates a call. We read this choice from the model before generation, and we steer it. For each tool, we average the model's hidden states from a few example requests to obtain a tool vector. Comparing a new request with these tool vectors predicts which tool it needs. The difference between two tool vectors gives a steering direction that can move the model toward a chosen tool without retraining. Across eight instruction-tuned models from 4B to 27B parameters, steering changes the generated call to a chosen target tool in 56--78\% of held-out tool pairs, depending on the model. We also observe steering on real APIs from $\tau$-bench and ToolBench, though less reliably. At the final layer, steering could work simply by raising the score of the target tool name. To test this, we compare the steering direction with a direction that only raises that score, layer by layer. In the middle layers, on the three models we examine in depth, the steering direction switches more calls than the score-raising direction. So the tool vectors capture part of the tool choice before the final layer. The same tool vectors also help identify likely tool-selection errors before generation. Calls are more likely to be wrong when a request is similarly close to two tool vectors. Across four models, this signal achieves an AUROC of 0.61--0.78 and outperforms a first-token confidence baseline on three of them. Together, these results suggest that the model's hidden state provides a way to read, steer, and check tool choice before a call is made.
\end{abstract}

\section{Introduction}
\label{sec:intro}
Language models now send messages, schedule meetings and issue refunds by calling tools. A wrong call can cause harm that a retry cannot undo \citep{hendrycks2023overview}. We study how a model picks among its tools, and we show that the choice can be seen, and changed, before the call is made. Most work on tool calling looks at the model from outside. Studies train models to call tools better \citep{schick2023toolformer, patil2023gorilla, qin2024toolllm}, measure which tools models pick and with what arguments \citep{yao2024taubench, li2023apibank, berkeley-function-calling-leaderboard, huang2024metatool}, or change how tools are presented \citep{hao2023toolkengpt}. Two studies look inside. \citet{healy2026internal} find a signal for tool-selection hallucinations in the hidden state, and leave the signal's effect on the call untested. \citet{wang2026asa} steer the decision to call a tool at all.

\begin{figure}[ht]
\centering
\includegraphics[width=1.0\linewidth]{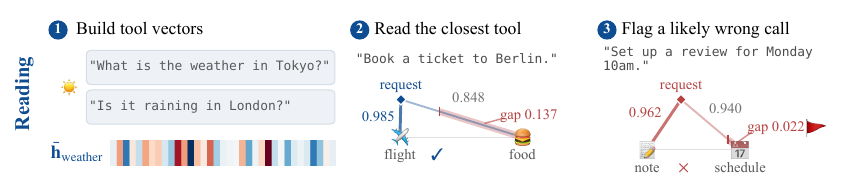}\par\vspace{-7pt}
\includegraphics[width=1.0\linewidth]{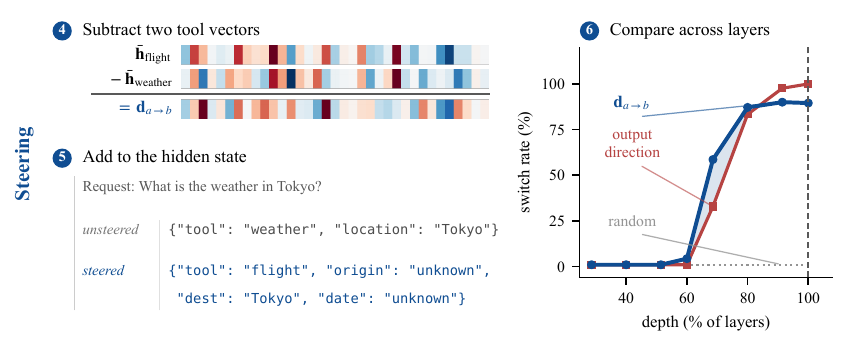}
\caption{Tool vectors and their two uses. Steps 1 to 3 read the tool on Gemma~3 4B. Averaging the hidden states of two example requests gives one tool vector per tool (1). A held-out request is read as the tool whose vector is closest (2). This example has the largest gap between the two highest similarity scores among 30 held-out requests. The request with the smallest gap is the only wrong readout, \texttt{schedule} read as \texttt{note} (3), and Section~\ref{sec:errorrisk} uses the gap to flag likely wrong calls. Steps 4 to 6 steer. The difference of two tool vectors (4), added to a new request's hidden state, switches the call on Gemma~3 4B (5). On Qwen~3 4B, the steering direction switches more held-out pairs than the output direction at middle layers (shaded) but not at the last layer (dashed, Section~\ref{sec:depth}) (6).}
\label{fig:hero}
\end{figure}

We study which tool the model calls. We read the choice among many tools from inside the model and switch it to one we pick. For each tool, we average the model's hidden state at the last prompt token over a few example requests, which gives one tool vector per tool. Comparing a new request with these tool vectors predicts which tool the request needs. Subtracting the tool vector of tool $a$ from the tool vector of tool $b$ gives a steering direction. Adding the steering direction to the hidden state pushes the model from $a$ toward $b$, with no training (Figure~\ref{fig:hero}). The main steering comparison covers eight instruction-tuned models from 4B to 27B in the Gemma~3 \citep{gemma2024}, Qwen~3/2.5 \citep{qwen3_2025,qwen25_2024} and Llama~3.1 \citep{llama31_2024} families. Four smaller models and the base versions let us compare model size and instruction tuning.

Across the eight models, steering switches the generated call to the target tool in 56--78\% of held-out tool pairs (Section~\ref{sec:steering}). On real APIs from $\tau$-bench and ToolBench the rates are lower. ToolBench groups its APIs by domain, such as three finance APIs whose names all start with \texttt{finance\_}. There, steering reaches the right domain in most pairs but the right API in only about a third. Steering at the last layer could work simply by raising the score of the target tool name. We therefore compare the steering direction with the output direction, a direction built from the tool names' output weights, at every layer. At the last layer, the output direction does at least as well. In the middle layers, on Gemma~3 4B, Qwen~3 4B and Llama~3.1 8B, the steering direction switches more pairs. So part of the tool choice is made before the output layer (Section~\ref{sec:depth}). To locate the choice inside the network, we run two requests that need different tools and copy one attention head's or feed-forward layer's output from one run into the other. If the copy changes which tool the model prefers, that head or layer carries part of the choice. This test, called activation patching, finds a few such heads and layers on Gemma~3 4B and Qwen~3 4B (Section~\ref{sec:circuit}). Finally, the same tool vectors warn before a call is written. A request that is about equally close to two tool vectors is more likely to produce a wrong call. On the BFCL benchmark, this signal reaches an AUROC of 0.61--0.78 on four models and beats a first-token confidence baseline on three (Section~\ref{sec:errorrisk}).

This paper makes four contributions. First, tool choice can be edited before generation. Adding the difference of two averaged hidden states switches the generated call in 56--78\% of held-out tool pairs across eight models, while the unsteered model names the target in almost none (Section~\ref{sec:steering}). Second, the edit does more than raise the target name's score. In the middle layers, the steering direction switches more pairs than a direction built from the output weights, and removing its output-aligned part leaves most of the effect (Section~\ref{sec:depth}). Third, the control has a resolution limit. Near-synonymous tools cannot be separated, and on ToolBench the direction moves the call to the right domain but rarely to the right API (Section~\ref{sec:steering}). Fourth, the same tool vectors give a signal for likely wrong calls before the call is written (Section~\ref{sec:errorrisk}).

\section{Tool vectors}
\label{sec:method}

The method needs a few example requests per tool, one forward pass for each, and no training or gradients. For each tool $t_i$, we take $n = 2\text{--}3$ requests that should call it. For each request, we record the model's hidden state $\resid_i^{(j)} \in \mathbb{R}^{d}$ at the last prompt token. The hidden state is the residual stream, the state that each attention and feed-forward block adds to. The main steering experiment uses the last layer, before the final normalization and the output projection that turns the hidden state into scores for the next token. The depth experiments rebuild the tool vectors at each tested layer. Averaging the states gives $\bar{\resid}_i = \frac{1}{n} \sum_j \resid_i^{(j)}$, which we call the tool vector of $t_i$ (Figure~\ref{fig:hero}, step 1).

To read which tool a new request needs, we compare the request's hidden state with every tool vector by cosine similarity and pick the closest (Figure~\ref{fig:hero}, step 2). To steer from tool $t_a$ to tool $t_b$, we use the steering direction $\direction_{a \to b} = \bar{\resid}_b - \bar{\resid}_a$, the difference of the two tool vectors with no normalization (Figure~\ref{fig:hero}, step 4). Before the model writes anything, we add a scaled copy of the steering direction to the hidden state at the same layer and token (Figure~\ref{fig:hero}, step 5):
\begin{equation}
    \resid_\ell \;\leftarrow\; \resid_\ell + \alpha \, \direction_{a \to b}
\end{equation}
The factor $\alpha$ multiplies the raw difference, so the size of the change differs from pair to pair. We use $\alpha{=}0.7$ for the synthetic tool lists, the ToolBench tool list and the depth comparison. We use $\alpha{=}1.0$ for the five-tool experiment in Section~\ref{sec:setup} and for the appendix controls built from output weights (Table~\ref{tab:strength} varies $\alpha$). Steering is applied once, at the last prompt token. The model's weights stay fixed, and the rest of the answer is generated normally. The method extends activation addition \citep{turner2024steering, arditi2024refusal} to selection among multiple tools.

\textbf{What last-layer steering can reach.} Steering at the last layer changes only the scores for the model's first output token. Later tokens still see the unedited prompt. So with the same prompt and greedy decoding, two steering directions that give the same first token give the same continuation. The limit matters for four synthetic tools whose names start with \texttt{get}: \texttt{get\_weather}, \texttt{get\_news}, \texttt{get\_stock\_price} and \texttt{get\_directions}. Steering a weather request toward \texttt{get\_news} targets the first token \texttt{get}, which the model writes anyway. Across the eight main models, whenever the answer to a \texttt{get\_} request starts with \texttt{get}, the model finishes it as the source tool's name, here \texttt{get\_weather}. So a \texttt{get\_} request cannot be switched to another \texttt{get\_} tool, and a request for any other tool can reach at most one of the four. Of the 210 ordered pairs, at most 165 can then be switched in these runs (Appendix~\ref{app:output_row}).

\section{Reading and steering tool choice}

\subsection{Experimental setup}
\label{sec:setup}

\textbf{Models.} We test 12 instruction-tuned models from the Gemma, Qwen and Llama families, and the base versions where available (Appendix~\ref{app:models}). The main steering results use the eight instruction-tuned models of 4B and above. The four smaller models show how results change with size. Most mechanism experiments use one model per family, which we call the three representative models: Gemma~3 4B, Qwen~3 4B and Llama~3.1 8B.

\textbf{Benchmarks.} The main experiments use single-turn requests with a fixed tool list. Appendix~\ref{sec:multiturn} reports multi-turn runs. We use synthetic tool lists that we control (Appendix~\ref{app:tools}), and real APIs from $\tau$-bench \citep{yao2024taubench}, ToolBench \citep{qin2024toolllm} and the BFCL v3 function-calling benchmark \citep{berkeley-function-calling-leaderboard}. On the synthetic five-tool list, each tool has three example requests. We build each tool vector from two and test on the third. Rotating which request is held out gives 60 trials per model over the 20 ordered pairs. In the synthetic experiments, the prompt lists the tools and the request, then starts the answer with \texttt{\{\{"tool": "} (the template's literal doubled brace), so the model's next words are the tool name (Appendix~\ref{app:eval_protocol}).

\textbf{Scoring.} The readout is scored against the \emph{label}, the tool the dataset says the request needs. In steering, the \emph{source} is the request's label and the \emph{target} is the tool we steer toward. A steering trial counts as a success when the model writes the target. The unsteered column of Table~\ref{tab:generalization} shows how often the model's own call already names the target. A generated tool name can be scored by its full name or only by its first token (the prefix rule). Four synthetic tools start with the same token, \texttt{get}, so the prefix rule cannot tell the four apart. Every cell of Table~\ref{tab:generalization} therefore matches the full name, with Wilson 95\% intervals for the synthetic scores. Appendix~\ref{app:realapi_scoring} scores the same generations by the first token for comparison. Picking a tool uniformly at random matches the full name $1/K$ of the time for $K$ tools. The prefix rule raises the chance rate to 12\% on the synthetic 15-tool list and 20\% on the ToolBench tool list of 15 APIs in five domains.

\subsection{Reading tool labels from hidden states}
\label{sec:reading}

We read a request's tool by picking the tool vector closest to the request's hidden state (Appendix~\ref{app:cosine_select}). The readout is scored against the dataset's label. The readout and the model's own generation receive the same prompt, so the two can be compared. We also tried subtracting the average tool vector before the cosine, and fitting a linear classifier. On BFCL, each beats the cosine on only one of five models, by 4 points or less (Appendix~\ref{app:cosine_select}).
The hidden state could carry the label because the model is preparing that tool call, or only because of what the request is about. To separate the two, we compare base and instruction-tuned models on BFCL, leaving each test request out of its tool vector (Table~\ref{tab:bfcl_reading}). Instruction-tuned models write the correct tool more often than the readout finds it. For the Gemma and Llama base models, the readout finds the label even though they generate the tool poorly. Base models might fail only on the answer format, so we also give them the start of the JSON answer. That does not rescue Gemma, while Qwen~3 4B base writes calls well with or without it (Appendix~\ref{app:base_it}). A sentence encoder that sees only the request reads the label about as well as the hidden state. On MetaTool, where the ten candidates are similar tools, the readout falls behind the sentence encoder on two of the three models (43--77\% against 75\%, Appendix~\ref{app:cosine_select}). At middle layers, tool vectors taken from a base model also read its instruction-tuned version, and the reverse, with most of their accuracy (Appendix~\ref{app:bfcl_reading}). So the hidden state carries the tool label, though the label may just reflect what the request is about. Section~\ref{sec:discussion} tests the topic explanation by comparing tools within a single topic.

\subsection{Steering tool selection}
\label{sec:steering}
\label{sec:emergence}
\label{sec:schema}
Reading shows that the tool label is in the hidden state. Steering shows that changing that label changes the call. We change the hidden state and count how often the call moves to a chosen target tool. On the five-tool list, instruction-tuned models of 4B or larger switch nearly all 60 trials at $\alpha{=}1.0$. At the weaker $\alpha{=}0.7$ they switch fewer (Figure~\ref{fig:matrix}, with counts in Appendix~\ref{app:matrix}).
The synthetic 15-tool list tests selection among 15 tools from eight domains (Table~\ref{tab:generalization}). Across the eight instruction-tuned models of 4B or larger, full-name switching succeeds on 56--78\% of held-out pairs. The 15-tool results use one fixed held-out request per tool. Rotating the held-out request moves the rate by up to 19 percentage points (Table~\ref{tab:fold_rotation}). The paired unsteered column runs the same 210 pairs and requests without intervention. The eight models already name the target in 0 to 1\% of pairs and name the tool the request asks for in 80 to 100\% (Appendix~\ref{app:unsteered_control}). Random vectors of matched length switch at most three of the 210 pairs at every tested layer (Figure~\ref{fig:depth_3fam}). With up to 150 tools, prefix-rule scores stay near 100\%, but full-name scores fall from 63--75\% at 15 tools to 20--25\% at 150 on the three representative models (Table~\ref{tab:scaling_v2}). At least 90\% of the full-name misses begin with the target's first word and end in another name, as the first-token edit predicts (Section~\ref{sec:method}).

\begin{table}[ht]
\centering
\caption{Steering accuracy (\%) is the share of ordered tool pairs whose steered call names the target tool, and every cell is scored on the full tool name. The synthetic columns use all 210 ordered pairs of 15 tools, each tested on a request that is held out of the tool vectors. \textsc{Unsteered} is the paired baseline, the same 210 pairs on the same held-out requests with no intervention, so it counts calls that already name the target (Appendix~\ref{app:unsteered_control}). The real-API columns use 30 ordered pairs each from $\tau$-bench airline, $\tau$-bench retail and ToolBench, and their test requests also build the tool vectors (Appendix~\ref{app:realapi_scoring} repeats three models with held-out requests). \textsc{Ambiguous} uses requests that fit several tools, with 15--19 trials per model, counting only trials where the unsteered generation does not already name the target (Appendix~\ref{app:generations}). Brackets give Wilson 95\% intervals. Each request and each target tool appears in 14 of the 210 pairs, so resampling target tools widens the interval to about $\pm$20 points (Table~\ref{tab:cluster_ci}).}
\label{tab:generalization}

\footnotesize
\setlength{\tabcolsep}{6pt}
\resizebox{\ifdim\width>\columnwidth\columnwidth\else\width\fi}{!}{\begin{tabular}{l cc ccc c}
\toprule
& \multicolumn{2}{c}{\textsc{Synthetic, 15 tools}} & \multicolumn{3}{c}{\textsc{Real API}} & \textsc{Ambiguous} \\
\cmidrule(lr){2-3}\cmidrule(lr){4-6}
& \textsc{Steered} [95\% CI] & \textsc{Unsteered} & \textsc{Airline} & \textsc{Retail} & \textsc{ToolBench} & \\
\midrule
\multicolumn{7}{@{}l}{\raisebox{-1pt}{\includegraphics[height=8pt]{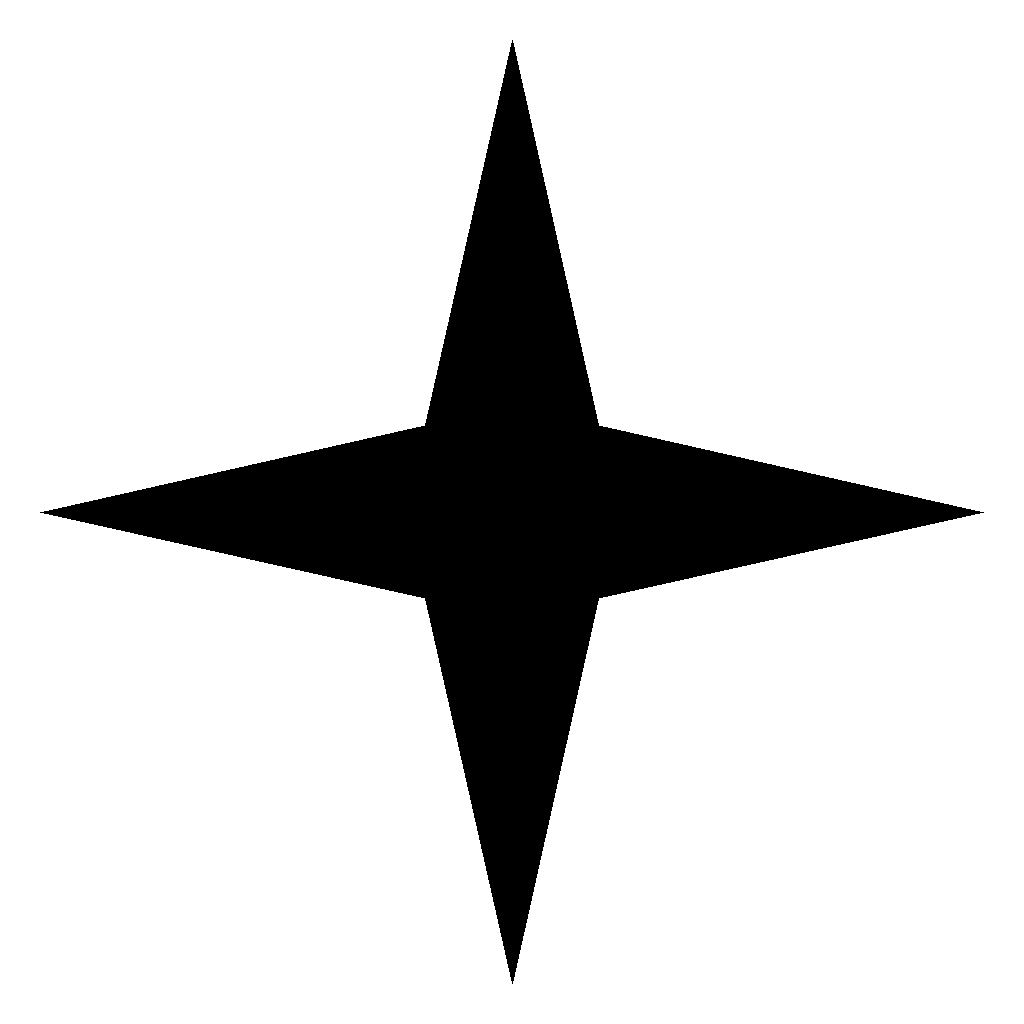}}\,\textbf{Gemma~3}} \\
\quad 1B & 31 [26,38] & 3 & 60 & 7 & 17 & 20 \\
\rowcolor{tablerow} \quad 4B & 56 [49,63] & 0 & 57 & 10 & 37 & 71 \\
\quad 12B & 73 [67,79] & 0 & 70 & 47 & 30 & 80 \\
\rowcolor{tablerow} \quad 27B & 78 [72,83] & 0 & 47 & 43 & 33 & 69 \\
\multicolumn{7}{@{}l}{\raisebox{-1pt}{\includegraphics[height=8pt]{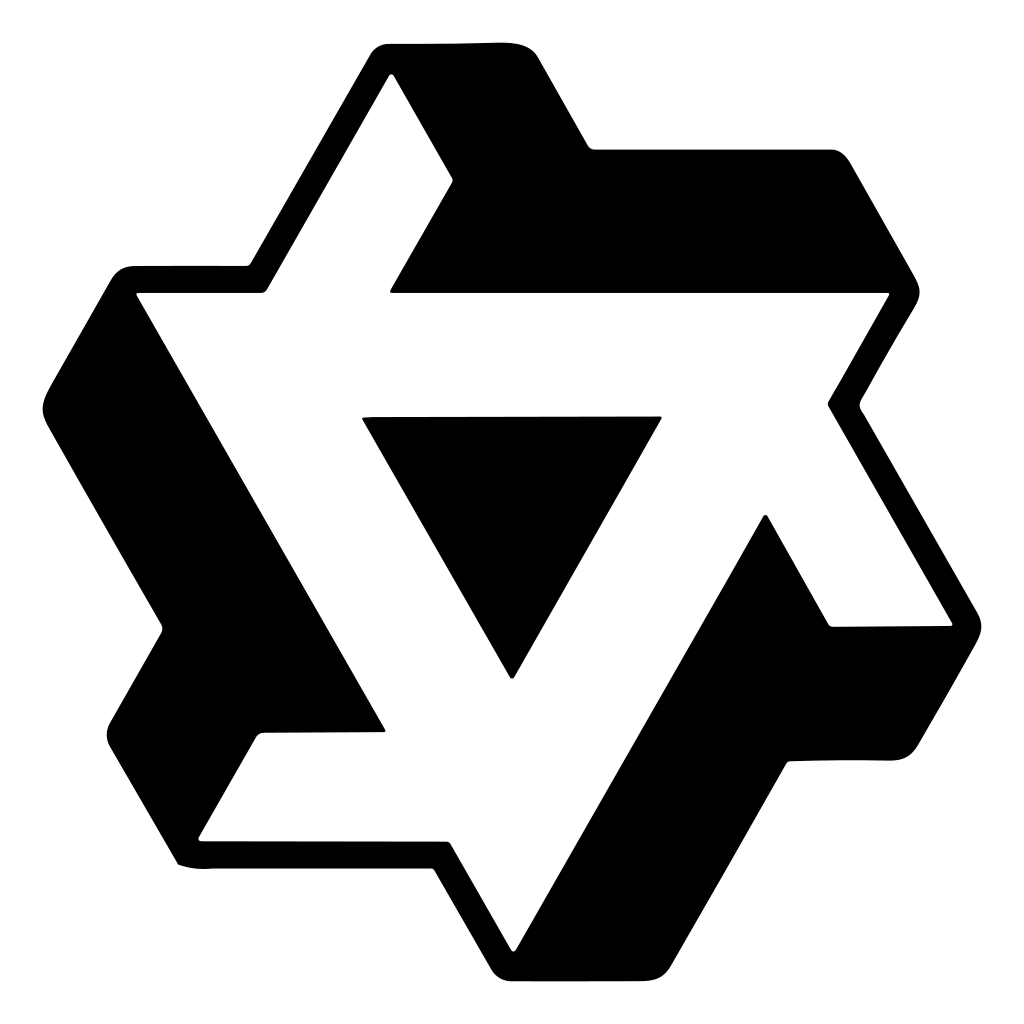}}\,\textbf{Qwen~3 / 2.5}} \\
\quad 0.6B & 45 [38,52] & 2 & 57 & 23 & 13 & 31 \\
\rowcolor{tablerow} \quad 1.7B & 69 [62,74] & 1 & 60 & 37 & 20 & 67 \\
\quad 4B & 71 [64,77] & 1 & 77 & 50 & 30 & 69 \\
\rowcolor{tablerow} \quad 8B & 77 [70,82] & 0 & 70 & 77 & 30 & 64 \\
\quad 14B & 78 [72,83] & 0 & 60 & 37 & 33 & 73 \\
\rowcolor{tablerow} \quad Qwen~2.5 7B & 73 [66,78] & 1 & 57 & 47 & 23 & 75 \\
\multicolumn{7}{@{}l}{\raisebox{-1pt}{\includegraphics[height=8pt]{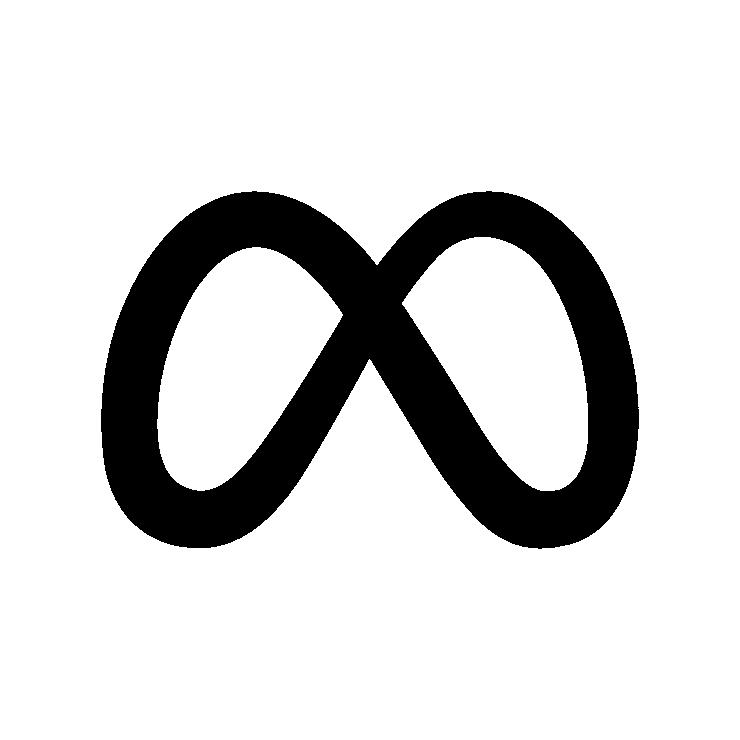}}\,\textbf{Llama~3.1}} \\
\quad 8B & 60 [53,66] & 0 & 60 & 57 & 27 & 73 \\
\bottomrule
\end{tabular}}

\end{table}

Steering depends little on the wording of the example requests, on the tool list or on knowing the source tool, but it fails on near-synonyms. Rebuilding the tool vectors from paraphrased requests leaves steering almost unchanged on the three representative models. The paraphrases include ``Is it raining in Tokyo?'' for ``What is the weather in Tokyo?'' and requests about other cities. Ambiguous or indirect requests such as ``Should I bring a jacket to Oslo?'' reduce steering (Appendix~\ref{app:robustness}). Removing five tools from the tool list, or adding five new ones, also changes little. Tool vectors built on the original tool list, applied to the changed tool list, still switch 18 to 20 of 20 sampled pairs on the two 4B models. Steering also works without knowing the source tool, by subtracting the average of all fifteen tool vectors from the target's tool vector instead. That direction switches 68--80\% as many calls as the steering direction at the last layer on the original tool list (Appendix~\ref{app:output_row}). Steering fails on near-synonyms. Pairs such as \texttt{send\_email} and \texttt{send\_message}, whose tool vectors are nearly parallel, switch in none of the trials. Among the ten similar candidates of MetaTool, steering still switches most calls, and fewer as the two tools grow more similar (Appendix~\ref{app:robustness}). Results with tool descriptions and with requests in other languages are in Appendices~\ref{app:desc_ablation} and~\ref{app:crosslingual}.

\phantomsection
\label{sec:alternatives}
\label{sec:discussion}
The direction carries more than the request's topic. On $\tau$-bench airline, where all tools share one topic, a linear probe on the hidden state predicts the upcoming call from real multi-turn trajectories and beats shuffled-label and noise controls. Steering between airline-specific tools still switches the full name in 52\% and 44\% of pairs on Gemma~3 4B and Qwen~3 4B (Appendix~\ref{app:within_topic_probe}). Steering also switches the call rather than blurring it. As we scale the direction up, the similarities to the tool vectors change smoothly but the emitted call flips at one point (Figure~\ref{fig:linearity}). Controls for the tool list, the tool names, the list order and the prompt are in Appendices~\ref{app:menu_removal}, \ref{app:anon_names}, \ref{app:position_bias} and~\ref{app:matched_prompt}.

\begin{figure}[t]
\centering
\includegraphics[width=1.0\linewidth]{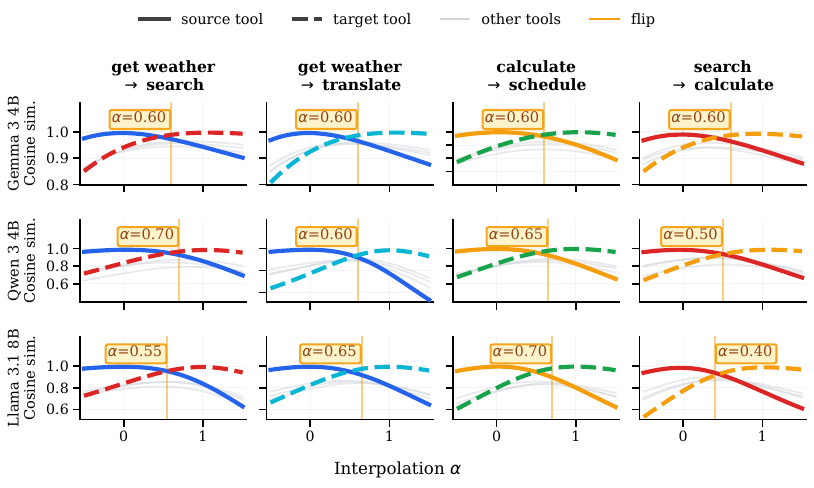}
\caption{Interpolating along $\direction_{a \to b}$ for four tool pairs (columns) on Gemma~3 4B, Qwen~3 4B and Llama~3.1 8B (rows). Similarities to the two tool vectors change smoothly, while the emitted call changes in one step in every panel. The transition point depends on the pair and model. Similarities to other tools (grey) remain lower throughout.}
\label{fig:linearity}
\end{figure}

\textbf{Model scale.} Scored by the prefix rule, steering works on every instruction-tuned model of 4B or larger and unevenly below that size, and a larger model is not always better (Figure~\ref{fig:emergence}). On the synthetic tool list, the instruction-tuned models of 4B or larger span 81 to 100\% and the four smaller models 27 to 92\%, so the two groups overlap. The smallest model, Gemma~3 270M, stays near its unsteered accuracy (Table~\ref{tab:emergence}). The steering advantage of instruction-tuned over base models is smallest at 27B, where base models sometimes match their instruction-tuned versions. Table~\ref{tab:emergence} and Appendix~\ref{app:benchmark_heatmap} report the individual benchmarks and uncertainty intervals.

\textbf{Shared name prefixes.} Prefix sharing limits steering on the original 15-tool list. None of the eight instruction-tuned models of 4B or larger switches between two \texttt{get\_} tools, and no other source reaches more than one \texttt{get\_} target, as the bound in Section~\ref{sec:method} predicts. The three strongest models reach 161 to 163 of the 165 pairs the bound allows, and the eight models switch 72 to 99\% of those 165 (Appendix~\ref{app:output_row}). Giving each tool a unique first token (the renamed tool list) improves the \texttt{get\_} targets to the level of the others (Appendix~\ref{app:failure}).

\textbf{Real API names.} On real API tool lists, full-name scores are substantially lower than prefix-rule scores (Table~\ref{tab:generalization}, Appendices~\ref{app:realapi_scoring} and~\ref{app:full_ci}). The cross-domain ToolBench tool list shows the same prefix limit. The tool list has three APIs in each of five domains, and the three APIs of a domain share a first token. So a prefix hit can mean the call moved to the right domain but not to the target API. We compare the steering direction with a direction built only from domain averages (the mean of a domain's three tool vectors) on the same pairs. The two directions give the same full-name hits and misses on all three representative models. Nearly all generated continuations are also identical between the two directions. Steering between APIs of one domain succeeds on none of the sampled pairs (Table~\ref{tab:toolbench_domain}). So on this tool list the direction carries the domain, not the API.

\textbf{Arguments after the name.} After a successful switch, some models adapt the JSON arguments to the new tool's schema (Table~\ref{tab:args_body}, with successes and failures in Appendix~\ref{app:generations}). Supplying the target name directly, without steering, produces similar adaptation, so the argument changes can come from the name alone (Appendix~\ref{app:schema_ar}). Attempts to control an argument value, change the output format or transfer a vector across model families failed in our tests (Appendix~\ref{app:limitations_extended}).

\begin{table}[h!]
\centering
\caption{Steered calls on Gemma~3 4B whose arguments carry the request (more in Table~\ref{tab:generations}).}
\label{tab:args_body}
\small
\setlength{\tabcolsep}{6pt}
\resizebox{\ifdim\width>\columnwidth\columnwidth\else\width\fi}{!}{%
\begin{tabular}{@{}l l l l@{}}
\toprule
\textsc{Request} & \textsc{Switch} & \textsc{Unsteered arguments} & \textsc{Steered arguments} \\
\midrule
\textit{``What is the weather in Tokyo?''} & \texttt{weather}$\to$\texttt{search} & \texttt{"input": "Tokyo"} & \texttt{\textcolor{steered}{"query": "weather in Tokyo"}} \\
\textit{``What is Apple stock price?''} & \texttt{stock}$\to$\texttt{search} & \texttt{"ticker": "AAPL"} & \texttt{\textcolor{steered}{"query": "Apple stock price"}} \\
\textit{``Translate hello to French.''} & \texttt{translate}$\to$\texttt{music} & \texttt{"text": "hello", "lang": "fr"} & \texttt{\textcolor{steered}{"song": "hello"}} \\
\textit{``Translate hello to French.''} & \texttt{translate}$\to$\texttt{calculate} & \texttt{"text": "hello", "lang": "fr"} & \texttt{\textcolor{steered}{"expr": "translate('hello', 'fr')"}} \\
\bottomrule
\end{tabular}}

\end{table}

\section{Where the direction acts}
\label{sec:mechanism}

\subsection{Across depth}
\label{sec:depth}

\begin{figure}[t]
\centering
\includegraphics[width=\linewidth]{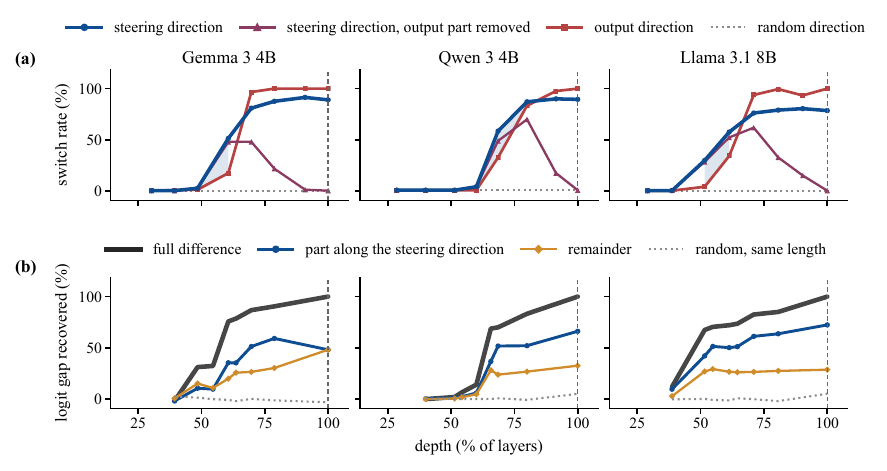}
\caption{Where the steering direction acts, one model per family. \textbf{(a)} Switch rate by layer when four directions of the same length are added: the steering direction, the same direction with its part along the output direction removed, the output direction, and a random direction (renamed 15-tool list, 210 held-out pairs). Shading marks layers where the steering direction beats the output direction. \textbf{(b)} Share of the logit gap between two requests recovered by patching their hidden-state difference in full, only along the steering direction, only in the remaining dimensions, or along a length-matched random direction. Dashed line: the last layer.}
\label{fig:depth_3fam}
\end{figure}

A direction that only raises the target name's score would switch the call without telling us anything about how the model chooses the tool. The steering direction partly aligns with the tool names' output weights (Figure~\ref{fig:alignment}). We therefore compare it at each depth with the \emph{output direction}, a direction built from those output weights. Appendices~\ref{app:output_row} and~\ref{app:schema_ar} also steer with the output direction alone and with a direct boost to the target name's first-token score.
Near the last layer, the output direction switches at least as many held-out pairs as the steering direction (Figure~\ref{fig:depth_3fam}a). Earlier in the network, the order reverses on all three models, on both the original and renamed tool lists (Table~\ref{tab:heldout_directions}). We call the layer with the largest difference between the two directions the peak layer. The peak layer is chosen on the same requests we evaluate. At that fixed layer, the steering direction still beats the output direction when a different request is held out (Table~\ref{tab:fold_peak}). The depth comparison thus points to a contribution of the steering direction before the last layer.
Next we remove the small part of the steering direction that lies along the output direction. What remains is rescaled to the original length and tested on held-out requests. Steering then keeps most of its effect at the peak layers but nearly vanishes at the last layer (Table~\ref{tab:heldout_directions}). Removing the parts along all fifteen tools' output directions gives the same picture. At the peak layers, 92--100\% of the steering direction's switches remain, and at the last layer at most 2 of 210 pairs switch (Table~\ref{tab:orthall_heldout}). So the part along the output direction does the work at the last layer but hardly matters in the middle layers.

The model's own hidden-state difference between two requests also lies partly along the steering direction (Figure~\ref{fig:depth_3fam}b). We take two requests that call different tools and patch the difference between their last-prompt-token states into the run of the first request, a form of activation patching. We compare patching the full difference with patching only its part along the steering direction, built from separate example requests. Recovery measures how much of the original logit gap between the two tools the patch closes. Appendix~\ref{app:direction_patch} defines the score, the selected layers and the controls.
Across Gemma~3 4B and 12B, Qwen~3 4B and 8B, and Llama~3.1 8B, patching the full difference recovers at least 0.2 of the original logit gap at 29 layers in total. At those layers, the steering-direction part recovers 30--84\% of the full patch's effect, and the rest of the difference also contributes. Random directions of matched length recover less. Per unit of length, the steering-direction part remains more effective than the remainder, although the advantage is small at some layers (Appendix~\ref{app:direction_patch}).
Together, the depth comparison, the removal and the patching show that the steering direction does more than raise the target name's score, and that part of it acts before the last layer. Steering itself still works best at the last layer in our tests (Appendix~\ref{app:output_row}).

\subsection{Which components carry the choice}
\label{sec:circuit}
Reading and steering both use the last prompt token. To examine how the model's components contribute to the tool choice at the last prompt token, we replace component outputs between requests.
Three observations suggest where to look. Sparse autoencoder features at sampled layers are selective for individual tools (Appendix~\ref{app:sae_full}). On Gemma~3 4B, attention from the last prompt token to the tool names rises late in the network (Figure~\ref{fig:attention}). Replacing the whole hidden state one layer at a time also points to the later layers (Appendix~\ref{app:attribution_method}). These observations are correlational. To test cause, we patch single components and token positions.

For each attention head and feed-forward layer, we swap in its output from a request for a different tool. We then measure how much the original tool's first-token score changes, averaged over 10 tool pairs (Appendix~\ref{app:patching_details}). On Gemma~3 4B, two attention heads in the middle of the network have the largest individual effects. Comparing the strongest components also shows a larger attention contribution than a feed-forward contribution on Gemma, whereas attention and feed-forward are comparable on Qwen (Table~\ref{tab:per_component_norm}). We compare the strongest components rather than sums, because an architecture with more heads would sum higher.

\begin{table}[!htb]
\centering
\caption{Activation-patching importance for attention heads and MLP layers. \textsc{Total} sums all components, and \textsc{Top-$k$ sum} the $k$ strongest in each category. On Gemma~3 4B, top-$k$ heads have 2--3$\times$ the effect of top-$k$ MLPs at each reported $k$. On Qwen~3 4B, the two are similar.}
\label{tab:per_component_norm}

\small
\setlength{\tabcolsep}{6pt}
\resizebox{\ifdim\width>\columnwidth\columnwidth\else\width\fi}{!}{%
\begin{tabular}{@{}l rr rr rr rr@{}}
\toprule
& \multicolumn{2}{c}{\textsc{Total}} & \multicolumn{2}{c}{\textsc{Top-1}} & \multicolumn{2}{c}{\textsc{Top-3 sum}} & \multicolumn{2}{c}{\textsc{Top-5 sum}} \\
\cmidrule(lr){2-3} \cmidrule(lr){4-5} \cmidrule(lr){6-7} \cmidrule(lr){8-9}
& Attn & MLP & Attn & MLP & Attn & MLP & Attn & MLP \\
\midrule
\raisebox{-1pt}{\includegraphics[height=8pt]{figures/logos/gemma.png}}\,\textbf{Gemma 4B} ($n_{\text{a}}{=}272$, $n_{\text{m}}{=}34$)    & 73.3 & 17.8 & 6.48 & 2.13 & 12.6 & 4.7  & 16.1 & 6.8  \\
\raisebox{-1pt}{\includegraphics[height=8pt]{figures/logos/qwen.png}}\,\textbf{Qwen 4B}  ($n_{\text{a}}{=}1{,}152$, $n_{\text{m}}{=}36$) & 25.8 & 3.4  & 0.44 & 0.41 & 0.96 & 1.05 & 1.45 & 1.49 \\
\bottomrule
\end{tabular}}

\end{table}

A separate experiment replaces each layer's attention or feed-forward output at selected token positions. In the middle of the network, replacing the attention output at the last prompt token restores about 0.4 of the gap between the two tools' first-token scores on Gemma and Qwen. At the first layer, replacing Qwen's attention or feed-forward output at every request token restores 0.99 of that gap. This nearly replaces the whole request with the other one, so it cannot locate the choice (Appendix~\ref{app:position_patching}).

\section{Reading a call before it is made}
\label{sec:errorrisk}
The same tool vectors also flag likely wrong calls. Comparing a request's hidden state with each tool vector ranks calls by error risk before any token is generated. Wrong calls tend to have a smaller gap between the request's similarities to its two closest tool vectors (Figure~\ref{fig:hero}, step 3). On BFCL, the gap ranks errors with an AUROC of 0.61 to 0.78 across four models (Table~\ref{tab:monitor}, Appendix~\ref{app:risk_coverage}).

\begin{table}[!t]
\centering
\caption{Ranking wrong calls on BFCL by AUROC, with 95\% bootstrap intervals, for the cosine gap and three output-probability signals. The column $n$ counts parseable calls, errors in parentheses. The last three columns give the paired-bootstrap probability that the gap scores higher (Appendix~\ref{app:risk_coverage}).}
\label{tab:monitor}
\scriptsize
\setlength{\tabcolsep}{4.2pt}
\resizebox{\ifdim\width>\columnwidth\columnwidth\else\width\fi}{!}{%
\begin{tabular}{@{}lccccccccc@{}}
\toprule
\textsc{Model} & $n$ (\textsc{err.}) & \multicolumn{4}{c}{\textsc{AUROC} [95\% CI]} & \multicolumn{3}{c}{$\Pr(\text{cos wins})$} \\
\cmidrule(lr){3-6}\cmidrule(lr){7-9}
 & & \textsc{Cosine gap} & \textsc{First token} & \textsc{Generated name} & \textsc{Full name} & vs first & vs gen. & vs full \\
\midrule
\raisebox{-1pt}{\includegraphics[height=8pt]{figures/logos/gemma.png}}\,Gemma~3 4B & 495 (20) & 0.76 [0.65,0.84] & 0.56 [0.47,0.64] & 0.48 [0.36,0.60] & 0.83 [0.75,0.90] & 1.00 & 1.00 & 0.14 \\
\rowcolor{tablerow} \raisebox{-1pt}{\includegraphics[height=8pt]{figures/logos/gemma.png}}\,Gemma~3 12B & 496 (13) & 0.78 [0.64,0.90] & 0.61 [0.49,0.69] & 0.61 [0.40,0.82] & 0.88 [0.74,0.96] & 0.98 & 0.89 & 0.11 \\
\raisebox{-1pt}{\includegraphics[height=8pt]{figures/logos/qwen.png}}\,Qwen~3 4B & 455 (61) & 0.61 [0.54,0.69] & 0.56 [0.48,0.63] & 0.56 [0.48,0.64] & 0.54 [0.47,0.62] & 0.86 & 0.84 & 0.91 \\
\rowcolor{tablerow} \raisebox{-1pt}{\includegraphics[height=8pt]{figures/logos/llama.png}}\,Llama~3.1 8B & 469 (16) & 0.78 [0.71,0.85] & 0.56 [0.45,0.67] & 0.67 [0.50,0.82] & 0.76 [0.65,0.87] & 1.00 & 0.87 & 0.58 \\
\bottomrule
\end{tabular}}

\end{table}

We compare the gap with three signals from the model's own output probabilities. The first is the margin between the two best candidates' first tokens, read from the probabilities the model already computes for its first output, so the margin costs nothing extra. The gap ranks errors better than the first-token margin on three of the four models. The second is the margin between summed log-probabilities of the full candidate names. The full-name margin ranks errors about as well as the gap, within noise on every model, but requires scoring every candidate name (Table~\ref{tab:monitor}). The third, free once the call is written, is the log-probability of the tool name the model wrote. The gap scores higher than the generated-name log-probability on all four models, but the difference is clear only on Gemma~3 4B, where the generated-name log-probability does no better than chance. Each model contributes only 13 to 61 errors, so we only rank calls and set no threshold.
Timed on all four models, the gap and the first-token margin add under a tenth of a millisecond to the forward pass over the prompt. Scoring the full names adds more than twice the time of the prompt pass (Appendix~\ref{app:monitor_cost}).

\section{Related work}

\citet{chen2026heading}, \citet{chen2026tunable} and \citet{wang2026asa} steer whether the model calls a tool at all. Our interventions target a different choice, the tool the model picks.
All of these interventions rest on the linear representation hypothesis, the idea that language models represent high-level concepts as directions in activation space \citep{park2024linear, nanda2023emergent}. Such directions have been found for sentiment \citep{tigges2023linear}, truthfulness \citep{marks2024geometry} and refusal \citep{arditi2024refusal}. Several methods use these directions to steer model behavior at inference time \citep{li2023inference, turner2024steering, zou2023representation, rimsky2024caa}. The closest comparison is \citet{todd2024function}, who identify ``function vectors'' for in-context tasks, whereas we study selection among named tools in a fixed tool list. \citet{todd2024function} test how much a function vector does beyond pushing the model toward its output tokens. We run the same test on the steering direction across network depth, and with its part along the output direction removed (Section~\ref{sec:depth}).
Base models also carry readable tool labels (Section~\ref{sec:reading}), which connects to work on how fine-tuning changes model behavior. Fine-tuning can adapt existing abilities with limited internal changes \citep{jain2024mechanistically} and strengthen mechanisms already present in the base model \citep{prakash2024finetuning}. Alignment tuning can likewise change how a direction affects generation while leaving the direction itself intact \citep{lee2024mechanistic}. Which training stage builds the tool-use mechanism remains open.
We examine internal representations with activation patching \citep{vig2020causal, wang2023interpretability, conmy2023automated} and sparse autoencoders (SAEs) \citep{cunningham2023sparse, bricken2023monosemanticity, templeton2024scaling}. Activation patching tests the effect of internal states, and SAEs separate activations into features. Principal component analysis (PCA) gives a geometric summary of the tool vectors, which we treat as an illustration only (Appendix~\ref{app:pca}).

\section{Conclusion}

One average hidden state per tool, its tool vector, is enough to read the tool a request asks for and to switch the model's call to another tool. At the last layer, the switch could simply be raising the target name's score. Earlier in the network, however, the steering direction switches more pairs than the output direction. Removing the part of the steering direction aligned with the output direction keeps steering working at middle layers. Patching only the part of two requests' hidden-state difference that lies along the steering direction recovers part of the effect of patching the whole difference. The three tests show that the direction helps choose the tool before the last layer, by an amount that varies across models and layers. On real API tool lists, the direction moved the call between domains rather than between APIs of one domain. Among similar tools on MetaTool it switched most calls, and fewer as the tools grew more similar. The same tool vectors also flag likely wrong calls in advance. How the model computes its preference and how to control multi-turn agents remain open.

\section*{Limitations}
Our steering results are single-turn. On twenty $\tau$-bench airline tasks, replacing the model's call with the readout's choice changes task success by $-$30 to $+$5 points against unsteered runs, across five models at the last layer. Steering inside agent loops remains untested (Appendix~\ref{sec:multiturn}).
Steering is weaker on real-API names and fails on near-synonymous tools, because near-synonyms have nearly parallel tool vectors. Changing which request is held out moves the last-layer rate by up to 19 percentage points on the three representative models. The peak layers were chosen on the evaluated requests and not re-selected on new ones. Mechanistic analyses use two to five of the eight models. The real-API columns reuse their test requests in the tool vectors, and only three models were rerun on held-out requests. The edit acts on the first output token, so it cannot separate two tools whose names share that token. The error signal rests on 13 to 61 errors per model, so we rank calls and set no threshold. On MetaTool, where the candidates are similar tools, the readout falls behind a sentence encoder on two of the three models. Component patching finds heads and layers that carry the preference, not a complete circuit. All requests are in English except the cross-lingual controls (Appendix~\ref{app:crosslingual}). Appendix~\ref{app:limitations_extended} gives the remaining limitations and evaluation sizes.

\subsection*{AI use statement}
In this work we used generative AI coding assistants (Anthropic's Claude through Claude Code, and OpenAI's Codex) for
\emph{implementation} and \emph{interpretation of results}.
They wrote and repaired the scripts that build every
figure in this paper from the stored experiment outputs, and they wrote the checks described in the
reproducibility statement. For interpretation, they computed quantities derived from those outputs,
such as ranges across cells and differences between conditions, and drafted sentences reporting
those quantities. Which results to report, and what the results mean, are our decisions.

OpenAI Codex proposed the new-query split with unsteered and equal-norm random controls in
Appendix~\ref{app:realapi_scoring}, and authored its fixed query strings from the existing tool list names
and signatures before running the models. These authored test inputs are separate from the official
benchmark test sets. Mathematical formulation and proofs are not applicable to this work.

We also used generative AI tools for three subtasks where disclosure is recommended. The tools drafted and edited prose for concision and to remove repetition, shortened figure captions, and searched the
manuscript for internal inconsistencies. We verified and fixed the inconsistencies found. We
used the tools to search for related work and to locate sources for claims we cite, and we checked what
the tools returned against those sources before citing any source. All
code the tools wrote is covered by the implementation disclosure above.

We take responsibility for the final content of this work, including text, claims or artifacts produced
with the aid of generative AI.

\subsection*{Ethics statement}
\textbf{Dual-use and defense.} Access to a model's internal states could allow an operator to redirect consequential tool calls without indicating the intervention in the generated response. The same access can help evaluators test which components affect selection and inspect signals associated with errors (Appendix~\ref{sec:monitor}). These signals are a research tool and would need further testing before use as a safeguard. The study uses publicly available models. Independent checks of tool permissions and argument schemas remain relevant to detecting or preventing harmful calls.

\subsection*{Reproducibility statement}
Model checkpoints and analysis resources are listed in Table~\ref{tab:models}. We use Gemma Scope~2 \citep{lieberum2024gemmascope}, NeuronPedia \citep{neuronpedia2024}, TransformerLens \citep{nanda2022transformerlens}, SAELens \citep{bloom2024saelens} and HuggingFace \texttt{transformers}. Appendix~\ref{app:eval_protocol} records the implementation details. Code and data are attached as supplementary archives. The verification suite, \texttt{verify\_paper.py}, runs 95 checks. It compares tables with stored outputs cell by cell, regenerates figures, checks manuscript consistency, verifies the new-query split, saved control vectors and per-generation labels, and deliberately alters table cells to test whether the checks detect errors. Any disagreement causes a non-zero exit. Tables without archived values are explicitly identified. Missing producer scripts and source records are also listed. The supplied code cannot regenerate those historical results.

\bibliographystyle{iclr2027_conference}
\bibliography{references}

\appendix
\renewcommand{\topfraction}{0.9}\renewcommand{\bottomfraction}{0.7}\renewcommand{\textfraction}{0.07}\renewcommand{\floatpagefraction}{0.75}
\setcounter{section}{0}
\renewcommand{\thesection}{A.\arabic{section}}
\setcounter{secnumdepth}{2}
\setcounter{figure}{0}\renewcommand{\thefigure}{A\arabic{figure}}
\setcounter{table}{0}\renewcommand{\thetable}{A\arabic{table}}

\section{Setup and protocol}

\subsection{Tool definitions}
\label{app:tools}
The 15 tools used in our experiments span diverse domains: information retrieval (get\_weather, search, get\_stock\_price, get\_news), computation (calculate, convert\_currency), communication (send\_email, create\_note), scheduling (schedule, set\_alarm, book\_flight), navigation (get\_directions), entertainment (play\_music), food ordering (order\_food), and language (translate).

\subsection{Models and resources}
\label{app:models}

Table~\ref{tab:models} lists all models used in this work. Table~\ref{tab:resources} lists SAE sources and software tools. The base models appear in the readout comparison, and the Qwen~3 4B base also in generation and output-projection controls. Component patching uses Gemma~3 4B and Qwen~3 4B. The output-row controls add the Gemma and Llama base models, and the SAE analysis uses whichever models its decoders cover (Appendix~\ref{app:sae_full}). These analyses interpret the hidden states but are not needed to build or apply the steering direction.

\begin{table}[htbp]
\centering
\caption{All models used in this work. PT/base = pre-trained; all others are instruction-tuned. The six base models of the main experiments are the five Gemma PT models and Llama~3.1 8B base; Qwen~3 4B Base appears only in the base-generation control (Appendix~\ref{app:base_it}).}
\label{tab:models}

\footnotesize
\setlength{\tabcolsep}{6pt}
\rowcolors{2}{tablerow}{white}
\resizebox{\ifdim\width>\columnwidth\columnwidth\else\width\fi}{!}{\begin{tabular}{@{}llrl@{}}
\toprule
\textsc{Family} & \textsc{Model} & \textsc{Params} & \textsc{HuggingFace ID} \\
\midrule
\raisebox{-1pt}{\includegraphics[height=8pt]{figures/logos/gemma.png}}\,Gemma~3 & Gemma~3 270M IT & 270M & \texttt{google/gemma-3-270m-it} \\
 & Gemma~3 1B IT & 1B & \texttt{google/gemma-3-1b-it} \\
 & Gemma~3 4B IT & 4B & \texttt{google/gemma-3-4b-it} \\
 & Gemma~3 12B IT & 12B & \texttt{google/gemma-3-12b-it} \\
 & Gemma~3 27B IT & 27B & \texttt{google/gemma-3-27b-it} \\
 & Gemma~3 270M PT & 270M & \texttt{google/gemma-3-270m} \\
 & Gemma~3 1B/4B/12B/27B PT & --- & \texttt{google/gemma-3-\{1b,4b,12b,27b\}-pt} \\
\midrule
\raisebox{-1pt}{\includegraphics[height=8pt]{figures/logos/qwen.png}}\,Qwen~3 & Qwen~3 0.6B & 0.6B & \texttt{Qwen/Qwen3-0.6B} \\
 & Qwen~3 1.7B & 1.7B & \texttt{Qwen/Qwen3-1.7B} \\
 & Qwen~3 4B & 4B & \texttt{Qwen/Qwen3-4B} \\
 & Qwen~3 8B & 8B & \texttt{Qwen/Qwen3-8B} \\
 & Qwen~3 14B & 14B & \texttt{Qwen/Qwen3-14B} \\
\midrule
 & Qwen~3 4B Base & 4B & \texttt{Qwen/Qwen3-4B-Base} \\
\midrule
\raisebox{-1pt}{\includegraphics[height=8pt]{figures/logos/qwen.png}}\,Qwen~2.5 & Qwen~2.5 7B Instruct & 7B & \texttt{Qwen/Qwen2.5-7B-Instruct} \\
\midrule
\raisebox{-1pt}{\includegraphics[height=8pt]{figures/logos/llama.png}}\,Llama~3.1 & Llama~3.1 8B Instruct & 8B & \texttt{meta-llama/Llama-3.1-8B-Instruct} \\
 & Llama~3.1 8B (base) & 8B & \texttt{meta-llama/Llama-3.1-8B} \\
\bottomrule
\end{tabular}}

\rowcolors{1}{}{}
\end{table}

\begin{table}[htbp]
\centering
\caption{SAE sources and software tools.}
\label{tab:resources}

\footnotesize
\setlength{\tabcolsep}{6pt}
\rowcolors{2}{tablerow}{white}
\resizebox{\ifdim\width>\columnwidth\columnwidth\else\width\fi}{!}{\begin{tabular}{@{}lll@{}}
\toprule
\textsc{Resource} & \textsc{Coverage} & \textsc{Source} \\
\midrule
\raisebox{-1pt}{\includegraphics[height=8pt]{figures/logos/gemma.png}}\,Gemma Scope 2 SAEs & Gemma~3 (L9/17/22/29) & \texttt{gemma-scope-2-4b-it-res} \\
NeuronPedia SAEs & Qwen~3 (all layers) & \href{https://neuronpedia.org}{\nolinkurl{neuronpedia.org}} \\
Goodfire SAEs & Llama~3.1 8B (L19) & \citet{goodfire2024} \\
\midrule
TransformerLens & Activation caching & v2.17+ \citep{nanda2022transformerlens} \\
SAELens & SAE loading/analysis & v6.37+ \citep{bloom2024saelens} \\
\bottomrule
\end{tabular}}
\rowcolors{1}{}{}

\end{table}

\subsection{Evaluation protocol}
\label{app:eval_protocol}

\textbf{Prompt templates.} The steering prompts below open the tool call before
generation, placing the final prompt token at the quote preceding the tool name. We read
and steer the residual at this position, the last prompt token. Instruction-tuned models use their chat
templates; base models use the bare answer prompt shown below. The doubled brace in
\texttt{\{\{"tool": "} is literal and shared by the compared conditions. In these
templates, \texttt{\{tools\}} denotes the tool list and \texttt{\{query\}} the user query:

{\small\begin{verbatim}
Gemma 3 (BOS in template)
<bos><start_of_turn>user
{tools}
{query}<end_of_turn>
<start_of_turn>model
{{"tool": "

Qwen 3 / 2.5 (BOS from tokenizer)
<|im_start|>user
{tools}
{query}<|im_end|>
<|im_start|>assistant
{{"tool": "

Llama 3.1 (BOS in template)
<|begin_of_text|><|start_header_id|>user
<|end_header_id|>

{tools}
{query}<|eot_id|><|start_header_id|>
assistant<|end_header_id|>

{{"tool": "

Base checkpoints (BOS from tokenizer)
{tools}
{query}
Answer: {{"tool": "
\end{verbatim}
}

\noindent Both braces reach the model because the harness uses string replacement
rather than \texttt{format}; generated calls mirror them with \texttt{\}\}}. Neither
prefix nor full-name scoring uses the brace count, but reproductions must retain the
doubled form to reproduce the activations. Repeating the held-out 15-tool run with a single
brace moves the full-name count by between $-3$ and $+9$ of 210 on the three representative
models (Table~\ref{tab:fold_rotation}). The last five token positions have the same
structure across the three families (\texttt{\{}, \texttt{\{"}, \texttt{tool}, \texttt{":},
\texttt{~"} on Gemma~3 4B), ending at the final quote.

\textbf{Steering evaluation.} The original harness reports prefix containment and
five-character containment over five generated tokens. Its ``exact'' field denotes the
latter, not a complete-name match. We recount the full-name 15-tool rates of Table~\ref{tab:generalization} from those
saved generations, requiring the complete target name with no following name character.
The full and prefix columns use the same 210 held-out ordered pairs. Older real-API
columns retain their original rules and sample sizes (Table~\ref{tab:full_ci}). The paired
reruns in Appendix~\ref{app:realapi_scoring} instead allow up to 24 tokens and require the
first identifier in the tool-name slot to equal the target; later mentions and longer
names containing it do not count. Wilson intervals describe sampling uncertainty. The
complete-name null $1/K$ does not apply to scores that group tools by prefix.

\textbf{BFCL evaluation.} Leave-one-out cross-validation constructs each tool vector
from the other requests mapping to the same function set, excluding the test request.
The generation parser (\texttt{parse\_fixed}) strips Gemma's \texttt{```tool\_code} prefix,
then tries exact-startswith matching followed by substring matching. Substring matching
can count a natural-language mention as a tool call, so the base-generation scores are
upper bounds on correctly formatted tool selection.

The monitor's \emph{format-excluded} population retains only outputs that parse into a
tool name; an error is a parseable call naming the wrong tool. Of 500 examples per model,
this excludes 5 on Gemma~3 4B, 4 on Gemma~3 12B, 31 on Llama~3.1 8B, and 45 on Qwen~3 4B.
The reported monitor therefore evaluates selection errors among parseable calls; malformed outputs are outside this population.

\textbf{PCA evaluation.} $k_{90}$ is the minimum number of principal components
capturing 90\% of variance across tool vectors (Appendix~\ref{app:pca}). We compare
it with random points in $\mathbb{R}^{d}$ using 50--100 Monte Carlo draws. The matched
control instead preserves the 15-tool prompt prefix and replaces tool requests with
non-tool topics (Appendix~\ref{app:matched_prompt}).

\subsection{Full-name checks on real API tool lists}
\label{app:realapi_scoring}

Appendix~\ref{app:full_ci} gives intervals for the earlier experiments that mixed the two scoring rules. Prefix matches overstate complete API selection on ToolBench's cross-domain tool list
(Table~\ref{tab:tb_fullname}), since each domain prefix is shared by three APIs. We rerun the
same 30 ordered cross-domain pairs on all 18 models and score each continuation
both by the first-token rule and by the full-name rule, which requires the first identifier in the tool-name slot to equal the complete target name. Generation allows up to 24 tokens,
enough for every name; later mentions in arguments or explanations do not count.

These reruns retain the prompt templates, two example requests per tool, and last-layer
displacement at $\alpha{=}0.7$. Test requests also contribute to the tool vectors. Both
scoring columns use the same generations. Table~\ref{tab:generalization} reports
their full-name ToolBench scores; prefix-scored summaries are in
Tables~\ref{tab:emergence} and~\ref{tab:full_ci}. Under uniform random choice among the
15 APIs, the domain-prefix baseline is 20\% and the full-name baseline is $1/15$.

\begin{table}[htbp]
\centering
\caption{ToolBench cross-domain target-match rates (\%, $N{=}30$ pairs per model), with both rules scored on the same generations. The first column counts the target domain's prefix in the first token; the second requires the complete target name at the start of the continuation. The tool list has three APIs per domain. Requests are not held out.}
\label{tab:tb_fullname}

\footnotesize
\setlength{\tabcolsep}{7pt}
\begin{tabular}{@{}l rr@{}}
\toprule
Model & Domain prefix & Full name \\
\midrule
Gemma~3 270M IT & 27 & 10 \\
Gemma~3 270M base & 27 & 10 \\
Gemma~3 1B IT & 93 & 17 \\
Gemma~3 1B base & 27 & 10 \\
Gemma~3 4B IT & 100 & 37 \\
Gemma~3 4B base & 27 & 10 \\
Gemma~3 12B IT & 100 & 30 \\
Gemma~3 12B base & 40 & 10 \\
Gemma~3 27B IT & 100 & 33 \\
Gemma~3 27B base & 90 & 27 \\
Qwen~3 0.6B & 77 & 13 \\
Qwen~3 1.7B & 100 & 20 \\
Qwen~3 4B & 100 & 30 \\
Qwen~3 8B & 100 & 30 \\
Qwen~3 14B & 100 & 33 \\
Qwen~2.5 7B & 100 & 23 \\
Llama~3.1 8B IT & 87 & 27 \\
Llama~3.1 8B base & 50 & 17 \\
\bottomrule
\end{tabular}

\end{table}

The compact $\tau$-bench rerun compares the same scoring rules on Gemma~3 4B, Qwen~3 4B,
and Llama~3.1 8B Instruct (Table~\ref{tab:tau_fullname}). Airline and retail list 14 and
15 tools, respectively, with nine tools per tool list eligible for pair sampling. Tool vectors use
two example requests, and cells use 30 pairs whose test requests also contribute to the tool vectors.
To reproduce the random sampling sequence, the harness consumes three 15-tool draws
before airline, then samples retail. The prefix rule reads the first five generated tokens; the full-name rule
matches the ToolBench rerun. Both measure emitted names, not API execution or agent
task completion.

\begin{table}[htbp]
\centering
\caption{Compact $\tau$-bench target-match rates (\%, $N{=}30$ pairs per cell). Prefix and full-name scores are paired within each environment. Each tool list has nine tools with tool vectors; the airline list has 14 tools and retail 15. These are last-layer checks with test requests reused in the tool vectors.}
\label{tab:tau_fullname}

\footnotesize
\setlength{\tabcolsep}{6pt}
\resizebox{\ifdim\width>\columnwidth\columnwidth\else\width\fi}{!}{\begin{tabular}{@{}l rrrr@{}}
\toprule
Model & Air prefix & Air full & Retail prefix & Retail full \\
\midrule
Gemma~3 4B IT & 77 & 57 & 50 & 10 \\
Qwen~3 4B & 83 & 77 & 73 & 50 \\
Llama~3.1 8B IT & 77 & 60 & 100 & 57 \\
\bottomrule
\end{tabular}}

\end{table}

\subsubsection{New request wordings and controls}
\label{app:realapi_heldout}

On new request wordings, the steering direction produces more full-name target hits than
either control in all nine model--tool-list combinations (Table~\ref{tab:realapi_heldout}). We
author one request per eligible tool from its name and signature before evaluation and
exclude it from every tool vector. The three models, tool lists, tool-vector construction, prompt
templates, and 30 ordered pairs match the compact checks above. We generate an unsteered
continuation for each request and test each pair with the steering direction and a Gaussian
random unit direction of the same displacement norm. Both edits act only on the first
forward pass, at the last prompt position after the final block, with $\alpha{=}0.7$.
All conditions use the same full-name rule and allow up to 24 generated tokens.

The Valid column counts target hits only where the unsteered model selected the intended
source; its separate denominator prevents pre-existing source errors from counting as
successful switches. Each source request is reused across targets, so the pairs are not
independent request samples. These authored wordings are separate from the official benchmark test cases, and the results measure tool names only.

\begin{table}[htbp]
\centering
\caption{Full-name counts on requests excluded from the tool vectors. Source: correct unsteered source choices over distinct requests. No edit, Random, and Steer: target hits on the same 30 ordered pairs. Valid: steered target hits over pairs whose unsteered output names the intended source. Each column reports its own denominator. Tool lists, tool-vector construction, and pairs match the compact scoring checks.}
\label{tab:realapi_heldout}

\footnotesize
\setlength{\tabcolsep}{6pt}
\resizebox{\ifdim\width>\columnwidth\columnwidth\else\width\fi}{!}{\begin{tabular}{@{}llrrrrr@{}}
\toprule
Model & Tool list & Source & No edit & Random & Steer & Valid \\
\midrule
Gemma~3 4B IT & TB-XD & 13/15 & 0/30 & 0/30 & 9/30 & 6/24 \\
Gemma~3 4B IT & $\tau$-air & 7/9 & 0/30 & 0/30 & 14/30 & 10/23 \\
Gemma~3 4B IT & $\tau$-ret & 9/9 & 0/30 & 0/30 & 8/30 & 8/30 \\
Qwen~3 4B & TB-XD & 14/15 & 0/30 & 0/30 & 10/30 & 10/27 \\
Qwen~3 4B & $\tau$-air & 8/9 & 1/30 & 1/30 & 19/30 & 17/28 \\
Qwen~3 4B & $\tau$-ret & 7/9 & 0/30 & 0/30 & 15/30 & 14/26 \\
Llama~3.1 8B IT & TB-XD & 14/15 & 0/30 & 0/30 & 6/30 & 6/27 \\
Llama~3.1 8B IT & $\tau$-air & 8/9 & 0/30 & 0/30 & 14/30 & 11/26 \\
Llama~3.1 8B IT & $\tau$-ret & 9/9 & 0/30 & 0/30 & 16/30 & 16/30 \\
\bottomrule
\end{tabular}}

\end{table}

\section{Reading tool labels}

\subsection{Cosine readout method}
\label{app:cosine_select}

The BFCL readout comparison in Section~\ref{sec:reading} uses tool vectors
and cosine similarity.

\textbf{Construct tool vectors.} For each candidate tool in a request's function list, we average the last-layer residual at the final prompt token, immediately before the tool name, over the other examples in that group whose answer is the same tool. The evaluated request is therefore left out, and the number of examples behind a mean varies by tool. A tool with no other example in its group uses its own description text in place of a request.

\textbf{Read out a new request.} We record the residual at the same position and predict the tool whose mean has the highest cosine similarity.

\textbf{Other readouts.} Two alternatives use the same leave-one-out examples (Table~\ref{tab:readout_extra}). One subtracts the average of the candidate tool vectors from the request's state and from each tool vector before taking the cosine. The other fits a linear discriminant classifier with shrinkage to those examples. Subtracting the average helps only on Llama~3.1 8B, by 4 points, and the classifier helps only on Qwen~3 4B, by 2 points; neither does better on the other models. The BFCL run skipped generation. Its plain cosine is within two points of Table~\ref{tab:bfcl_reading} except on Llama~3.1 8B, where it is 5 points higher. On MetaTool, where the ten candidates are similar tools, the plain cosine reads 43--77\% across the three models. It matches the model's own answer on Gemma and Llama and falls 14 points below it on Qwen. The linear classifier is higher than the cosine on every model. A text encoder on the request alone reaches 75\%, which only Llama's readout exceeds.

\begin{table}[htbp]
\centering
\caption{BFCL tool-name accuracy (\%) of three readouts on the same leave-one-out examples at the last layer: cosine to the tool vectors (Cosine), cosine after subtracting the average of the candidate tool vectors (Centered), and a shrinkage linear discriminant fit on the same examples (LDA). $n$ matches Table~\ref{tab:bfcl_reading}. LDA needs more examples than candidate tools, so its BFCL columns count only the examples where it could be fit. MetaTool: the similar-choice tool-selection task of \citet{huang2024metatool}, ten candidates per request in shuffled order, three other requests per candidate for its tool vector, and the instruction to answer with one tool name; Gen is the model's own answer (Qwen with a 200-token budget for its reasoning); the base models were not asked to generate.}
\label{tab:readout_extra}
\footnotesize
\resizebox{\ifdim\width>\columnwidth\columnwidth\else\width\fi}{!}{%
\begin{tabular}{@{}l r rr r rr | rrrr@{}}
\toprule
& \multicolumn{6}{c|}{\textsc{BFCL}} & \multicolumn{4}{c}{\textsc{MetaTool}, $n{=}995$} \\
& & \multicolumn{2}{c}{\textsc{All examples}} & \multicolumn{3}{c|}{\textsc{Examples where LDA fits}} & & & & \\
\cmidrule(lr){3-4}\cmidrule(lr){5-7}\cmidrule(lr){8-11}
\textsc{Model} & $n$ & Cosine & Centered & $n$ & Cosine & LDA & Gen & Cosine & Centered & LDA \\
\midrule
\raisebox{-1pt}{\includegraphics[height=8pt]{figures/logos/gemma.png}}\,Gemma~3 4B & 200 & 76 & 74 & 120 & 70 & 59 & 72 & 72 & 71 & 76 \\
\rowcolor{tablerow} \raisebox{-1pt}{\includegraphics[height=8pt]{figures/logos/gemma.png}}\,Gemma~3 4B base & 300 & 74 & 66 & 148 & 60 & 56 & -- & 47 & 37 & 53 \\
\raisebox{-1pt}{\includegraphics[height=8pt]{figures/logos/qwen.png}}\,Qwen~3 4B & 300 & 69 & 68 & 148 & 56 & 58 & 57 & 43 & 41 & 57 \\
\rowcolor{tablerow} \raisebox{-1pt}{\includegraphics[height=8pt]{figures/logos/llama.png}}\,Llama~3.1 8B & 200 & 58 & 62 & 120 & 53 & 53 & 75 & 77 & 76 & 79 \\
\raisebox{-1pt}{\includegraphics[height=8pt]{figures/logos/llama.png}}\,Llama~3.1 8B base & 300 & 82 & 82 & 148 & 70 & 61 & -- & 70 & 67 & 76 \\
\rowcolor{tablerow} Text encoder (BGE-small), request only & & & & & & & -- & 75 & 73 & 74 \\
\bottomrule
\end{tabular}}

\end{table}

Groups keep at least three candidate tools and at least two examples. For the generation
baseline, we strip \texttt{<think>...</think>} content and allow
\texttt{max\_new\_tokens = 200}, giving thinking-mode models such as Qwen~3 room to emit a
tool name after reasoning. A 30-token limit would truncate some outputs before the
tool name and inflate the readout advantage.

\subsection{BFCL v3 tool-name accuracy}
\label{app:bfcl_reading}

\begin{table}[htbp]
\centering
\caption{BFCL v3 tool-name accuracy: greedy generation versus cosine readout of the last-layer residual (max\_new\_tokens=200, thinking tags stripped; $N{=}300$ requests per base row and $N{=}200$ per instruction-tuned row). $\Delta$ uses unrounded scores and can differ by one point from the displayed subtraction. Other model's vectors: Gemma~3 4B readout against the other model's leave-one-out tool vectors at L17 / L22 / L33 ($N{=}150$), under the same bare prompt template. Text encoder: BGE-small embeds only the request, with no tool list or language-model activations, and classifies against leave-one-out tool vectors ($N{=}150$).}
\label{tab:bfcl_reading}

\footnotesize
\setlength{\tabcolsep}{7pt}
\resizebox{\ifdim\width>\columnwidth\columnwidth\else\width\fi}{!}{%
\begin{tabular}{l c c r c}
\toprule
\textsc{Model} & \textsc{Gen (\%)} & \textsc{Readout (\%)} & \textsc{$\Delta$ (pp)} & \textsc{Other model's vectors} \\
\midrule
\multicolumn{5}{@{}l}{\textit{Base models}} \\
\rowcolor{tablerow} \raisebox{-1pt}{\includegraphics[height=8pt]{figures/logos/gemma.png}}\,Gemma~3 1B base   & 2  & 69 & +67 & -- \\
\raisebox{-1pt}{\includegraphics[height=8pt]{figures/logos/gemma.png}}\,Gemma~3 4B base   & 3  & 75 & +72 & 55 / 51 / 31 \\
\rowcolor{tablerow} \raisebox{-1pt}{\includegraphics[height=8pt]{figures/logos/gemma.png}}\,Gemma~3 12B base  & 5 & 61 & +56 & -- \\
\raisebox{-1pt}{\includegraphics[height=8pt]{figures/logos/llama.png}}\,Llama~3.1 8B base & 10 & 82 & +72 & -- \\
\rowcolor{tablerow} Text encoder (BGE-small), request only & -- & 81 & -- & -- \\
\midrule
\multicolumn{5}{@{}l}{\textit{Instruction-tuned models}} \\
\rowcolor{tablerow} \raisebox{-1pt}{\includegraphics[height=8pt]{figures/logos/gemma.png}}\,Gemma~3 4B IT     & 92 & 74 & $-$18 & 53 / 52 / 52 \\
\raisebox{-1pt}{\includegraphics[height=8pt]{figures/logos/gemma.png}}\,Gemma~3 12B IT    & 95 & 66 & $-$30 & -- \\
\rowcolor{tablerow} \raisebox{-1pt}{\includegraphics[height=8pt]{figures/logos/llama.png}}\,Llama~3.1 8B IT   & 90 & 53 & $-$37 & -- \\
\bottomrule
\end{tabular}}

\end{table}

The base models' tool labels are readable despite weak generation, but a text encoder (BGE-small) reads them about as well from the request text alone (Table~\ref{tab:bfcl_reading}). Thus readout accuracy on this benchmark may reflect the request alone.
Instruction-tuned models generate more accurately than this readout.

For the cross-model comparison, a common prompt template matters. Transfer was much weaker when each model used its own template. With their own tool vectors under
the common template, base accuracy is 60 / 55 / 57 and IT accuracy is 68 / 62 / 55 at
L17 / L22 / L33; chance is 20\%. Transfer retains 78--93\% of own-vector accuracy at the
middle layers in both directions. At the last layer, IT tool vectors read the base model at
54\% of the base model's own-vector accuracy.

\subsection{Base and instruction-tuned models}
\label{app:base_it}

\subsubsection{Base vs instruction-tuned comparison}
\label{app:base_it_fig}

\textbf{Alignment with output-token directions.}
At the final layer used for steering, instruction-tuned directions align more strongly with the target token's output direction than base-model directions, by a factor of 1.5 to 1.9. This alignment does not explain the difference between the steering direction and the output direction at earlier layers (Section~\ref{sec:depth}). Instruction-tuned models also show larger logit changes under residual replacement at the matched layer (Table~\ref{tab:attribution_summary}). SAE counts follow neither effect. The 270M base model has more tool-specific features despite ineffective steering (Appendix~\ref{app:sae_full}).

\begin{figure}[htbp]
\centering
\includegraphics[width=\linewidth]{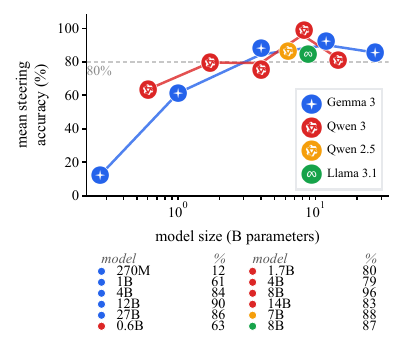}
\caption{Steering accuracy against model size, averaged over the synthetic 15-tool list, $\tau$-bench airline and $\tau$-bench retail under prefix scoring. The 15-tool term uses 210 held-out pairs; the two $\tau$-bench terms use $N{=}30$ per cell. Accuracy generally increases with scale, but the average combines different protocols. Table~\ref{tab:emergence} reports each benchmark separately.}
\label{fig:emergence}
\end{figure}

\textbf{Average across benchmarks.} Figure~\ref{fig:emergence}, cited from Section~\ref{sec:emergence}, averages the three benchmark columns specified in its caption.

Table~\ref{tab:emergence} gives per-cell steering accuracy for every model, with the instruction-tuned and base figures side by side wherever both models were tested; the patching peak layer and importance for every model are in Table~\ref{tab:attribution_summary}.

\newcommand{\ci}[2]{{\scriptsize[#1,#2]}}
\begin{table}[htbp]
\centering
\caption{Steering accuracy (\%, prefix match) across all 12 models, with instruction-tuned/base results where both were tested. Sw-15 uses 210 held-out pairs; the other columns use $N{=}30$ per cell with test requests reused in the tool vectors. These protocols differ from the held-out 5-tool test in Appendix~\ref{app:matrix}. Qwen base models are absent from this steering sweep; Qwen~3 4B base is included in Appendix~\ref{app:base_it}. \textbf{Bold}: ${\geq}$93\%. TB-XD results vary slightly across hardware reruns: Gemma~3 1B gives 24, 26 and 27 of 30, and Gemma~3 270M gives 6, 8 and 9. These estimates lie within one another's intervals; the table retains the stored run's values. Table~\ref{tab:full_ci} gives confidence intervals.}
\label{tab:emergence}

\footnotesize
\setlength{\tabcolsep}{6pt}
\renewcommand{\arraystretch}{1.05}
\resizebox{\ifdim\width>\columnwidth\columnwidth\else\width\fi}{!}{\begin{tabular}{@{}l cccc@{}}
\toprule
& \textsc{Sw-15} & \textsc{$\tau$-Air} & \textsc{$\tau$-Ret} & \textsc{TB-XD prefix} \\
& \multicolumn{4}{c}{\scriptsize\emph{IT / base where available}} \\
\midrule
\multicolumn{5}{@{}l}{\raisebox{-1pt}{\includegraphics[height=8pt]{figures/logos/gemma.png}}\,\textbf{Gemma~3}} \\
\rowcolor{tablerow} \quad 270M & 27 / 27 & 10 / 10 & 0 / 0 & 27 / 30 \\
\quad 1B   & 54 / 27 & 77 / 10 & 53 / 0 & 80 / 27 \\
\rowcolor{tablerow} \quad 4B   & 83 / 48 & \textbf{94} / 60 & 76 / 60 & \textbf{100} / 30 \\
\quad 12B  & \textbf{100} / 55 & 90 / 53 & 80 / \textbf{93} & \textbf{100} / 33 \\
\rowcolor{tablerow} \quad 27B  & \textbf{100} / 84 & 77 / \textbf{97} & 80 / 83 & \textbf{100} / 90 \\
\midrule
\multicolumn{5}{@{}l}{\raisebox{-1pt}{\includegraphics[height=8pt]{figures/logos/qwen.png}}\,\textbf{Qwen~3 / 2.5}} \\
\rowcolor{tablerow} \quad 0.6B  & 67 & 77 & 47 & 77 \\
\quad 1.7B  & 92 & 87 & 60 & \textbf{100} \\
\rowcolor{tablerow} \quad 4B    & 88 & 80 & 70 & \textbf{100} \\
\quad 8B    & \textbf{98} & 90 & \textbf{100}& \textbf{100} \\
\rowcolor{tablerow} \quad 14B   & \textbf{99} & 87 & 63 & \textbf{100} \\
\quad Qwen~2.5 7B  & \textbf{100} & 90 & 73 & \textbf{100} \\
\midrule
\multicolumn{5}{@{}l}{\raisebox{-1pt}{\includegraphics[height=8pt]{figures/logos/llama.png}}\,\textbf{Llama~3.1}} \\
\rowcolor{tablerow} \quad 8B & 81 / 66 & 80 / 80 & \textbf{100} / \textbf{100} & 90 / 50 \\
\bottomrule
\end{tabular}}

\end{table}

\subsubsection{Cosine readout vs.\ the logit lens on base models}
\label{app:lens_compare}

Cosine similarity to tool vectors reads tool labels more accurately than direct output projection at several middle layers (Table~\ref{tab:lens_compare}). We compare both methods on the same cached base-model residuals from BFCL. The output projection, or \emph{logit lens}, applies the final normalization and unembedding to each layer's residual and scores candidate first tokens. Chance accuracy is approximately 29\%, with 4.2 candidate functions per example on average.

\begin{table}[htbp]
\centering
\caption{Tool-label readout accuracy (\%) from the same residual vectors at each layer ($N{=}150$ BFCL examples, leave-one-out tool vectors). Cosine readout generally performs better at middle layers; the methods approach similar accuracy at the final layer. The logit lens applies normalization calibrated for the final layer and scores only candidate first tokens. A lens trained for each depth remains untested.}
\label{tab:lens_compare}

\small
\setlength{\tabcolsep}{5pt}
\begin{tabular}{@{}ll rrr@{}}
\toprule
\textsc{Model} & \textsc{L} & \textsc{Cosine} & \textsc{Lens} & \textsc{$\Delta$} \\
\midrule
\raisebox{-1pt}{\includegraphics[height=8pt]{figures/logos/gemma.png}}\,Gemma~3 4B base & 17 & 51.7 & 34.7 & $+$17.1 \\
\rowcolor{tablerow} \raisebox{-1pt}{\includegraphics[height=8pt]{figures/logos/gemma.png}}\,Gemma~3 4B base & 22 & 42.8 & 32.7 & $+$10.1 \\
\raisebox{-1pt}{\includegraphics[height=8pt]{figures/logos/gemma.png}}\,Gemma~3 4B base & 27 & 44.1 & 45.3 & $-$1.2 \\
\rowcolor{tablerow} \raisebox{-1pt}{\includegraphics[height=8pt]{figures/logos/gemma.png}}\,Gemma~3 4B base & 33 & 53.1 & 49.3 & $+$3.8 \\
\midrule
\raisebox{-1pt}{\includegraphics[height=8pt]{figures/logos/qwen.png}}\,Qwen~3 4B base & 18 & 51.0 & 29.3 & $+$21.7 \\
\rowcolor{tablerow} \raisebox{-1pt}{\includegraphics[height=8pt]{figures/logos/qwen.png}}\,Qwen~3 4B base & 23 & 58.6 & 30.0 & $+$28.6 \\
\raisebox{-1pt}{\includegraphics[height=8pt]{figures/logos/qwen.png}}\,Qwen~3 4B base & 28 & \textbf{61.4} & \textbf{28.0} & $+$\textbf{33.4} \\
\rowcolor{tablerow} \raisebox{-1pt}{\includegraphics[height=8pt]{figures/logos/qwen.png}}\,Qwen~3 4B base & 35 & 62.1 & 62.0 & $+$0.1 \\
\bottomrule
\end{tabular}

\end{table}

\subsubsection{Base-model generation with a prefilled JSON prefix}
\label{app:base_prefill}

Difficulty starting the output format could explain why a base model's tool label is readable while its generated call is incorrect. To test this, we append the JSON prefix \texttt{\{"name": "} after the answer prompt, leaving the model to generate the function name (Table~\ref{tab:base_prefill}).

\begin{table}[htbp]
\centering
\caption{Base-model generation accuracy (\%) on BFCL ($N{=}150$), with a bare answer prompt or prefilled JSON prefix. Prefill does not improve the tested Gemma base models, so failure to start this JSON format does not explain their readout--generation gap. Qwen~3 4B base performs well under both conditions and improves with prefill. Poor base-model generation is therefore specific to particular models: these Gemma base models, and Llama~3.1 8B base in Table~\ref{tab:bfcl_reading}.}
\label{tab:base_prefill}

\small
\begin{tabular}{@{}l rr@{}}
\toprule
\textsc{Model} & \textsc{Bare} & \textsc{Prefilled} \\
\midrule
\raisebox{-1pt}{\includegraphics[height=8pt]{figures/logos/gemma.png}}\,Gemma~3 1B base & 3.3 & 2.0 \\
\raisebox{-1pt}{\includegraphics[height=8pt]{figures/logos/gemma.png}}\,Gemma~3 4B base & 2.0 & 1.3 \\
\raisebox{-1pt}{\includegraphics[height=8pt]{figures/logos/qwen.png}}\,Qwen~3 4B base & 58.0 & 80.0 \\
\bottomrule
\end{tabular}

\end{table}

\section{Steering results and controls}

\subsection{Full pairwise switching matrix (5 tools)}
\label{app:matrix}

\begin{figure}[htbp]
\centering
\includegraphics[width=\linewidth]{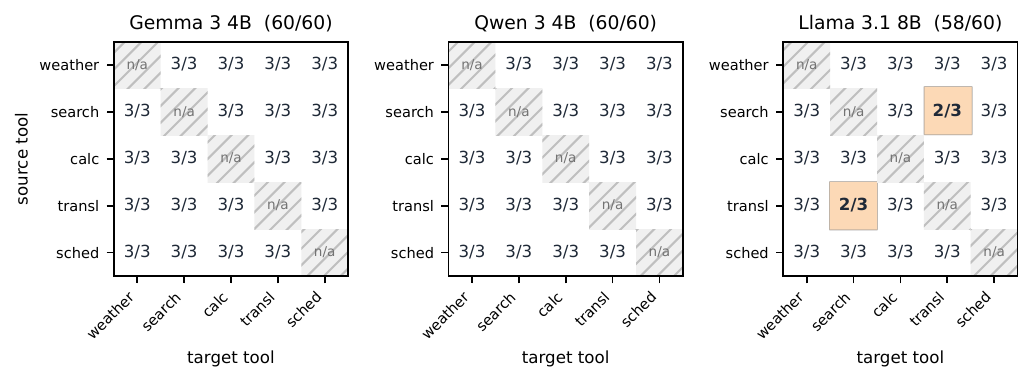}
\caption{5-tool pairwise switching matrix at $\alpha{=}1.0$: exact-match successes out of 3 held-out splits per ordered pair, with vectors built from the other two requests. The two 4B models switch every pair; Llama~3.1 8B has two failed trials.}
\label{fig:matrix}
\end{figure}

Table~\ref{tab:matrix5_all} lists exact-match successes for all 18 models under both $\alpha$ values; per-trial details are in the supplementary archive.

\begin{table}[htbp]
\centering
\caption{5-tool matrix, exact-match successes out of 60 held-out trials per model (20 ordered pairs $\times$ 3 leave-one-out splits; vectors never see the test request).}
\label{tab:matrix5_all}
\footnotesize
\setlength{\tabcolsep}{6pt}
\begin{tabular}{@{}lcc@{}}
\toprule
\textsc{Model} & $\alpha{=}0.7$ & $\alpha{=}1.0$ \\
\midrule
\raisebox{-1pt}{\includegraphics[height=8pt]{figures/logos/gemma.png}}\,Gemma~3 270M IT & 12/60 & 12/60 \\
\rowcolor{tablerow} \raisebox{-1pt}{\includegraphics[height=8pt]{figures/logos/gemma.png}}\,Gemma~3 270M base & 12/60 & 12/60 \\
\raisebox{-1pt}{\includegraphics[height=8pt]{figures/logos/gemma.png}}\,Gemma~3 1B IT & 20/60 & 39/60 \\
\rowcolor{tablerow} \raisebox{-1pt}{\includegraphics[height=8pt]{figures/logos/gemma.png}}\,Gemma~3 1B base & 12/60 & 13/60 \\
\raisebox{-1pt}{\includegraphics[height=8pt]{figures/logos/gemma.png}}\,Gemma~3 4B IT & 50/60 & 60/60 \\
\rowcolor{tablerow} \raisebox{-1pt}{\includegraphics[height=8pt]{figures/logos/gemma.png}}\,Gemma~3 4B base & 35/60 & 46/60 \\
\raisebox{-1pt}{\includegraphics[height=8pt]{figures/logos/gemma.png}}\,Gemma~3 12B IT & 54/60 & 60/60 \\
\rowcolor{tablerow} \raisebox{-1pt}{\includegraphics[height=8pt]{figures/logos/gemma.png}}\,Gemma~3 12B base & 40/60 & 54/60 \\
\raisebox{-1pt}{\includegraphics[height=8pt]{figures/logos/gemma.png}}\,Gemma~3 27B IT & 59/60 & 60/60 \\
\rowcolor{tablerow} \raisebox{-1pt}{\includegraphics[height=8pt]{figures/logos/gemma.png}}\,Gemma~3 27B base & 48/60 & 58/60 \\
\raisebox{-1pt}{\includegraphics[height=8pt]{figures/logos/qwen.png}}\,Qwen~3 0.6B & 39/60 & 47/60 \\
\rowcolor{tablerow} \raisebox{-1pt}{\includegraphics[height=8pt]{figures/logos/qwen.png}}\,Qwen~3 1.7B & 44/60 & 56/60 \\
\raisebox{-1pt}{\includegraphics[height=8pt]{figures/logos/qwen.png}}\,Qwen~3 4B & 53/60 & 60/60 \\
\rowcolor{tablerow} \raisebox{-1pt}{\includegraphics[height=8pt]{figures/logos/qwen.png}}\,Qwen~2.5 7B & 50/60 & 55/60 \\
\raisebox{-1pt}{\includegraphics[height=8pt]{figures/logos/qwen.png}}\,Qwen~3 8B & 58/60 & 60/60 \\
\rowcolor{tablerow} \raisebox{-1pt}{\includegraphics[height=8pt]{figures/logos/qwen.png}}\,Qwen~3 14B & 56/60 & 60/60 \\
\raisebox{-1pt}{\includegraphics[height=8pt]{figures/logos/llama.png}}\,Llama~3.1 8B IT & 55/60 & 58/60 \\
\rowcolor{tablerow} \raisebox{-1pt}{\includegraphics[height=8pt]{figures/logos/llama.png}}\,Llama~3.1 8B base & 34/60 & 45/60 \\
\bottomrule
\end{tabular}

\end{table}

\subsection{Strength and number of tools}
\label{app:strength_menu}

\subsubsection{Scaling results}
\label{app:scaling}
Switching remains high as the number of sampled tools grows in this exploratory experiment (Table~\ref{tab:scaling_v2}). The experiment uses prefix scoring and reuses the test requests in the tool vectors, so the evaluation is looser than in the main text.

\begin{table}[htbp]
\centering
\caption{Switch accuracy (\%) as the number of tools $K$ increases, among randomly sampled tools. Each tool vector comes from one templated request and is tested on a second, differently worded templated request. Left: the first-token prefix rule, $N{=}20$ pairs per cell. Right: the full target name on 60 pairs per cell ($K{=}5$: all 20), with the edit applied on the first forward pass and eight generated tokens, rerun on the three representative models. G=Gemma, Q=Qwen, Q2.5=Qwen~2.5, L=Llama. The $K{=}15$ row uses the Sw-15 tool list of Table~\ref{tab:generalization} but a less restrictive evaluation: Gemma~3 4B scores 100 here and 83 with held-out requests, both under prefix scoring.}
\label{tab:scaling_v2}

\small
\setlength{\tabcolsep}{7pt}
\rowcolors{2}{tablerow}{white}
\resizebox{\ifdim\width>\columnwidth\columnwidth\else\width\fi}{!}{\begin{tabular}{rcccccccc|ccc}
\toprule
& \multicolumn{8}{c|}{\textsc{Prefix rule, 20 pairs}} & \multicolumn{3}{c}{\textsc{Full name, 60 pairs}} \\
\textsc{$K$} & \textsc{G-4B} & \textsc{G-12B} & \textsc{G-27B} & \textsc{Q-4B} & \textsc{Q-8B} & \textsc{Q-14B} & \textsc{Q2.5-7B} & \textsc{L-8B} & \textsc{G-4B} & \textsc{Q-4B} & \textsc{L-8B} \\
\midrule
5 & 100 & 100 & 100 & 100 & 75 & 100 & 100 & 100 & 100 & 100 & 100 \\
15 & 100 & 100 & 100 & 100 & 100 & 100 & 95 & 95 & 67 & 75 & 63 \\
30 & 100 & 100 & 100 & 100 & 95 & 100 & 90 & 100 & 67 & 70 & 62 \\
50 & 95 & 100 & 95 & 100 & 100 & 100 & 100 & 100 & 62 & 60 & 58 \\
75 & 95 & 100 & 95 & 100 & 100 & 100 & 100 & 100 & 38 & 47 & 40 \\
100 & 100 & 100 & 100 & 100 & 100 & 95 & 95 & 100 & 58 & 47 & 58 \\
150 & 90 & 100 & 100 & 100 & 100 & 95 & 100 & 100 & 20 & 23 & 25 \\
\bottomrule
\end{tabular}}
\rowcolors{1}{}{}

\end{table}

\subsubsection{Steering strength and tool-count scaling}
\label{app:strength_scaling_figs}

\begin{figure}[htbp]
\centering
\includegraphics[width=\linewidth]{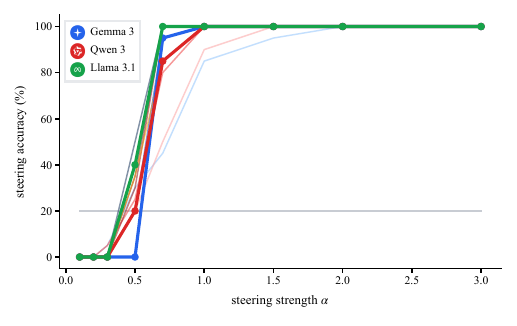}
\caption{Steering accuracy against strength $\alpha$ (20 pairs per point). The eight instruction-tuned models of 4B and above reach high accuracy between $\alpha{=}0.5$ and $1.0$, without a decline at higher tested strengths. Pale lines show smaller models: most need stronger interventions, while Gemma~3 270M remains unchanged. Table~\ref{tab:strength} includes the sampled values, three base models, and the transition estimate.}
\label{fig:strength}
\end{figure}

\begin{figure}[htbp]
\centering
\includegraphics[width=\linewidth]{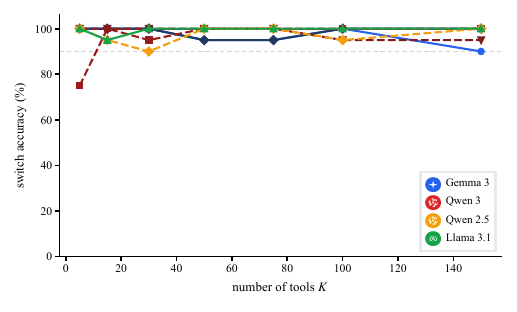}
\caption{Steering accuracy against the number of tools $K$ (20 pairs per point). Accuracy remains high from $K{=}15$ to $150$, without a consistent ordering by model size. Qwen~3 8B at $K{=}5$ is the exception to the otherwise high scores. Full values and evaluation restrictions are in Table~\ref{tab:scaling_v2}.}
\label{fig:scaling_v2}
\end{figure}

Steering accuracy rises sharply over a narrow range of strengths (Figure~\ref{fig:strength}). With strength fixed, larger sampled tool lists do not consistently reduce accuracy in this prefix-scored experiment (Figure~\ref{fig:scaling_v2}).

\subsubsection{Steering strength calibration}
\label{app:strength}

Table~\ref{tab:strength} shows how the same intervention changes as its strength increases, including where a stronger intervention stops helping.

\begin{table}[htbp]
\centering
\caption{Steering strength calibration. ``Trans.'' is the first sampled $\alpha$ with accuracy above 50\%. Linear interpolation between the two bracketing samples gives crossings at 0.50--0.61 for the 4B+ instruction-tuned models; these are estimates from a coarse grid. The base models either remain weak or lose accuracy at high strength.}
\label{tab:strength}

\small
\setlength{\tabcolsep}{7pt}
\rowcolors{2}{tablerow}{white}
\resizebox{\ifdim\width>\columnwidth\columnwidth\else\width\fi}{!}{\begin{tabular}{lccccccc}
\toprule
\textsc{Model} & $\alpha{=}0.3$ & $\alpha{=}0.5$ & $\alpha{=}0.7$ & $\alpha{=}1.0$ & $\alpha{=}2.0$ & $\alpha{=}3.0$ & \textsc{Trans.} \\
\midrule
\raisebox{-1pt}{\includegraphics[height=8pt]{figures/logos/gemma.png}}\,Gemma~3 270M & 20\% & 20\% & 20\% & 20\% & 20\% & 20\% & --- \\
\raisebox{-1pt}{\includegraphics[height=8pt]{figures/logos/gemma.png}}\,Gemma~3 1B & 5\% & 30\% & 45\% & 85\% & 100\% & 100\% & $\sim$1.0 \\
\raisebox{-1pt}{\includegraphics[height=8pt]{figures/logos/gemma.png}}\,Gemma~3 4B & 0\% & 0\% & 95\% & 100\% & 100\% & 100\% & 0.7 \\
\raisebox{-1pt}{\includegraphics[height=8pt]{figures/logos/gemma.png}}\,Gemma~3 12B & 0\% & 20\% & 95\% & 100\% & 100\% & 100\% & 0.7 \\
\raisebox{-1pt}{\includegraphics[height=8pt]{figures/logos/gemma.png}}\,Gemma~3 27B & 0\% & 50\% & 100\% & 100\% & 100\% & 100\% & 0.7 \\
\raisebox{-1pt}{\includegraphics[height=8pt]{figures/logos/qwen.png}}\,Qwen~3 0.6B & 0\% & 20\% & 50\% & 90\% & 100\% & 100\% & $\sim$1.0 \\
\raisebox{-1pt}{\includegraphics[height=8pt]{figures/logos/qwen.png}}\,Qwen~3 1.7B & 5\% & 25\% & 80\% & 100\% & 100\% & 100\% & 0.7 \\
\raisebox{-1pt}{\includegraphics[height=8pt]{figures/logos/qwen.png}}\,Qwen~3 4B & 0\% & 20\% & 85\% & 100\% & 100\% & 100\% & 0.7 \\
\raisebox{-1pt}{\includegraphics[height=8pt]{figures/logos/qwen.png}}\,Qwen~3 8B & 0\% & 30\% & 100\% & 100\% & 100\% & 100\% & 0.7 \\
\raisebox{-1pt}{\includegraphics[height=8pt]{figures/logos/qwen.png}}\,Qwen~3 14B & 0\% & 35\% & 100\% & 100\% & 100\% & 100\% & 0.7 \\
\raisebox{-1pt}{\includegraphics[height=8pt]{figures/logos/qwen.png}}\,Qwen~2.5 7B & 0\% & 35\% & 85\% & 100\% & 100\% & 100\% & 0.7 \\
\raisebox{-1pt}{\includegraphics[height=8pt]{figures/logos/llama.png}}\,Llama~3.1 8B & 0\% & 40\% & 100\% & 100\% & 100\% & 100\% & 0.7 \\
\midrule
\multicolumn{8}{@{}l}{\emph{Base (pre-trained) models:}} \\
\raisebox{-1pt}{\includegraphics[height=8pt]{figures/logos/gemma.png}}\,Gemma~3 270M-base & 20\% & 20\% & 20\% & 20\% & 25\% & 35\% & --- \\
\raisebox{-1pt}{\includegraphics[height=8pt]{figures/logos/gemma.png}}\,Gemma~3 1B-base & 20\% & 20\% & 20\% & 20\% & 40\% & 45\% & --- \\
\raisebox{-1pt}{\includegraphics[height=8pt]{figures/logos/gemma.png}}\,Gemma~3 4B-base & 5\% & 30\% & 55\% & 80\% & 75\% & 65\% & $\sim$0.7 \\
\bottomrule
\end{tabular}}
\rowcolors{1}{}{}

\end{table}

\subsection{The unsteered control on the held-out requests}
\label{app:unsteered_control}

The unsteered column of Table~\ref{tab:generalization} is measured on the same held-out requests and ordered pairs as the steered column. We use the steered run's prompts, tool list and decoding with no intervention, then score the result against all 210 ordered pairs. The continuation depends only on the source request, so one generation serves fourteen targets and fifteen generations cover the table.

We separate two rates. The first is how often the unsteered continuation already names the
pair's target, which is what a steered number has to beat. The second is how often it names the tool
the request actually asks for, which measures how often the model was already doing the right thing before the intervention. Full names are counted with the rule the steered column uses, not the archives'
five-character field.

\begin{table}[htbp]
\centering
\caption{Full-name counts over the same 210 ordered pairs. \textsc{Steered}: target hits under
steering, the column Table~\ref{tab:generalization} prints. \textsc{Steered, source correct}: the same
hits counted only over pairs whose unsteered continuation names the source tool, so the denominator
changes with the model. \textsc{Unsteered target}: target hits with no intervention.
\textsc{Unsteered source}: continuations that name the tool their own request asks for. Both Gemma~3 270M rows sit at the $1/15$ chance rate on both unsteered counts, which is what a
model that cannot do the task should give, and the paper reports no effective steering for them.}
\label{tab:unsteered_control}

\footnotesize
\setlength{\tabcolsep}{6pt}
\resizebox{\ifdim\width>\columnwidth\columnwidth\else\width\fi}{!}{\begin{tabular}{@{}lrrrr@{}}
\toprule
Model & Steered & Steered, source correct & Unsteered target & Unsteered source \\
\midrule
\raisebox{-1pt}{\includegraphics[height=8pt]{figures/logos/gemma.png}}\,Gemma~3 270M IT & 14/210 & 0/14 & 14/210 & 14/210 \\
\raisebox{-1pt}{\includegraphics[height=8pt]{figures/logos/gemma.png}}\,Gemma~3 270M base & 14/210 & 0/14 & 14/210 & 14/210 \\
\raisebox{-1pt}{\includegraphics[height=8pt]{figures/logos/gemma.png}}\,Gemma~3 1B IT & 66/210 & 36/112 & 7/210 & 112/210 \\
\raisebox{-1pt}{\includegraphics[height=8pt]{figures/logos/gemma.png}}\,Gemma~3 4B IT & 118/210 & 113/196 & 1/210 & 196/210 \\
\raisebox{-1pt}{\includegraphics[height=8pt]{figures/logos/gemma.png}}\,Gemma~3 4B base & 57/210 & 37/140 & 5/210 & 140/210 \\
\raisebox{-1pt}{\includegraphics[height=8pt]{figures/logos/gemma.png}}\,Gemma~3 12B IT & 154/210 & 154/210 & 0/210 & 210/210 \\
\raisebox{-1pt}{\includegraphics[height=8pt]{figures/logos/gemma.png}}\,Gemma~3 27B IT & 163/210 & 163/210 & 0/210 & 210/210 \\
\raisebox{-1pt}{\includegraphics[height=8pt]{figures/logos/gemma.png}}\,Gemma~3 27B base & 130/210 & 122/196 & 1/210 & 196/210 \\
\raisebox{-1pt}{\includegraphics[height=8pt]{figures/logos/qwen.png}}\,Qwen~3 0.6B & 94/210 & 67/140 & 5/210 & 140/210 \\
\raisebox{-1pt}{\includegraphics[height=8pt]{figures/logos/qwen.png}}\,Qwen~3 1.7B & 144/210 & 124/182 & 2/210 & 182/210 \\
\raisebox{-1pt}{\includegraphics[height=8pt]{figures/logos/qwen.png}}\,Qwen~3 4B & 149/210 & 119/168 & 3/210 & 168/210 \\
\raisebox{-1pt}{\includegraphics[height=8pt]{figures/logos/qwen.png}}\,Qwen~3 8B & 161/210 & 150/196 & 0/210 & 196/210 \\
\raisebox{-1pt}{\includegraphics[height=8pt]{figures/logos/qwen.png}}\,Qwen~3 14B & 163/210 & 163/210 & 0/210 & 210/210 \\
\raisebox{-1pt}{\includegraphics[height=8pt]{figures/logos/qwen.png}}\,Qwen~2.5 7B & 153/210 & 132/182 & 2/210 & 182/210 \\
\raisebox{-1pt}{\includegraphics[height=8pt]{figures/logos/llama.png}}\,Llama~3.1 8B IT & 126/210 & 126/210 & 0/210 & 210/210 \\
\raisebox{-1pt}{\includegraphics[height=8pt]{figures/logos/llama.png}}\,Llama~3.1 8B base & 91/210 & 83/182 & 2/210 & 182/210 \\
\bottomrule
\end{tabular}}

\end{table}

Restricting the steered count to pairs whose unsteered continuation names the source tool moves it very little, from 56--78\% to 58--78\% across the models of 4B and above. So the headline rate does not depend on pairs where the model was already off task.

\subsection{Full confidence intervals}
\label{app:full_ci}

Each of the 15 requests and each of the 15 target tools appears in 14 of the 210 pairs, so the pairs are not independent. A bootstrap that resamples whole requests instead of single pairs barely changes the intervals (Table~\ref{tab:cluster_ci}). Resampling the target tools as well widens them to about $\pm$20 points. The reason is that the four tools whose names start with \texttt{get} share a first token, and steering reaches them by full name far less often than the other eleven (Appendix~\ref{app:failure}).

\begin{table}[htbp]
\centering
\caption{95\% intervals for the full-name \textsc{Sw-15} rate of Table~\ref{tab:generalization}. \textsc{Wilson} treats the 210 ordered pairs as independent. \textsc{Requests} resamples the 15 source requests with replacement; \textsc{Requests and tools} also resamples the target tools ($B{=}10{,}000$ each).}
\label{tab:cluster_ci}

\small
\resizebox{\ifdim\width>\columnwidth\columnwidth\else\width\fi}{!}{%
\begin{tabular}{@{}l r c c c@{}}
\toprule
\textsc{Model} & \textsc{Sw-15} & \textsc{Wilson} & \textsc{Requests} & \textsc{Requests and tools} \\
\midrule
Gemma~3 4B & 56 & [49,63] & [49,63] & [35,77] \\
Gemma~3 12B & 73 & [67,79] & [71,76] & [53,91] \\
Gemma~3 27B & 78 & [72,83] & [76,79] & [57,96] \\
Qwen~3 4B & 71 & [64,77] & [70,72] & [50,90] \\
Qwen~3 8B & 77 & [70,82] & [74,79] & [57,93] \\
Qwen~3 14B & 78 & [72,83] & [76,79] & [57,95] \\
Qwen~2.5 7B & 73 & [66,78] & [70,75] & [52,91] \\
Llama~3.1 8B & 60 & [53,66] & [55,65] & [40,79] \\
\bottomrule
\end{tabular}}

\end{table}

Table~\ref{tab:full_ci} reports confidence intervals for the runs that mix scoring rules. Its caption specifies the scoring rules and sample sizes; these real-API entries differ from the complete-name scores in Table~\ref{tab:generalization}.

\begin{table}[htbp]
\centering
\caption{Wilson 95\% CIs for the benchmark runs that mix scoring rules. Apart from Sw-15 full, the columns use first-token prefix scoring; the real-API entries use prefix scoring, unlike the full-name scores in Table~\ref{tab:generalization}. $N$ is 210 in the two Sw-15 columns (all ordered pairs, test request held out) and 30 elsewhere, except that Gemma~3 4B is 47/50 on $\tau$-air and 38/50 on $\tau$-ret. Format: accuracy [\%] [lower, upper].}
\label{tab:full_ci}

\footnotesize
\setlength{\tabcolsep}{6pt}
\rowcolors{2}{tablerow}{white}
\resizebox{\ifdim\width>\columnwidth\columnwidth\else\width\fi}{!}{\begin{tabular}{@{}l ccccc@{}}
\toprule
\rowcolor{white}
\textsc{Model} & \textsc{Sw-15 full} & \textsc{Sw-15 prefix} & \textsc{$\tau$-Air} & \textsc{$\tau$-Ret} & \textsc{TB-XD prefix} \\
\midrule
\raisebox{-1pt}{\includegraphics[height=8pt]{figures/logos/gemma.png}}\,Gemma~3 270M & 7 [4,11] & 27 [21,33] & 10 [3,26] & 0 [0,11] & 30 [17,48] \\
\raisebox{-1pt}{\includegraphics[height=8pt]{figures/logos/gemma.png}}\,Gemma~3 1B   & 31 [26,38] & 54 [48,61] & 77 [59,88] & 53 [36,70] & 80 [63,90] \\
\raisebox{-1pt}{\includegraphics[height=8pt]{figures/logos/gemma.png}}\,Gemma~3 4B   & 56 [49,63] & 83 [78,88] & 94 [84,98] & 76 [63,86] & 100 [89,100] \\
\raisebox{-1pt}{\includegraphics[height=8pt]{figures/logos/gemma.png}}\,Gemma~3 12B  & 73 [67,79] & 100 [98,100] & 90 [74,97] & 80 [63,90] & 100 [89,100] \\
\raisebox{-1pt}{\includegraphics[height=8pt]{figures/logos/gemma.png}}\,Gemma~3 27B  & 78 [72,83] & 100 [98,100] & 77 [59,88] & 80 [63,90] & 100 [89,100] \\
\midrule
\raisebox{-1pt}{\includegraphics[height=8pt]{figures/logos/qwen.png}}\,Qwen~3 0.6B  & 45 [39,52] & 67 [61,73] & 77 [59,88] & 47 [30,64] & 77 [59,88] \\
\raisebox{-1pt}{\includegraphics[height=8pt]{figures/logos/qwen.png}}\,Qwen~3 1.7B  & 69 [62,75] & 92 [87,95] & 87 [70,95] & 60 [42,75] & 100 [89,100] \\
\raisebox{-1pt}{\includegraphics[height=8pt]{figures/logos/qwen.png}}\,Qwen~3 4B    & 71 [64,77] & 88 [83,92] & 80 [63,90] & 70 [52,83] & 100 [89,100] \\
\raisebox{-1pt}{\includegraphics[height=8pt]{figures/logos/qwen.png}}\,Qwen~3 8B    & 77 [70,82] & 98 [95,99] & 90 [74,97] & 100 [89,100] & 100 [89,100] \\
\raisebox{-1pt}{\includegraphics[height=8pt]{figures/logos/qwen.png}}\,Qwen~3 14B   & 78 [72,83] & 99 [97,100] & 87 [70,95] & 63 [46,78] & 100 [89,100] \\
\midrule
\raisebox{-1pt}{\includegraphics[height=8pt]{figures/logos/llama.png}}\,Llama~3.1 8B & 60 [53,66] & 81 [75,86] & 80 [63,90] & 100 [89,100] & 90 [74,97] \\
\raisebox{-1pt}{\includegraphics[height=8pt]{figures/logos/qwen.png}}\,Qwen~2.5 7B  & 73 [66,78] & 100 [97,100] & 90 [74,97] & 73 [56,86] & 100 [89,100] \\
\bottomrule
\end{tabular}}
\rowcolors{1}{}{}

\end{table}

\subsection{Cross-benchmark and ToolBench analysis}
\label{app:benchmark_heatmap}
\label{app:toolbench_domains}

\begin{figure}[htbp]
\centering
\begin{minipage}[t]{\linewidth}
\centering
\includegraphics[width=\linewidth]{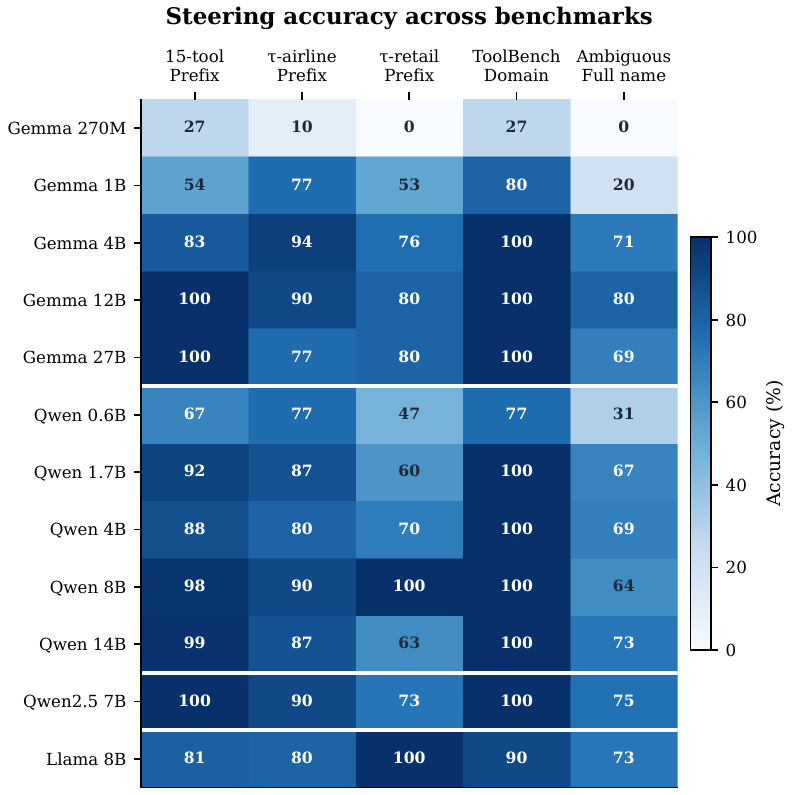}
\caption{Steering accuracy across 12 models $\times$ 5 benchmarks.}
\label{fig:benchmark_heatmap}
\end{minipage}

\begin{minipage}[t]{\linewidth}
\centering
\includegraphics[width=\linewidth]{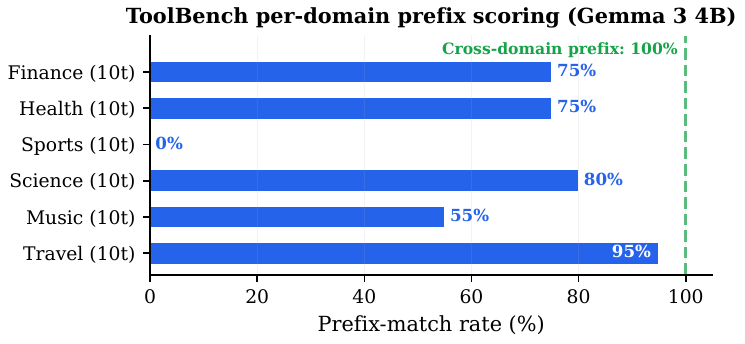}
\caption{ToolBench prefix-matching accuracy by domain (Gemma~3 4B). The dashed line is an earlier cross-domain run under prefix scoring, not exact API selection (Appendix~\ref{app:realapi_scoring}). Steering fails on the near-synonymous Sports tools.}
\label{fig:toolbench_domains}
\end{minipage}
\end{figure}

Figure~\ref{fig:benchmark_heatmap} shows steering accuracy across 12 models and 5 benchmarks; Figure~\ref{fig:toolbench_domains} shows ToolBench per-domain breakdown on Gemma~3 4B.

\subsection{Failure mode analysis}
\label{app:failure}
Steering failures fall into three categories.

\emph{Shared name prefixes.} Four tools (\texttt{get\_weather}, \texttt{get\_stock\_price}, \texttt{get\_news}, \texttt{get\_directions}) share the first token \texttt{get}. Steering can produce that token without producing the correct continuation. Across the eight instruction-tuned models of 4B and above, full-name accuracy is 11--20\% on \texttt{get\_*} targets, compared with 71--99\% on unique-prefix targets. Prefix scoring therefore overstates successful selection among these tools.

\emph{Specific requests.} The model can retain the original tool when the request strongly specifies it. ``Calculate 15 * 37'' resists steering away from \texttt{calculate}, for example.

\emph{Near-synonyms.} Steering fails on the 10 similar Sports APIs. Across the 90 stored steered generations, all 10 failures name a tool on the tool list rather than inventing one. We do not report a comparable malformed-output count, because generation stops at a fixed token budget and can end mid-JSON for reasons unrelated to steering.

\subsection{Full generation examples}
\label{app:generations}

The \textsc{Ambiguous} column in Table~\ref{tab:generalization} uses full-name scoring. Prefix scoring would accept some unsteered defaults because four tools begin with \texttt{get\_}; it would therefore credit steering even when the selected tool has not changed. Table~\ref{tab:ambiguous_full} shows the distinction for every Gemma~3 4B trial.

The examples in Table~\ref{tab:generations} were generated autoregressively from the literal prefix \texttt{\{\{"tool": "}, with the steering direction applied only on the first forward pass. The table shows argument excerpts; Table~\ref{tab:ambiguous_full} gives the ambiguous-request variant.

\begin{table}[htbp]
\centering
\caption{Stored steered generations. For each model, we select pairs where both the tool name and argument keys changed, order them by source tool, and show the first four. The same pairs qualify on all three models. Argument excerpts preserve the recorded text, including nested braces, but omit the outer call wrapper. Ellipses mark omitted continuation or shortening at a key boundary; they are not generated tokens. The Llama ticker \texttt{ALRM} is model-generated. This selection illustrates changed outputs, not a random sample of steering trials.}
\label{tab:generations}

\scriptsize
\setlength{\tabcolsep}{6pt}
\rowcolors{2}{tablerow}{white}
\resizebox{\ifdim\width>\columnwidth\columnwidth\else\width\fi}{!}{\begin{tabular}{ c l p{0.17\linewidth} p{0.22\linewidth} p{0.34\linewidth} }
\toprule
\textsc{Model} & \textsc{Direction} & \textsc{Request} & \textsc{Original argument excerpts} & \textsc{Steered argument excerpts} \\
\midrule
\raisebox{-1pt}{\includegraphics[height=7pt]{figures/logos/gemma.png}} & \texttt{alarm}$\to$\texttt{directions} & ``Set alarm for 7am.'' & \texttt{"time": "7am"} & \texttt{\textcolor{steered}{"from": "current location", "to": "nearest hospital"}} \\
\raisebox{-1pt}{\includegraphics[height=7pt]{figures/logos/gemma.png}} & \texttt{alarm}$\to$\texttt{stock} & ``Set alarm for 7am.'' & \texttt{"time": "7am"} & \texttt{\textcolor{steered}{"ticker": "AAPL"}} \\
\raisebox{-1pt}{\includegraphics[height=7pt]{figures/logos/gemma.png}} & \texttt{calculate}$\to$\texttt{translate} & ``Calculate 15 * 37.'' & \texttt{"expr": "15 * 37"} & \texttt{\textcolor{steered}{"text": "15 * 37", "lang": "en", "result": "555"}} \\
\raisebox{-1pt}{\includegraphics[height=7pt]{figures/logos/gemma.png}} & \texttt{currency}$\to$\texttt{alarm} & ``Convert 100 USD to EUR.'' & \texttt{"amount": 100, "from\_cur": "USD", "to\_cur": "EUR"} & \texttt{\textcolor{steered}{"time": "10:00 AM"\,\ldots}} \\
\cmidrule(lr){1-5}
\raisebox{-1pt}{\includegraphics[height=7pt]{figures/logos/qwen.png}} & \texttt{alarm}$\to$\texttt{directions} & ``Set alarm for 7am.'' & \texttt{"time": "07:00"\,\ldots} & \texttt{\textcolor{steered}{"from": "home", "to": "work"\,\ldots}} \\
\raisebox{-1pt}{\includegraphics[height=7pt]{figures/logos/qwen.png}} & \texttt{alarm}$\to$\texttt{stock} & ``Set alarm for 7am.'' & \texttt{"time": "07:00"\,\ldots} & \texttt{\textcolor{steered}{"ticker": "AAPL"\,\ldots}} \\
\raisebox{-1pt}{\includegraphics[height=7pt]{figures/logos/qwen.png}} & \texttt{calculate}$\to$\texttt{translate} & ``Calculate 15 * 37.'' & \texttt{"params": \{"expr": "15 * 37"\}\,\ldots} & \texttt{\textcolor{steered}{"text": "Calculate 15 * 37.", "lang": "en"\,\ldots}} \\
\raisebox{-1pt}{\includegraphics[height=7pt]{figures/logos/qwen.png}} & \texttt{currency}$\to$\texttt{alarm} & ``Convert 100 USD to EUR.'' & \texttt{"params": \{"expr": "100 * 0.93"\}\,\ldots} & \texttt{\textcolor{steered}{"time": "08:00", "message": "Convert 100 USD to EUR."\,\ldots}} \\
\cmidrule(lr){1-5}
\raisebox{-1pt}{\includegraphics[height=7pt]{figures/logos/llama.png}} & \texttt{alarm}$\to$\texttt{directions} & ``Set alarm for 7am.'' & \texttt{"args": \{"time": "7am"\}} & \texttt{\textcolor{steered}{"from": "user's current location", "to": "user's home"}} \\
\raisebox{-1pt}{\includegraphics[height=7pt]{figures/logos/llama.png}} & \texttt{alarm}$\to$\texttt{stock} & ``Set alarm for 7am.'' & \texttt{"args": \{"time": "7am"\}} & \texttt{\textcolor{steered}{"ticker": "ALRM"}} \\
\raisebox{-1pt}{\includegraphics[height=7pt]{figures/logos/llama.png}} & \texttt{calculate}$\to$\texttt{translate} & ``Calculate 15 * 37.'' & \texttt{"expr": "15 * 37"} & \texttt{\textcolor{steered}{"text": "15 * 37", "lang": "en"}} \\
\raisebox{-1pt}{\includegraphics[height=7pt]{figures/logos/llama.png}} & \texttt{currency}$\to$\texttt{alarm} & ``Convert 100 USD to EUR.'' & \texttt{"amount": 100, "from\_cur": "USD", "to\_cur": "EUR"} & \texttt{\textcolor{steered}{"time": "now"\,\ldots}} \\
\bottomrule
\end{tabular}}
\rowcolors{1}{}{}

\end{table}

\begin{table}[htbp]
\centering
\caption{All 18 ambiguous-request trials on Gemma~3 4B, with the tool actually generated; $\times$ marks a failed target switch. Prefix containment scores this set at 78\%, compared with 67\% under full-name scoring. In 22\% of trials, the unsteered default already satisfies the prefix rule. For example, steering the dinner request toward \texttt{get\_directions} produces \texttt{get\_weather}, which shares the prefix but is the wrong tool.}
\label{tab:ambiguous_full}

\footnotesize
\setlength{\tabcolsep}{7pt}
\rowcolors{2}{tablerow}{white}
\resizebox{\ifdim\width>\columnwidth\columnwidth\else\width\fi}{!}{\begin{tabular}{@{} l l l @{}}
\toprule
\textsc{Request} & \textsc{Steered To} & \textsc{Tool Named} \\
\midrule
``Help me plan my trip to Paris'' & \texttt{book\_flight} & \texttt{\textcolor{steered}{book\_flight}} \\
 & \texttt{get\_directions} & \texttt{book\_flight\,$\times$} \\
 & \texttt{search} & \texttt{\textcolor{steered}{search}} \\
\midrule
``I need to prepare for my meeting …'' & \texttt{schedule} & \texttt{\textcolor{steered}{schedule}} \\
 & \texttt{create\_note} & \texttt{\textcolor{steered}{create\_note}} \\
 & \texttt{send\_email} & \texttt{create\_note\,$\times$} \\
\midrule
``What should I know about Tokyo?'' & \texttt{search} & \texttt{\textcolor{steered}{search}} \\
 & \texttt{get\_news} & \texttt{search\,$\times$} \\
 & \texttt{translate} & \texttt{\textcolor{steered}{translate}} \\
\midrule
``Help me with dinner tonight'' & \texttt{search} & \texttt{\textcolor{steered}{search}} \\
 & \texttt{get\_directions} & \texttt{get\_weather\,$\times$} \\
\midrule
``How much will my vacation cost?'' & \texttt{calculate} & \texttt{\textcolor{steered}{calculate}} \\
 & \texttt{convert\_currency} & \texttt{\textcolor{steered}{convert\_currency}} \\
 & \texttt{book\_flight} & \texttt{\textcolor{steered}{book\_flight}} \\
 & \texttt{search} & \texttt{\textcolor{steered}{search}} \\
\midrule
``Keep track of Apple for me'' & \texttt{get\_stock\_price} & \texttt{get\_news\,$\times$} \\
 & \texttt{create\_note} & \texttt{\textcolor{steered}{create\_note}} \\
 & \texttt{get\_news} & \texttt{search\,$\times$} \\
\bottomrule
\end{tabular}}
\rowcolors{1}{}{}

\end{table}

\subsection{Schema-correctness controls}
\label{app:schema_ar}

\textbf{Argument generation after the tool name.}
Writing the target tool name into the prompt (prefilling) gives the same schema correctness as steering (Table~\ref{tab:schema_ar}). Gemma~3 4B and Qwen~3 4B differ by at most one trial between the two conditions. On Llama~3.1 8B, the difference disappears when analysis is restricted to the 23 pairs where steering switches the name. So the arguments adapt after the tool name is chosen, whichever way the name gets there. In the fourth condition, we keep each tool name and replace its argument keys with positional labels such as \texttt{p1} and \texttt{p2}. Conformity to these new keys falls on all three models, suggesting that familiar key names help argument generation.

\textbf{Biasing the output token.} On the same 30 sampled pairs (seed 42), we add a constant to the target tool's first-token logit at the first generation step, then generate four tokens normally. Small biases switch few calls; even a bias of 20 is less effective than activation steering on Gemma~3 4B and Qwen~3 4B. The success rates for Gemma~3 4B / Qwen~3 4B / Llama~3.1 8B are: bias 2, 0/0/0\%; bias 5, 0/0/6.7\%; bias 10, 3.3/0/33.3\%; and bias 20, 23.3/66.7/66.7\%. Activation steering under the same five-character scoring rule reaches 80\% on Gemma and 93\% on Qwen. These scores use name-prefix containment, not full-name matching. Per-trial outputs are in the supplementary archive.

\begin{table}[htbp]
\centering
\caption{Schema correctness on the same 30 source$\to$target pairs. A: no intervention, evaluated against the source schema. B: target-name prefill followed by argument generation. C: activation steering instead of prefill. D: target-name prefill with parameter keys replaced by positional labels (\texttt{p1}, \texttt{p2}, etc.). B and C are similar; their Llama~3.1 8B difference disappears on the 23 pairs where steering switches the name. D is evaluated against these replacement keys.}
\label{tab:schema_ar}

\small
\setlength{\tabcolsep}{6pt}
\resizebox{\ifdim\width>\columnwidth\columnwidth\else\width\fi}{!}{\begin{tabular}{@{}l cccc@{}}
\toprule
 & \textsc{A: base} & \textsc{B: prefill} & \textsc{C: steer} & \textsc{D: swapped} \\
 & src schema & tgt schema & tgt schema & tgt schema \\
\midrule
\raisebox{-1pt}{\includegraphics[height=8pt]{figures/logos/gemma.png}}\,Gemma~3 4B IT     & 87\% & 70\% & 70\% & 10\% \\
\raisebox{-1pt}{\includegraphics[height=8pt]{figures/logos/qwen.png}}\,Qwen~3 4B         & 60\% & 27\% & 23\% & 0\% \\
\raisebox{-1pt}{\includegraphics[height=8pt]{figures/logos/llama.png}}\,Llama~3.1 8B IT   & 90\% & 53\% & 43\% & 30\% \\
\bottomrule
\end{tabular}}

\end{table}

\subsection{Robustness of the steering directions}
\label{app:robustness}

\textbf{Example-request robustness.} Ambiguous and out-of-distribution example requests reduce steering more than readout accuracy (Table~\ref{tab:seed_robustness}). We rebuild the tool vectors from 2--3 requests per tool in each of five alternative sets of example requests, then test Gemma~3 4B, Qwen~3 4B and Llama~3.1 8B with 30 trials per cell. The sets contain paraphrases, indirect requests, changed entities, ambiguous requests, and abbreviated requests unlike the examples in the tool list. Examples include replacing the original cities with Reykjavik and Kuala Lumpur, ambiguous requests such as ``Should I bring a jacket to Oslo?'' and ``Put something on.'', and abbreviations such as ``tokyo weather now'' and ``AAPL px''. Paraphrases and changed entities have little effect; ambiguous requests produce the lowest steering accuracy. So the readout tolerates variation in the example requests that weakens steering.

\begin{table}[htbp]
\centering
\caption{Example-request robustness on the 15-tool list, 30 trials per cell. Read: top-1 cosine readout; steer: successful tool switches after rebuilding the tool vectors from each set. Ambiguous requests reduce steering most on each model, tied with indirect requests on Llama~3.1 8B. Out-of-distribution requests also reduce steering while largely preserving readout. Llama starts from lower steering accuracy under the original requests.}
\label{tab:seed_robustness}
\small
\setlength{\tabcolsep}{7pt}
\resizebox{\ifdim\width>\columnwidth\columnwidth\else\width\fi}{!}{%
\begin{tabular}{@{}l cc cc cc@{}}
\toprule
& \multicolumn{2}{c}{Gemma~3 4B} & \multicolumn{2}{c}{Qwen~3 4B} & \multicolumn{2}{c}{Llama~3.1 8B} \\
\textsc{Seed set} & read & steer & read & steer & read & steer \\
\midrule
\rowcolor{tablerow} Original & 30/30 & 28/30 & 30/30 & 29/30 & 30/30 & 22/30 \\
Paraphrased & 29/30 & 30/30 & 30/30 & 29/30 & 30/30 & 22/30 \\
\rowcolor{tablerow} Atypical (indirect) & 26/30 & 21/30 & 28/30 & 25/30 & 29/30 & 17/30 \\
Entity-varied & 30/30 & 28/30 & 30/30 & 29/30 & 30/30 & 20/30 \\
\rowcolor{tablerow} Ambiguous & 26/30 & 12/30 & 25/30 & 15/30 & 26/30 & 17/30 \\
Out-of-distribution & 30/30 & 19/30 & 30/30 & 24/30 & 28/30 & 18/30 \\
\bottomrule
\end{tabular}}

\end{table}

\textbf{Number of example requests.} Tool vectors can also be built from more than two requests per tool (Table~\ref{tab:seedcount}). We pool eight requests per tool, the two used everywhere else plus two paraphrased, two with changed entities and two indirect ones, and draw one, two or four of them at random three times. With one request per tool, the last-layer count changes by up to 56 pairs from one draw to the next. With four requests the draws agree more closely. On the original tool list, the two requests used in the main experiments do better than every random draw and than all eight together, which suggests that their typical wording matters more than the number of requests.

\begin{table}[htbp]
\centering
\caption{Held-out full-name switches out of 210 ordered pairs at the last layer when each tool vector averages $n$ requests drawn at random from a pool of eight per tool. Mean over three draws with the range in parentheses; $n{=}8$ uses the whole pool. The main experiments use the first two requests of the pool, which give 119, 150 and 123 on the original tool list (Table~\ref{tab:heldout_directions}).}
\label{tab:seedcount}
\footnotesize
\resizebox{\ifdim\width>\columnwidth\columnwidth\else\width\fi}{!}{%
\begin{tabular}{@{}l r r c c@{}}
\toprule
\textsc{Model} & L & $n$ & \textsc{Original tool list} & \textsc{Renamed tool list} \\
\midrule
\raisebox{-1pt}{\includegraphics[height=8pt]{figures/logos/gemma.png}}\,Gemma~3 4B & 33 & 1 & 90 (71--117) & 154 (132--188) \\
\rowcolor{tablerow}  &  & 2 & 96 (91--100) & 173 (165--181) \\
 &  & 4 & 105 (94--111) & 178 (171--184) \\
\rowcolor{tablerow}  &  & 8 & 103 & 182 \\
\raisebox{-1pt}{\includegraphics[height=8pt]{figures/logos/qwen.png}}\,Qwen~3 4B & 35 & 1 & 121 (110--129) & 176 (167--185) \\
\rowcolor{tablerow}  &  & 2 & 129 (126--134) & 185 (179--193) \\
 &  & 4 & 131 (128--134) & 184 (180--187) \\
\rowcolor{tablerow}  &  & 8 & 143 & 186 \\
\raisebox{-1pt}{\includegraphics[height=8pt]{figures/logos/llama.png}}\,Llama~3.1 8B & 31 & 1 & 96 (75--117) & 141 (131--150) \\
\rowcolor{tablerow}  &  & 2 & 105 (100--110) & 153 (147--158) \\
 &  & 4 & 112 (110--113) & 156 (153--161) \\
\rowcolor{tablerow}  &  & 8 & 114 & 156 \\
\bottomrule
\end{tabular}}

\end{table}

\textbf{Changing the tool list without recomputing vectors.} Removing five tools or adding five new ones leaves the surviving tools' vectors similar on Gemma~3 4B, Qwen~3 4B and Llama~3.1 8B: same-tool cosine averages 0.975--0.995, with a minimum of 0.941. On these changed tool lists, vectors from the original tool list perform similarly to recomputed vectors. Both switch 18--20 of 20 pairs on the two 4B models and 11 of 20 on Llama. The largest difference is on Gemma~3 4B after adding five tools: 18/20 with original vectors versus 20/20 with new ones. This small test supports limited reuse. Recomputation uses 2--3 requests per tool.

\textbf{Near-synonymous pairs.} Steering becomes less reliable as the tool vectors become more similar, but cosine similarity does not give a universal failure threshold (Table~\ref{tab:cliff}). We test 12 fixed pairs, four informative trials per pair, at $\alpha{=}0.7$. On Gemma, every pair that switches in all four trials has cosine at most 0.93, while every less successful pair is above 0.94. Qwen has failures below 0.93, and Llama misses at least one trial on every pair at 0.90 or above. With only four trials per pair, these cutoffs are descriptive. The broader decline is consistent with failed steering on the 10 near-synonymous Sports APIs (Appendix~\ref{app:failure}).

\begin{table}[htbp]
\centering
\caption{Steering success against cosine similarity between tool vectors, for all 12 pairs and three models (four informative trials per pair, $\alpha{=}0.7$). Success requires the generated text to \emph{begin with the full target name}. This is stricter than the prefix and full-name containment rules used elsewhere, so scores are not directly comparable across tables. Half the pairs share their first five characters; five-character containment would accept \texttt{send\_email} when targeting \texttt{send\_message}. Fisher: a Fisher discriminant direction from the same four example states, run together with the steering direction; in that run the steering direction scores 32, 33 and 28 of 48, one trial off the hits column on Qwen and Llama.}
\label{tab:cliff}
\small
\setlength{\tabcolsep}{7pt}
\resizebox{\ifdim\width>\columnwidth\columnwidth\else\width\fi}{!}{%
\begin{tabular}{@{}l ccc ccc ccc@{}}
\toprule
& \multicolumn{3}{c}{Gemma~3 4B} & \multicolumn{3}{c}{Qwen~3 4B} & \multicolumn{3}{c}{Llama~3.1 8B} \\
\textsc{Pair} & cos & hits & Fisher & cos & hits & Fisher & cos & hits & Fisher \\
\midrule
\rowcolor{tablerow} weather$\to$stock & 0.920 & 4/4 & 4/4 & 0.776 & 4/4 & 4/4 & 0.891 & 4/4 & 4/4 \\
weather$\to$directions & 0.888 & 4/4 & 4/4 & 0.724 & 4/4 & 4/4 & 0.844 & 4/4 & 4/4 \\
\rowcolor{tablerow} flight$\to$hotel & 0.927 & 4/4 & 4/4 & 0.865 & 4/4 & 4/4 & 0.926 & 0/4 & 0/4 \\
search$\to$news & 0.967 & 3/4 & 1/4 & 0.941 & 4/4 & 3/4 & 0.911 & 3/4 & 2/4 \\
\rowcolor{tablerow} book\_flight$\to$cancel\_flight & 0.915 & 4/4 & 4/4 & 0.772 & 4/4 & 4/4 & 0.908 & 2/4 & 2/4 \\
schedule$\to$reminder & 0.912 & 4/4 & 4/4 & 0.882 & 1/4 & 1/4 & 0.923 & 3/4 & 3/4 \\
\rowcolor{tablerow} search\_web$\to$lookup\_online & 0.980 & 2/4 & 2/4 & 0.990 & 2/4 & 2/4 & 0.988 & 2/4 & 2/4 \\
order\_food$\to$book\_restaurant & 0.950 & 3/4 & 3/4 & 0.890 & 4/4 & 4/4 & 0.940 & 3/4 & 2/4 \\
\rowcolor{tablerow} find\_flights$\to$search\_flights & 0.989 & 2/4 & 2/4 & 0.973 & 3/4 & 2/4 & 0.993 & 2/4 & 2/4 \\
get\_weather$\to$fetch\_weather & 0.997 & 2/4 & 2/4 & 0.970 & 2/4 & 2/4 & 0.989 & 2/4 & 2/4 \\
\rowcolor{tablerow} send\_email$\to$send\_message & 0.960 & 0/4 & 0/4 & 0.894 & 0/4 & 0/4 & 0.977 & 0/4 & 0/4 \\
play\_song$\to$play\_track & 0.984 & 0/4 & 0/4 & 0.946 & 0/4 & 1/4 & 0.989 & 2/4 & 2/4 \\
\bottomrule
\end{tabular}}

\end{table}

\textbf{Steering among similar tools on MetaTool.} We also steer every MetaTool request from its labelled tool to one other candidate of its ten, chosen at random, with the protocol of the main experiments (tool vectors from other requests, $\alpha{=}0.7$ at the last layer). The call switches to the target on 64--68\% of the 995 requests on the three representative models (Gemma~3 4B 65, Qwen~3 4B 68, Llama~3.1 8B 64\%), and on 66--71\% of the requests whose unsteered call names the labelled tool. Splitting the pairs into thirds by the cosine between the two tool vectors, the most similar third switches on 47--57\% against 67--84\% for the least similar third. So similar tools are harder to switch between, but on this tool list steering does not collapse the way it does on the twelve near-synonym pairs above.

\subsection{Cross-lingual tool space alignment}
\label{app:crosslingual}

We test English tool vectors for readout and steering in other languages.
The experiment covers 5 tools and 8 request languages: English, Chinese, Japanese, French,
Spanish, German, Korean, and Arabic. Tool definitions remain in English. For each tool
and language, we build the tool vector from 1 request and classify 3 held-out requests. Tools
use the main text's \texttt{name(params)} format without descriptions, except for the
separate description comparison below. Table~\ref{tab:crosslingual} reports Gemma~3 4B,
Qwen~3 4B, and Llama~3.1 8B. The Llama run uses the archived script with descriptions
disabled; the other two runs predate that script.

\begin{table}[htbp]
\centering
\caption{Cross-lingual alignment, steering, and classification without tool descriptions. \textsc{same}: mean cosine between each tool's English tool vector and the same tool's vectors in other languages. \textsc{cross}: cosine between different tools across languages. \textsc{gap}: their difference. Steering uses English steering directions on 10 pairs per language. Classification compares held-out requests with English tool vectors only, using 15 requests per language.}
\label{tab:crosslingual}
\small
\setlength{\tabcolsep}{6pt}
\resizebox{\ifdim\width>\columnwidth\columnwidth\else\width\fi}{!}{\begin{tabular}{l ccc cccc cccccccc}
\toprule
& \multicolumn{3}{c}{\textsc{Cosine}} & \multicolumn{4}{c}{\textsc{Steering (\%)}} & \multicolumn{8}{c}{\textsc{Classification (\%)}} \\
\cmidrule(lr){2-4}\cmidrule(lr){5-8}\cmidrule(lr){9-16}
\textsc{Model} & same & cross & gap & zh & ja & fr & es & en & zh & ja & fr & es & de & ko & ar \\
\midrule
\rowcolor{tablerow} Gemma~3 4B & 0.980 & 0.926 & 0.054 & 100 & 90 & 90 & 90 & 87 & 73 & 67 & 93 & 87 & 80 & 60 & 73 \\
Qwen~3 4B & 0.968 & 0.802 & 0.167 & 100 & 100 & 100 & 100 & 93 & 93 & 100 & 93 & 93 & 93 & 93 & 93 \\
\rowcolor{tablerow} Llama~3.1 8B & 0.855 & 0.666 & 0.189 & 100 & 100 & 90 & 90 & 100 & 93 & 100 & 93 & 93 & 100 & 100 & 100 \\
\bottomrule
\end{tabular}}

\end{table}

\textbf{Steering across languages.} English steering directions switch tool
selection in the four tested non-English steering languages: Chinese, Japanese, French,
and Spanish (Table~\ref{tab:crosslingual}). The same 10 ordered pairs are used in each
language. This supports transfer within the tested languages, prompts, and models. Other languages remain untested.

\textbf{Similarity beyond the shared prompt.} Different tools also have high
cross-lingual cosine similarity, so same-tool similarity alone does not show tool identity. Table~\ref{tab:crosslingual} compares the two. The smaller gap on Gemma
shows that its high raw similarity is less tool-specific than Qwen's or Llama's.

\textbf{Classification on held-out requests.} Qwen~3 4B and Llama~3.1 8B classify
consistently across the eight languages, while Gemma~3 4B varies more
(Table~\ref{tab:crosslingual}). Relative to Gemma's English baseline, French, Spanish,
and German are similar; Chinese, Arabic, Japanese, and Korean are lower. A separate run
with descriptions improves Korean from 60\% to 93\% and Japanese from 67\% to 100\%,
suggesting that descriptions can help cross-lingual readout in these cases.

\subsection{Tool descriptions}
\label{app:desc_ablation}

We add short descriptions to the main text's \texttt{name(params)} format, for example
\texttt{weather(location): Check current weather conditions}. Under the
five-character containment rule, all eight instruction-tuned models of 4B and above have both cells (Table~\ref{tab:desc_ablation}). Five are unchanged, Gemma~3 4B falls from 97 to 90\%, Qwen~3 4B rises from 93 to 97\% and Llama~3.1 8B rises from 77 to 90\%. Descriptions
improve Gemma~1B IT but reduce scores on Qwen~0.6B IT and every base model above
the floor.

\begin{table}[htbp]
\centering
\caption{Steering accuracy with and without descriptions, scored by five-character containment over five generated tokens on 30 pairs. A dash denotes a missing matched archive, not a zero score. The two unchanged base models are both at the 3\% floor.}
\label{tab:desc_ablation}

\small
\setlength{\tabcolsep}{7pt}
\rowcolors{2}{tablerow}{white}
\begin{tabular}{@{}l cc cc@{}}
\toprule
& \multicolumn{2}{c}{\textsc{IT}} & \multicolumn{2}{c}{\textsc{Base}} \\
\cmidrule(lr){2-3} \cmidrule(lr){4-5}
\textsc{Model} & \textsc{Name} & \textsc{+Desc} & \textsc{Name} & \textsc{+Desc} \\
\midrule
\multicolumn{5}{@{}l}{\raisebox{-1pt}{\includegraphics[height=8pt]{figures/logos/gemma.png}}\,\textbf{Gemma~3}} \\
\quad 270M & 3\% & 3\% & 3\% & 3\% \\
\quad 1B & 57\% & 63\% $\uparrow$ & 3\% & 3\% \\
\quad 4B & 97\% & 90\% $\downarrow$ & 50\% & 20\% $\downarrow$ \\
\quad 12B & 100\% & 100\% & 47\% & 37\% $\downarrow$ \\
\quad 27B & 100\% & 100\% & 67\% & 63\% $\downarrow$ \\
\midrule
\multicolumn{5}{@{}l}{\raisebox{-1pt}{\includegraphics[height=8pt]{figures/logos/qwen.png}}\,\textbf{Qwen~3}} \\
\quad 0.6B & 67\% & 50\% $\downarrow$ & \multicolumn{2}{c}{---} \\
\quad 1.7B & 87\% & 87\% & \multicolumn{2}{c}{---} \\
\quad 4B & 93\% & 97\% $\uparrow$ & \multicolumn{2}{c}{---} \\
\quad 8B & 100\% & 100\% & \multicolumn{2}{c}{---} \\
\quad 14B & 100\% & 100\% & \multicolumn{2}{c}{---} \\
\midrule
\multicolumn{5}{@{}l}{\textbf{Others}} \\
\quad \raisebox{-1pt}{\includegraphics[height=8pt]{figures/logos/qwen.png}}\,Qwen~2.5 7B & 97\% & 97\% & \multicolumn{2}{c}{---} \\
\quad \raisebox{-1pt}{\includegraphics[height=8pt]{figures/logos/llama.png}}\,Llama~3.1 8B & 77\% & 90\% $\uparrow$ & 77\% & 50\% $\downarrow$ \\
\bottomrule
\end{tabular}
\rowcolors{1}{}{}

\end{table}

The comparison measures emitted names under the five-character rule, so the effect of descriptions on the underlying circuit stays open. Because tool names share prefixes, these rates also overstate complete API selection.

\subsection{Anonymous-name control}
\label{app:anon_names}

Replacing tool names with anonymous IDs changes PCA dimensionality little
(Table~\ref{tab:anon_names}). We replace the 15 names with \texttt{tool\_00} through
\texttt{tool\_14}, preserving parameter lists and descriptions, then recompute the tool vectors
and PCA. The table reports geometry and unsteered tool-calling accuracy.

\begin{table}[htbp]
\centering
\caption{Anonymous-name control. \textsc{Base}: unsteered tool-calling accuracy. \textsc{top-10}: cumulative variance of the top 10 principal components. Similar $k_{90}$ and top-10 variance under both naming schemes show that low dimensionality does not require the original tool names.}
\label{tab:anon_names}

\small
\setlength{\tabcolsep}{5pt}
\begin{tabular}{@{}l cc cc cc@{}}
\toprule
 & \multicolumn{2}{c}{\textsc{Base}} & \multicolumn{2}{c}{\textsc{$k_{90}$}} & \multicolumn{2}{c}{\textsc{top-10}} \\
\cmidrule(lr){2-3} \cmidrule(lr){4-5} \cmidrule(lr){6-7}
& real & anon & real & anon & real & anon \\
\midrule
\raisebox{-1pt}{\includegraphics[height=8pt]{figures/logos/gemma.png}}\,Gemma~3 4B IT    & 100\% & 80\%  & 10 & 9  & 0.91 & 0.93 \\
\raisebox{-1pt}{\includegraphics[height=8pt]{figures/logos/gemma.png}}\,Gemma~3 4B base  & 73\%  & 47\%  & 10 & 10 & 0.90 & 0.90 \\
\raisebox{-1pt}{\includegraphics[height=8pt]{figures/logos/qwen.png}}\,Qwen~3 4B        & 100\% & 100\% & 11 & 11 & 0.88 & 0.88 \\
\raisebox{-1pt}{\includegraphics[height=8pt]{figures/logos/llama.png}}\,Llama~3.1 8B IT  & 100\% & 100\% & 11 & 11 & 0.88 & 0.87 \\
\bottomrule
\end{tabular}

\end{table}

This control changes the names while preserving descriptive content, so it cannot separate
tool identity from the descriptions' topics. The within-topic analysis addresses that
question (Section~\ref{sec:discussion}). Steering under anonymous IDs tests the same first-token limitation (Section~\ref{sec:method}). The IDs share the token \texttt{tool\_}; selecting that token alone leaves the model to complete the ID from its existing state. The stored runs show little successful redirection under this naming scheme. Under the
held-out 15-tool protocol with descriptions, the readable names give 187 to 209 full-name
switches out of 210 on seven models and the anonymous IDs at most 2. When the readable
names are shuffled across functions, the direction follows the label printed on the tool list
rather than the label a reader would expect on five of the seven.

\subsection{List-ordering control}
\label{app:position_bias}

Tool vectors remain similar across list orderings (Table~\ref{tab:position_bias}). We
reorder the 15-tool list five ways: identity, reverse, and three random permutations from
a fixed seed. Each ordering uses the same two requests per tool as
Appendix~\ref{app:pca}. We compare each tool's vectors by pairwise cosine,
recompute $k_{90}$ within each ordering, and decompose variance into tool and position
effects. The control tests representation geometry and unsteered selection.

\begin{table}[htbp]
\centering
\caption{List-ordering control. \textsc{cos}: mean pairwise cosine for the same tool across five orderings. \textsc{$k_{90}$}: components needed for 90\% variance, reported as the range across orderings. \textsc{Tool} and \textsc{Pos}: fractions of total variance attributed to identity and position under an additive decomposition. \textsc{Tool/Pos}: their ratio. Tool identity accounts for more variance than list position on all five models.}
\label{tab:position_bias}

\small
\setlength{\tabcolsep}{2pt}
\begin{tabular}{@{}l c c c c c@{}}
\toprule
\textsc{Model} & \textsc{cos} & \textsc{$k_{90}$} & \textsc{Tool} & \textsc{Pos} & \textsc{Tool/Pos} \\
\midrule
\raisebox{-1pt}{\includegraphics[height=8pt]{figures/logos/gemma.png}}\,Gemma~3 1B IT    & 0.93 & 9--10  & 16.3\% & 3.6\% & 4.5$\times$ \\
\rowcolor{tablerow} \raisebox{-1pt}{\includegraphics[height=8pt]{figures/logos/gemma.png}}\,Gemma~3 4B IT    & 0.98 & 9--11  & 16.6\% & 3.9\% & 4.3$\times$ \\
\raisebox{-1pt}{\includegraphics[height=8pt]{figures/logos/gemma.png}}\,Gemma~3 4B base  & \textbf{1.00} & 10--11 & 15.0\% & 3.3\% & 4.5$\times$ \\
\rowcolor{tablerow} \raisebox{-1pt}{\includegraphics[height=8pt]{figures/logos/qwen.png}}\,Qwen~3 4B        & 0.96 & 11     & 17.6\% & 3.6\% & 4.9$\times$ \\
\raisebox{-1pt}{\includegraphics[height=8pt]{figures/logos/llama.png}}\,Llama~3.1 8B IT  & 0.96 & 11     & 17.0\% & 3.5\% & 4.9$\times$ \\
\bottomrule
\end{tabular}

\end{table}

Unsteered accuracy remains near 100\% at each position on the instruction-tuned 4B+
models, with occasional declines when a longer name occupies the last slot. Gemma~3 1B
IT is less accurate overall, but its errors are spread across positions. These runs do
not show a consistent preference for the beginning or end of the list.

\section{Depth analysis}

\subsection{The direction beyond the output row}
\label{app:output_row}

\subsubsection{Steering with the unembedding direction alone}
\label{app:unembed_steer_control}

If steering worked only through alignment with the target's output row, that row alone
should be equally effective. We compare four magnitude-matched directions on the 15-tool
benchmark: the steering direction, the target's first-token unembedding row, the difference
between target and source rows, and a random unit vector. Each receives a displacement
norm of $\alpha{=}1.0$ times the distance between the tool vectors at that layer. This
run (Table~\ref{tab:unembed_steer_control}) mixes protocols; the depth comparison in
Figure~\ref{fig:depth_3fam}, top row, uses one protocol throughout.

\begin{table}[htbp]
\centering
\caption{Full-name steering accuracy (\%) for magnitude-matched directions: $N{=}100$ sampled ordered pairs, $\alpha{=}1.0$, seven layers per model. \textsc{$W_U$ diff} is the difference of the tools' first-token output directions; \textsc{$W_U$ row} is the target's alone. Both are unit-normalised and given the same coefficient as the steering direction. The four \texttt{get\_} tools share a first token, making \textsc{$W_U$ diff} zero for 7 of the 100 pairs; those pairs are excluded from that column. The mean difference outperforms both output-direction controls in the separation band, while the controls overtake it near the output, consistent with the crossover in Figure~\ref{fig:depth_3fam}.}
\label{tab:unembed_steer_control}

\small
\setlength{\tabcolsep}{2pt}
\resizebox{\ifdim\width>\columnwidth\columnwidth\else\width\fi}{!}{\begin{tabular}{@{}ll rrrr@{}}
\toprule
\textsc{Model} & \textsc{L} & \textsc{Steer} & \textsc{$W_U$ diff} & \textsc{$W_U$ row} & \textsc{Rand} \\
\midrule
\raisebox{-1pt}{\includegraphics[height=8pt]{figures/logos/gemma.png}}\,Gemma~3 4B & 8 & 0 & 0 & 0 & 0 \\
  & 17 & 0 & 1 & 1 & 0 \\
  & 22 & 41 & 20 & 10 & 0 \\
  & 27 & 45 & 70 & 54 & 0 \\
  & 30 & 49 & 71 & 66 & 0 \\
  & 32 & 48 & 71 & 66 & 0 \\
  & 33 & 44 & 71 & 66 & 0 \\
\midrule
\raisebox{-1pt}{\includegraphics[height=8pt]{figures/logos/qwen.png}}\,Qwen~3 4B & 9 & 0 & 0 & 0 & 0 \\
  & 18 & 0 & 0 & 0 & 0 \\
  & 23 & 46 & 0 & 0 & 0 \\
  & 28 & 59 & 36 & 10 & 0 \\
  & 32 & 66 & 78 & 54 & 0 \\
  & 34 & 66 & 75 & 39 & 0 \\
  & 35 & 67 & 78 & 73 & 0 \\
\midrule
\raisebox{-1pt}{\includegraphics[height=8pt]{figures/logos/llama.png}}\,Llama~3.1 8B & 8 & 0 & 0 & 0 & 0 \\
  & 16 & 15 & 1 & 1 & 0 \\
  & 20 & 26 & 9 & 2 & 0 \\
  & 25 & 44 & 73 & 50 & 0 \\
  & 28 & 55 & 73 & 58 & 0 \\
  & 30 & 62 & 73 & 53 & 0 \\
  & 31 & 53 & 73 & 68 & 0 \\
\bottomrule
\end{tabular}}

\end{table}

This control and Appendices~\ref{app:base_it}, \ref{app:base_it}, and
\ref{app:position_patching} use HuggingFace forward hooks on raw layer outputs before the
final normalisation. The other experiments use TransformerLens. These controls use a
single seed, so small differences may be noise.

\subsubsection{Unembedding decomposition of the steering direction}
\label{app:unembed_decomp}
\begin{figure}[htbp]
\centering
\includegraphics[width=\linewidth]{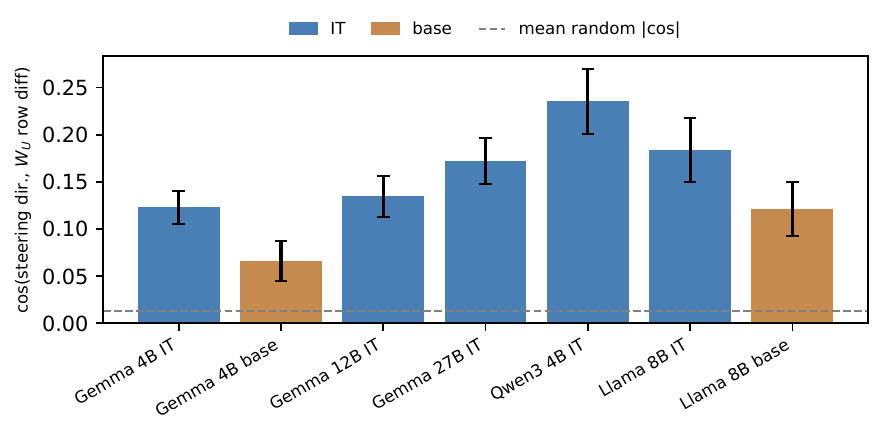}
\caption{Mean cosine between each steering direction and the difference of the two tools' first-token output directions (unembedding rows scaled by the final-norm weight; 198 pairs per model, excluding pairs whose tool names share a first token, for which the unembedding difference is zero; error bars one standard deviation; dashed line the mean random-direction level, 0.013). IT models show 1.5 to 1.9 times the base-model alignment (Gemma~3 4B 0.12 vs.\ 0.07, Llama~3.1 8B 0.18 vs.\ 0.12).}
\label{fig:alignment}
\end{figure}

Steering directions align more strongly with the difference of the tools'
first-token output directions than random vectors do (Figure~\ref{fig:alignment}). Across
198 pairs per model, cosine similarity is $+$0.12 to $+$0.24 for instruction-tuned models
and 0.07 to 0.12 for base models. Mean random-direction $|\cos|$ is 0.011 to 0.015
(200 draws per model); even the weakest observed alignment is 4.3 times the largest
random baseline.

First-token output directions cannot distinguish tools with a shared tokenizer prefix,
such as \texttt{get\_weather}, \texttt{get\_stock\_price}, \texttt{get\_news}, and
\texttt{get\_directions}. We therefore repeat the alignment check using sums of raw
unembedding rows over each complete tool name, before final-norm scaling. Mean alignment
changes little: Gemma~3 4B gives 0.185 for full names versus 0.190 for first tokens, and
Qwen~3 4B gives 0.249 versus 0.243. The full-name construction also covers the 12 shared-first-token pairs excluded when the first-token difference is zero.

We split each steering direction $d = \bar{\resid}_b - \bar{\resid}_a$ into a component
parallel to the target's first-token output direction and an orthogonal remainder. The
parallel component has only 7--17\% of the full norm. To separate direction from magnitude,
we test each component at its own norm and rescaled to the full vector's L2 norm
(``par$^{\star}$'' and ``orth$^{\star}$''; Table~\ref{tab:unembed_decomp}).

\begin{table}[htbp]
\centering
\caption{Steering-vector decomposition against the target's first-token output direction (unembedding row scaled by the final-norm weight), on 30 source$\to$target pairs. Full: the original $\bar{\resid}_b - \bar{\resid}_a$ at $\alpha{=}0.7$. Par and Orth: parallel and orthogonal components at their own norms. Par$^{\star}$ and Orth$^{\star}$: components rescaled to the full L2 norm. Zero: no steering. The depth sweep in Table~\ref{tab:orth_depth} uses $\alpha{=}1.0$, so its last-layer rates differ. In both runs, the orthogonal component alone is much less effective than the full vector at the last layer.}
\label{tab:unembed_decomp}

\footnotesize
\setlength{\tabcolsep}{2pt}
\begin{tabular}{@{}l ccccc c@{}}
\toprule
\textsc{Model} & \textsc{Full} & \textsc{par} & \textsc{par$^{\star}$} & \textsc{orth} & \textsc{orth$^{\star}$} & \textsc{Zero} \\
\midrule
Gemma~3 4B IT    & 97\%  & 0\% & \textbf{100\%} & 17\% & 17\% & 0\% \\
\rowcolor{tablerow} Gemma~3 4B base  & 50\%  & 0\% & \textbf{100\%} & 7\%  & 7\%  & 0\% \\
Qwen~3 4B        & 93\%  & 0\% & \textbf{100\%} & 0\%  & 0\%  & 0\% \\
\rowcolor{tablerow} Llama~3.1 8B IT  & 77\%  & 7\% & \textbf{93\%}  & 17\% & 17\% & 0\% \\
Llama~3.1 8B base& 77\%  & 0\% & \textbf{100\%} & 7\%  & 7\%  & 0\% \\
\bottomrule
\end{tabular}

\end{table}

At matched magnitude, the parallel component steers both base and instruction-tuned
models successfully (Table~\ref{tab:unembed_decomp}). This test alone therefore cannot
explain their full-vector performance gap. Gemma~3 4B's higher instruction-tuned
full-vector rate coincides with stronger alignment (Figure~\ref{fig:alignment}).

\subsubsection{Depth checks on the steering direction}
\label{app:depth_checks}

Two other analyses locate tool-choice information near the peak layer. The largest
component-patching effect occurs one or two layers away (Section~\ref{sec:circuit}). In base
models, cosine readout identifies tools there even when the output-layer readout remains
near chance (Appendix~\ref{app:base_it}).

The depth experiments in Section~\ref{sec:depth} preserve the corresponding steering or patching
protocol while varying the layer. Figure~\ref{fig:depth_3fam} reports two distinct outcomes. The top row measures generated-tool switches after adding a vector, and the bottom row measures logit-gap recovery after copying components of the model's own state. These
rates are not directly comparable, even when measured at the same layer.

\textbf{Removing the output-row component.} Throughout this appendix and Figure~\ref{fig:depth_3fam}, a tool's output direction is its first-token unembedding row scaled elementwise by the final norm's weight (on Gemma, $1{+}w$). That row is the direction along which the residual sets that token's logit. The raw row has cosine 0.81 to it on Gemma~3 4B. For each ordered pair we split the
steering direction $\direction_{a \to b}$ against the target tool's first-token unembedding row
$u_b$ into a parallel and an orthogonal part, then steer with the orthogonal part alone.
Each condition is injected at $\alpha$ times the part it steers with. The orthogonal condition therefore displaces the state by almost exactly what the full vector does. Per cell it averages 98.5--99.99\% of the vector's length, 98.5--99.6\%
of it at the last layers, and it never drops below 96.9\% on a single pair.

Unsteered generation reaches the target in 0 of 30 pairs
in all eighteen cells, and the fifteen tool names share no four-character prefix. At its own magnitude the
parallel part switches at most 3\% of pairs at every layer below the last, and 13--57\% at the last layer. Scaled back up to the full magnitude, it reaches 93--100\% at the last layer across five models (Gemma~3 4B IT and base, Qwen~3 4B, Llama~3.1 8B IT and base). So the rescaled component is sufficient for the tested target-name switches
at that depth. Comparisons at the components' natural norms are valid only within a layer.
The parallel component carries between 1\% and 19\% of the full norm across models and
depths, so comparing it across rows also changes the injected magnitude.

\begin{table}[htbp]
\centering
\caption{Switch rate (\%) for the full steering direction, for the same vector with its component along the target tool's output direction projected out (\textsc{Orth$_1$}), and with its component along the span of \emph{every} listed tool's output direction projected out (\textsc{Orth$_{\rm all}$}), at six depths per model ($n{=}30$ ordered pairs, $\alpha{=}1.0$, greedy decoding, first-token prefix rule). Output directions are the unembedding rows scaled by the final norm's weight, the direction that sets the token's logit. The last column is \textsc{Orth$_{\rm all}$} as a percentage of \textsc{Full}, blank where the full vector switches nothing.}
\label{tab:orth_depth}
\small
\setlength{\tabcolsep}{7pt}
\resizebox{\ifdim\width>\columnwidth\columnwidth\else\width\fi}{!}{%
\begin{tabular}{@{}l rr rrr r@{}}
\toprule
\textsc{Model} & L & \% & \textsc{Full} & \textsc{Orth$_1$} & \textsc{Orth$_{\rm all}$} & \textsc{Orth$_{\rm all}$/Full} \\
\midrule
\raisebox{-1pt}{\includegraphics[height=8pt]{figures/logos/gemma.png}}\,Gemma~3 4B & 15 & 45 & 0 & 0 & 0 & -- \\
\rowcolor{tablerow}  & 18 & 55 & 33 & 33 & 33 & 100 \\
 & 21 & 64 & 97 & 97 & 97 & 100 \\
\rowcolor{tablerow}  & 24 & 73 & 100 & 100 & 93 & 93 \\
 & 27 & 82 & 100 & 100 & 73 & 73 \\
\rowcolor{tablerow}  & 33 & 100 & 100 & 30 & \textbf{0} & 0 \\
\midrule
\raisebox{-1pt}{\includegraphics[height=8pt]{figures/logos/qwen.png}}\,Qwen~3 4B & 19 & 54 & 0 & 0 & 0 & -- \\
\rowcolor{tablerow}  & 22 & 63 & 57 & 57 & 53 & 94 \\
 & 25 & 71 & 93 & 93 & 93 & 100 \\
\rowcolor{tablerow}  & 28 & 80 & 93 & 93 & 93 & 100 \\
 & 31 & 89 & 93 & 93 & 60 & 64 \\
\rowcolor{tablerow}  & 35 & 100 & 93 & 0 & \textbf{0} & 0 \\
\midrule
\raisebox{-1pt}{\includegraphics[height=8pt]{figures/logos/llama.png}}\,Llama~3.1 8B & 12 & 39 & 0 & 0 & 0 & -- \\
\rowcolor{tablerow}  & 15 & 48 & 53 & 53 & 53 & 100 \\
 & 18 & 58 & 87 & 87 & 87 & 100 \\
\rowcolor{tablerow}  & 21 & 68 & 93 & 90 & 93 & 100 \\
 & 24 & 77 & 93 & 80 & 73 & 79 \\
\rowcolor{tablerow}  & 31 & 100 & 97 & 20 & \textbf{0} & 0 \\
\bottomrule
\end{tabular}}

\end{table}

\textbf{Held-out steering at the peak layer.} We also checked selected layers from in-sample sweeps. One sweep built each tool vector from a single request and steered that same request. With one example request and unit strength, this is an exact activation swap at the intervened token. A broader sweep used 30 to 50 pairs per cell and at most two example requests per tool, again reusing test requests in the tool vectors. Table~\ref{tab:midstack_heldout} repeats both steering protocols at the selected layer with held-out requests. On the renamed 15-tool list, tool vectors use two requests and the held-out third is tested across all 210 ordered pairs. Gemma~3 4B and Qwen~3 4B closely match their in-sample sweeps: 51.4\% versus 52.0\%, and 73.8\% versus 74.0\%, respectively, at the same layer and $\alpha$. Llama~3.1 8B reaches 47.6\% versus 38.0\% at the sweep's nearest measured layer; that sweep samples odd layers and omits L18. Its higher rate may reflect the use of two example requests rather than one, but the layer difference prevents a direct comparison. Request reuse causes little inflation in these cases. These runs test steering at selected layers and do not repeat the output-direction controls. Appendix~\ref{app:output_row} gives the full comparison on held-out requests. The state-difference patching in Figure~\ref{fig:depth_3fam}, bottom row, is separate, since its directions exclude both evaluated requests (Appendix~\ref{app:direction_patch}).

\begin{table}[htbp]
\centering
\caption{Steering at the last layer against the peak layer, on the paper's own tool
sets and protocols. 5-tool: leave-one-out over three requests per tool, 20 ordered pairs
$\times$ 3 splits. 15-tool: tool vectors from requests 0--1, tested on a held-out third request, all
210 ordered pairs at $\alpha{=}0.7$, scored on the full name. Increasing the displacement in the 5-tool protocol narrows the gap between peak-layer and last-layer steering.}
\label{tab:midstack_heldout}
\small
\setlength{\tabcolsep}{6pt}
\begin{tabular}{@{}l l rrrr@{}}
\toprule
& & \multicolumn{2}{c}{5-tool} & \multicolumn{2}{c}{15-tool} \\
\cmidrule(lr){3-4}\cmidrule(l){5-6}
\textsc{Model} & \textsc{Layer} & $\alpha{=}0.7$ & $\alpha{=}1$ & orig & renamed \\
\midrule
\raisebox{-1pt}{\includegraphics[height=8pt]{figures/logos/gemma.png}}\,Gemma~3 4B & L33 last & 83 & 100 & 56 & 89 \\
 & L21 (64\%) & 70 & 95 & 47 & 51 \\
\midrule
\raisebox{-1pt}{\includegraphics[height=8pt]{figures/logos/qwen.png}}\,Qwen~3 4B & L35 last & 88 & 100 & 71 & 90 \\
 & L25 (71\%) & 78 & 100 & 66 & 74 \\
\midrule
\raisebox{-1pt}{\includegraphics[height=8pt]{figures/logos/llama.png}}\,Llama~3.1 8B & L31 last & 92 & 97 & 60 & 77 \\
 & L18 (58\%) & 70 & 98 & 35 & 48 \\
\bottomrule
\end{tabular}

\end{table}

\subsubsection{Held-out comparison of directions across depth}
\label{app:heldout_directions}

Figure~\ref{fig:depth_3fam}, top row, and Table~\ref{tab:heldout_directions} repeat the
depth comparison under the held-out protocol of the 15-tool run: tool vectors from two requests per
tool, the third request held out, all 210 ordered pairs, $\alpha{=}0.7$. At each of eight
depths and for both tool lists, five unit directions receive the same displacement, $0.7$ times
the distance between the pair's tool vectors, so the conditions differ in direction alone. They
are the steering direction; the steering direction with its component along the two tools'
output-direction difference removed and renormalised; that output-direction difference; the
target's output direction alone; and a random direction fixed per pair. Output directions
are the first-token unembedding rows scaled by the final norm's weight, as elsewhere in this
appendix. On the original tool list the four \texttt{get\_} tools share a first token, so the
output-direction difference is zero for their 12 pairs and that column counts out of 198.
At the last layer, the steering-direction condition of this run exactly reproduces the accompanying standard steering run and matches Table~\ref{tab:generalization} to within one pair. The full-name counts are 119 versus 118 on Gemma~3 4B, 150 versus 149 on Qwen~3 4B, and 125 versus 126 on Llama~3.1 8B. Table~\ref{tab:generalization} reports a separate run of the same protocol; the depth and fold comparisons below use the run described here.

The removed component is small at every depth. Its cosine to the steering direction, in the
last column, stays below 0.11 up to the peak layer and reaches 0.13 to 0.25 at the
last layer. Removing it changes little at the peak layer and removes the effect at the last layer. The random direction switches at most three pairs in any cell.

We add a direction that needs no source tool, the target's tool vector minus the average of the fifteen tool vectors, to the held-out run at the peak layer and the last layer (Table~\ref{tab:srcfree}). It switches fewer pairs than the steering direction, 51--63\% as many at the peak layer and 68--80\% as many at the last layer on the original tool list, and far more than the random direction. The steering-direction cells of that run match Table~\ref{tab:heldout_directions}. Gemma and Qwen ran on the same device as that table; Llama ran on a different GPU, where the counts differ by up to three pairs. A Fisher discriminant direction, which accounts for how the example states spread around each tool vector, switches 14 and 20 more pairs than the steering direction at Gemma's peak layer, and between 4 more and 18 fewer in the other ten cells. With two example requests per tool, that spread is poorly estimated, which limits what the Fisher direction can add. The LAT direction, built from the same example states, switches between 19 fewer and 3 more pairs than the steering direction; it is below the steering direction in all four Gemma cells and within 4 pairs of it on Qwen and Llama.

\begin{table}[htbp]
\centering
\caption{Held-out full-name switches out of 210 ordered pairs for the steering direction (Steer), for the target's tool vector minus the average of all fifteen tool vectors (Target$-$avg), and for a Fisher discriminant direction (the mean difference multiplied by the inverse of the within-tool covariance of the example states, shrunk halfway toward a scaled identity), and for the LAT direction of \citet{zou2023representation} (the top singular vector of the differences between the two tools' example states, sign-aligned with the mean difference), at the peak layer and the last layer of each model, same displacement and protocol as Table~\ref{tab:heldout_directions}. The Fisher counts come from separate runs; their steering-direction cells match this table except on Qwen, which ran on a different GPU and differs by up to three pairs.}
\label{tab:srcfree}
\footnotesize
\resizebox{\ifdim\width>\columnwidth\columnwidth\else\width\fi}{!}{%
\begin{tabular}{@{}l r rrrr rrrr@{}}
\toprule
& & \multicolumn{4}{c}{\textsc{Original tool list}} & \multicolumn{4}{c}{\textsc{Renamed tool list}} \\
\cmidrule(lr){3-6}\cmidrule(lr){7-10}
\textsc{Model} & L & Steer & Target$-$avg & Fisher & LAT & Steer & Target$-$avg & Fisher & LAT \\
\midrule
\raisebox{-1pt}{\includegraphics[height=8pt]{figures/logos/gemma.png}}\,Gemma~3 4B & 20 & 92 & 47 & 112 & 82 & 108 & 40 & 122 & 98 \\
\rowcolor{tablerow}  & 33 & 119 & 81 & 115 & 101 & 187 & 125 & 191 & 184 \\
\raisebox{-1pt}{\includegraphics[height=8pt]{figures/logos/qwen.png}}\,Qwen~3 4B & 24 & 125 & 69 & 112 & 123 & 123 & 49 & 107 & 125 \\
\rowcolor{tablerow}  & 35 & 150 & 117 & 144 & 148 & 188 & 145 & 178 & 190 \\
\raisebox{-1pt}{\includegraphics[height=8pt]{figures/logos/llama.png}}\,Llama~3.1 8B & 16 & 52 & 33 & 52 & 50 & 64 & 34 & 63 & 64 \\
\rowcolor{tablerow}  & 31 & 123 & 98 & 107 & 125 & 162 & 136 & 144 & 164 \\
\bottomrule
\end{tabular}}

\end{table}

\begin{table}[tb]
\centering
\caption{Held-out full-name switches out of 210 ordered pairs by depth (L, and \% of layers) for five unit directions at the same displacement: the steering direction (Steer), the steering direction with its $\Delta W_U$ component removed (Orth), the difference of the two tools' output directions ($\Delta W_U$; out of 198 on the original tool list), the target's output direction alone ($W_U$), and a random direction (Rand). The last column is the mean cosine between $\Delta W_U$ and the steering direction on the renamed tool list. Means from two requests per tool, third request held out, $\alpha{=}0.7$.}
\label{tab:heldout_directions}
\footnotesize
\setlength{\tabcolsep}{2.6pt}
\resizebox{\ifdim\width>\columnwidth\columnwidth\else\width\fi}{!}{%
\begin{tabular}{@{}l rr rrrrr rrrrr r@{}}
\toprule
& & & \multicolumn{5}{c}{\textsc{Original tool list}} & \multicolumn{5}{c}{\textsc{Renamed tool list}} & \\
\cmidrule(lr){4-8}\cmidrule(lr){9-13}
\textsc{Model} & L & \% & Steer & Orth & $\Delta W_U$ & $W_U$ & Rand & Steer & Orth & $\Delta W_U$ & $W_U$ & Rand & cos \\
\midrule
\raisebox{-1pt}{\includegraphics[height=8pt]{figures/logos/gemma.png}}\,Gemma~3 4B & 10 & 30 & 1 & 1 & 1 & 1 & 1 & 1 & 1 & 1 & 1 & 1 & 0.00 \\
\rowcolor{tablerow}  & 13 & 39 & 1 & 1 & 3 & 2 & 1 & 1 & 1 & 2 & 2 & 1 & 0.00 \\
 & 16 & 48 & 5 & 5 & 1 & 1 & 1 & 6 & 6 & 3 & 2 & 1 & 0.04 \\
\rowcolor{tablerow}  & 20 & 61 & 92 & 93 & 22 & 2 & 1 & 108 & 101 & 37 & 13 & 1 & 0.05 \\
 & 23 & 70 & 105 & 78 & 115 & 100 & 1 & 170 & 101 & 203 & 171 & 1 & 0.15 \\
\rowcolor{tablerow}  & 26 & 79 & 118 & 59 & 136 & 133 & 1 & 184 & 46 & 210 & 207 & 1 & 0.20 \\
 & 30 & 91 & 112 & 7 & 136 & 135 & 1 & 192 & 3 & 210 & 210 & 1 & 0.22 \\
\rowcolor{tablerow}  & 33 & 100 & 119 & 0 & 136 & 136 & 1 & 187 & 1 & 210 & 210 & 1 & 0.13 \\
\midrule
\raisebox{-1pt}{\includegraphics[height=8pt]{figures/logos/qwen.png}}\,Qwen~3 4B & 10 & 29 & 2 & 2 & 3 & 2 & 3 & 2 & 2 & 2 & 2 & 2 & 0.03 \\
\rowcolor{tablerow}  & 14 & 40 & 3 & 3 & 3 & 3 & 2 & 2 & 2 & 2 & 2 & 2 & 0.02 \\
 & 18 & 51 & 4 & 4 & 3 & 4 & 3 & 2 & 2 & 2 & 2 & 2 & 0.03 \\
\rowcolor{tablerow}  & 21 & 60 & 8 & 9 & 5 & 4 & 3 & 9 & 9 & 2 & 2 & 2 & 0.03 \\
 & 24 & 69 & 125 & 113 & 49 & 45 & 2 & 123 & 103 & 69 & 47 & 2 & 0.10 \\
\rowcolor{tablerow}  & 28 & 80 & 145 & 131 & 128 & 96 & 1 & 183 & 147 & 175 & 133 & 2 & 0.16 \\
 & 32 & 91 & 150 & 63 & 163 & 139 & 1 & 189 & 37 & 205 & 170 & 2 & 0.29 \\
\rowcolor{tablerow}  & 35 & 100 & 150 & 3 & 165 & 165 & 2 & 188 & 2 & 210 & 210 & 2 & 0.25 \\
\midrule
\raisebox{-1pt}{\includegraphics[height=8pt]{figures/logos/llama.png}}\,Llama~3.1 8B & 9 & 29 & 0 & 0 & 0 & 0 & 0 & 1 & 1 & 1 & 1 & 1 & 0.00 \\
\rowcolor{tablerow}  & 12 & 39 & 3 & 3 & 1 & 2 & 0 & 1 & 2 & 1 & 1 & 1 & 0.00 \\
 & 16 & 52 & 53 & 51 & 9 & 6 & 1 & 63 & 60 & 9 & 4 & 1 & 0.02 \\
\rowcolor{tablerow}  & 19 & 61 & 87 & 76 & 35 & 33 & 1 & 121 & 110 & 72 & 57 & 1 & 0.05 \\
 & 22 & 71 & 124 & 110 & 129 & 89 & 1 & 160 & 130 & 197 & 150 & 1 & 0.11 \\
\rowcolor{tablerow}  & 25 & 81 & 130 & 61 & 152 & 133 & 0 & 166 & 69 & 209 & 195 & 1 & 0.19 \\
 & 28 & 90 & 132 & 31 & 158 & 128 & 1 & 169 & 32 & 196 & 160 & 1 & 0.20 \\
\rowcolor{tablerow}  & 31 & 100 & 125 & 0 & 159 & 159 & 2 & 165 & 1 & 210 & 210 & 1 & 0.21 \\
\bottomrule
\end{tabular}}

\end{table}

\textbf{A conditional first-token limit.} The last-layer edit changes only the first generated token's distribution. In the eight headline models, every observed \texttt{get}-prefixed continuation from a \texttt{get\_} source request returns that source name. Under this observed behavior, the eleven unique-first-token targets contribute at most $11 \times 14 = 154$ switches, and each of the eleven non-\texttt{get\_} sources can reach at most one \texttt{get\_} target through the shared first token. This gives the conditional reference of 165 of 210 used in Section~\ref{sec:method}. Different continuations or tokenizations could permit other outcomes. Table~\ref{tab:first_token_cap} reports all 18 models against this common reference. None exceeds it in the stored run, but the smaller models do include \texttt{get\_}-to-\texttt{get\_} hits where the unsteered generation was already wrong. Across the paired records, steered and unsteered generations sharing the first token are identical in 182 of 185 cases; the three exceptions occur on Qwen~3 0.6B after the tool name.

\begin{table}[htbp]
\centering
\caption{Full-name outcomes on the original 15-tool list for all 18 models. \textsc{Switches}: full-name hits out of 210. \textsc{\% of cap}: the count as a share of the conditional 165-pair reference derived for the eight headline models. \texttt{get\_}$\to$\texttt{get\_}: hits between two \texttt{get\_} tools, out of 12. \textsc{Max}: the largest number of distinct \texttt{get\_} targets any non-\texttt{get\_} source reached, out of 4.}
\label{tab:first_token_cap}
\small
\setlength{\tabcolsep}{7pt}
\resizebox{\ifdim\width>\columnwidth\columnwidth\else\width\fi}{!}{%
\begin{tabular}{@{}l rrrr@{}}
\toprule
\textsc{Model} & \textsc{Switches} & \textsc{\% of cap} & \texttt{get\_}$\to$\texttt{get\_} & \textsc{Max} \texttt{get\_} \textsc{per source} \\
\midrule
\raisebox{-1pt}{\includegraphics[height=8pt]{figures/logos/gemma.png}}\,Gemma~3 270M base & 14 & 8 & 3 & 1 \\
\rowcolor{tablerow} \raisebox{-1pt}{\includegraphics[height=8pt]{figures/logos/gemma.png}}\,Gemma~3 270M & 14 & 8 & 3 & 1 \\
\raisebox{-1pt}{\includegraphics[height=8pt]{figures/logos/gemma.png}}\,Gemma~3 1B base & 14 & 8 & 3 & 1 \\
\rowcolor{tablerow} \raisebox{-1pt}{\includegraphics[height=8pt]{figures/logos/gemma.png}}\,Gemma~3 1B & 66 & 40 & 0 & 1 \\
\raisebox{-1pt}{\includegraphics[height=8pt]{figures/logos/gemma.png}}\,Gemma~3 4B base & 57 & 35 & 1 & 1 \\
\rowcolor{tablerow} \raisebox{-1pt}{\includegraphics[height=8pt]{figures/logos/gemma.png}}\,Gemma~3 4B & 118 & 72 & 0 & 1 \\
\raisebox{-1pt}{\includegraphics[height=8pt]{figures/logos/gemma.png}}\,Gemma~3 12B base & 71 & 43 & 1 & 1 \\
\rowcolor{tablerow} \raisebox{-1pt}{\includegraphics[height=8pt]{figures/logos/gemma.png}}\,Gemma~3 12B & 154 & 93 & 0 & 1 \\
\raisebox{-1pt}{\includegraphics[height=8pt]{figures/logos/gemma.png}}\,Gemma~3 27B base & 130 & 79 & 0 & 1 \\
\rowcolor{tablerow} \raisebox{-1pt}{\includegraphics[height=8pt]{figures/logos/gemma.png}}\,Gemma~3 27B & 163 & 99 & 0 & 1 \\
\raisebox{-1pt}{\includegraphics[height=8pt]{figures/logos/qwen.png}}\,Qwen~3 0.6B & 94 & 57 & 1 & 1 \\
\rowcolor{tablerow} \raisebox{-1pt}{\includegraphics[height=8pt]{figures/logos/qwen.png}}\,Qwen~3 1.7B & 144 & 87 & 0 & 1 \\
\raisebox{-1pt}{\includegraphics[height=8pt]{figures/logos/qwen.png}}\,Qwen~3 4B & 149 & 90 & 0 & 1 \\
\rowcolor{tablerow} \raisebox{-1pt}{\includegraphics[height=8pt]{figures/logos/qwen.png}}\,Qwen~3 8B & 161 & 98 & 0 & 1 \\
\raisebox{-1pt}{\includegraphics[height=8pt]{figures/logos/qwen.png}}\,Qwen~3 14B & 163 & 99 & 0 & 1 \\
\rowcolor{tablerow} \raisebox{-1pt}{\includegraphics[height=8pt]{figures/logos/qwen.png}}\,Qwen~2.5 7B & 153 & 93 & 0 & 1 \\
\raisebox{-1pt}{\includegraphics[height=8pt]{figures/logos/llama.png}}\,Llama~3.1 8B base & 91 & 55 & 0 & 1 \\
\rowcolor{tablerow} \raisebox{-1pt}{\includegraphics[height=8pt]{figures/logos/llama.png}}\,Llama~3.1 8B & 126 & 76 & 0 & 1 \\
\bottomrule
\end{tabular}}

\end{table}

\textbf{Which request is held out.} The 15-tool run holds out the third request of every
tool. Table~\ref{tab:fold_rotation} holds out each request in turn, building the tool vectors from the
other two, on the three representative models. The choice moves the last-layer count by up to
39 pairs on the original tool list and 19 on the renamed tool list, and the third request gives the
highest count on two of the three models. The table also repeats the third-request run with a
single opening brace in the answer prefix in place of the doubled one described in
Appendix~\ref{app:eval_protocol}; the count moves by between $-3$ and $+9$.

\begin{table}[htbp]
\centering
\caption{Full-name switches out of 210 at the last layer when request 0, 1 or 2 of each tool is held out and the other two build the tool vectors. q2 is the run accompanying the held-out depth comparison, and the q2 counts differ by one pair per model from Table~\ref{tab:generalization}. ``q2, \{'' repeats it with a single opening brace in the answer prefix.}
\label{tab:fold_rotation}
\small
\setlength{\tabcolsep}{7pt}
\resizebox{\ifdim\width>\columnwidth\columnwidth\else\width\fi}{!}{%
\begin{tabular}{@{}l rrrr rrrr@{}}
\toprule
& \multicolumn{4}{c}{\textsc{Original tool list}} & \multicolumn{4}{c}{\textsc{Renamed tool list}} \\
\cmidrule(lr){2-5}\cmidrule(lr){6-9}
\textsc{Model} & q0 & q1 & q2 & q2, \{ & q0 & q1 & q2 & q2, \{ \\
\midrule
\raisebox{-1pt}{\includegraphics[height=8pt]{figures/logos/gemma.png}}\,Gemma~3 4B & 80 & 104 & 119 & 120 & 168 & 186 & 187 & 195 \\
\rowcolor{tablerow} \raisebox{-1pt}{\includegraphics[height=8pt]{figures/logos/qwen.png}}\,Qwen~3 4B & 115 & 140 & 150 & 147 & 172 & 181 & 188 & 197 \\
\raisebox{-1pt}{\includegraphics[height=8pt]{figures/logos/llama.png}}\,Llama~3.1 8B & 102 & 129 & 125 & 131 & 156 & 169 & 165 & 169 \\
\bottomrule
\end{tabular}}

\end{table}

\textbf{Peak layer on other held-out requests.} The peak layers of Table~\ref{tab:heldout_directions} were chosen with request 2 held out. Table~\ref{tab:fold_peak} keeps those layers and holds out request 0 or 1 instead. The steering direction switches more pairs than the output direction in every cell.

\begin{table}[htbp]
\centering
\caption{Full-name switches out of 210 held-out ordered pairs at the peak layer L of Table~\ref{tab:heldout_directions}, on the original tool list, when request 0, 1 or 2 of each tool is held out.}
\label{tab:fold_peak}
\small
\setlength{\tabcolsep}{7pt}
\resizebox{\ifdim\width>\columnwidth\columnwidth\else\width\fi}{!}{%
\begin{tabular}{@{}l r rrr rrr@{}}
\toprule
& & \multicolumn{3}{c}{\textsc{Steering direction}} & \multicolumn{3}{c}{\textsc{Output direction}} \\
\cmidrule(lr){3-5}\cmidrule(lr){6-8}
\textsc{Model} & L & q0 & q1 & q2 & q0 & q1 & q2 \\
\midrule
\raisebox{-1pt}{\includegraphics[height=8pt]{figures/logos/gemma.png}}\,Gemma~3 4B & 20 & 83 & 87 & 92 & 0 & 5 & 2 \\
\rowcolor{tablerow} \raisebox{-1pt}{\includegraphics[height=8pt]{figures/logos/qwen.png}}\,Qwen~3 4B & 24 & 89 & 108 & 125 & 9 & 33 & 45 \\
\raisebox{-1pt}{\includegraphics[height=8pt]{figures/logos/llama.png}}\,Llama~3.1 8B & 16 & 36 & 56 & 53 & 0 & 4 & 6 \\
\bottomrule
\end{tabular}}

\end{table}

\textbf{Every tool's output direction removed.} Table~\ref{tab:orth_depth} uses 30 pairs, $\alpha{=}1.0$ and the prefix rule. Table~\ref{tab:orthall_heldout} repeats the removal on the 210 held-out pairs with full-name scoring at $\alpha{=}0.7$. Both columns of each layer come from the same run.

\begin{table}[htbp]
\centering
\caption{Full-name switches out of 210 held-out ordered pairs for the steering direction (\textsc{Steer}) and for the same direction with its part along every listed tool's output direction removed and rescaled to the original length (\textsc{Removed}), original tool list, $\alpha{=}0.7$.}
\label{tab:orthall_heldout}
\small
\setlength{\tabcolsep}{7pt}
\resizebox{\ifdim\width>\columnwidth\columnwidth\else\width\fi}{!}{%
\begin{tabular}{@{}l rrr rrr@{}}
\toprule
& \multicolumn{3}{c}{\textsc{Peak layer}} & \multicolumn{3}{c}{\textsc{Last layer}} \\
\cmidrule(lr){2-4}\cmidrule(lr){5-7}
\textsc{Model} & L & \textsc{Steer} & \textsc{Removed} & L & \textsc{Steer} & \textsc{Removed} \\
\midrule
\raisebox{-1pt}{\includegraphics[height=8pt]{figures/logos/gemma.png}}\,Gemma~3 4B & 20 & 91 & 91 & 33 & 120 & 1 \\
\rowcolor{tablerow} \raisebox{-1pt}{\includegraphics[height=8pt]{figures/logos/qwen.png}}\,Qwen~3 4B & 24 & 122 & 112 & 35 & 148 & 2 \\
\raisebox{-1pt}{\includegraphics[height=8pt]{figures/logos/llama.png}}\,Llama~3.1 8B & 16 & 54 & 52 & 31 & 126 & 0 \\
\bottomrule
\end{tabular}}

\end{table}

\textbf{Domain against tool on ToolBench.} The cross-domain ToolBench tool list has three
APIs per domain, and the three share their first token. Table~\ref{tab:toolbench_domain}
compares the steering direction with a direction built only from domain averages (the mean of a
domain's three tool vectors, target domain minus source domain) on the same 30 cross-domain pairs. The table adds 30 pairs whose source and target lie in one domain. The two directions
produce the same generation on 30, 30 and 28 of the 30 pairs and agree on every hit and
miss, so the tool-specific part of the direction changes nothing here. Within a domain, the tool
direction switches nothing, and the unsteered generation names the source in 28 to 30 of
30 pairs. These runs use an example request as the test request, as the ToolBench run in
Table~\ref{tab:generalization} does, and are not held out.
\subsection{The direction in the model's own state differences}
\label{app:direction_patch}
We test how much of the model's own tool-choice state lies along the steering direction. For each tool pair, two equal-length prompts request different tools.
We record their last-prompt-token residuals and call the difference $\delta$. We then
displace the clean residual by the full $\delta$, its projection onto
$\direction_{a \to b}$, its orthogonal remainder, or a random direction with the
projection's norm. Table~\ref{tab:direction_patch} reports the fraction of the two runs'
first-token logit difference recovered by each displacement.
The steering direction is built from example requests excluded from both evaluated prompts.
We exclude pairs whose original logit difference is less than one, leaving 16 to 24 pairs
per model.

\begin{table}[htbp]
\centering
\caption{ToolBench cross-domain steering, out of 30 pairs. \textsc{target} counts calls naming the target API, and \textsc{domain} calls naming any API of the target domain. \textsc{Within domain} uses 30 pairs inside one domain, steered and unsteered.}
\label{tab:toolbench_domain}
\small
\setlength{\tabcolsep}{7pt}
\resizebox{\ifdim\width>\columnwidth\columnwidth\else\width\fi}{!}{%
\begin{tabular}{@{}l rr rr rr@{}}
\toprule
& \multicolumn{2}{c}{\textsc{Steering direction}} & \multicolumn{2}{c}{\textsc{Domain direction}} & \multicolumn{2}{c}{\textsc{Within domain}} \\
\cmidrule(lr){2-3}\cmidrule(lr){4-5}\cmidrule(lr){6-7}
\textsc{Model} & target & domain & target & domain & steered & unsteered \\
\midrule
\raisebox{-1pt}{\includegraphics[height=8pt]{figures/logos/gemma.png}}\,Gemma~3 4B & 12 & 29 & 12 & 29 & 0 & 0 \\
\rowcolor{tablerow} \raisebox{-1pt}{\includegraphics[height=8pt]{figures/logos/qwen.png}}\,Qwen~3 4B & 11 & 30 & 11 & 30 & 0 & 0 \\
\raisebox{-1pt}{\includegraphics[height=8pt]{figures/logos/llama.png}}\,Llama~3.1 8B & 8 & 25 & 8 & 23 & 0 & 0 \\
\bottomrule
\end{tabular}}

\end{table}

\begin{table}[htbp]
\centering
\caption{Recovered share (\%) of the first-token logit difference between two requests that call different tools (the clean and corrupt runs), when the
clean run's last-prompt-token residual is displaced by $\delta$ (\textsc{Full}), by its component
along $\direction_{a \to b}$ (\textsc{Dir}), by the remainder (\textsc{Rem}), or by a random
direction scaled to the same length as that component (\textsc{Rand}). The last column is the share of $\delta$'s
length that lies along $\direction_{a \to b}$. Rows follow each model's own depth grid; the three models of
Figure~\ref{fig:depth_3fam} are the first, third and fifth blocks. Each row is a mean over the pairs.}
\small
\setlength{\tabcolsep}{7pt}
\rowcolors{2}{tablerow}{white}
\resizebox{\ifdim\width>\columnwidth\columnwidth\else\width\fi}{!}{%
\begin{tabular}{@{}l rr rrr r r@{}}
\toprule
\textsc{Model} & L & \% & \textsc{Full} & \textsc{Dir} & \textsc{Rem} & \textsc{Rand} & $|d_\parallel|/|\delta|$ \\
\midrule
\raisebox{-1pt}{\includegraphics[height=8pt]{figures/logos/gemma.png}}\,Gemma~3 4B & 13 & 39 & -2 & -2 & 1 & 2.5 & 65 \\
  & 16 & 48 & 31 & 11 & 15 & 1.4 & 46 \\
  & 18 & 55 & 32 & 10 & 11 & -0.1 & 56 \\
  & 20 & 61 & 76 & 35 & 20 & -0.8 & 67 \\
  & 21 & 64 & 79 & 35 & 26 & -2.0 & 66 \\
  & 23 & 70 & 87 & 51 & 27 & 0.2 & 76 \\
  & 26 & 79 & 90 & 59 & 30 & -1.2 & 79 \\
  & 33 & 100 & 100 & 48 & 48 & -3.1 & 65 \\
\midrule
\raisebox{-1pt}{\includegraphics[height=8pt]{figures/logos/gemma.png}}\,Gemma~3 12B & 19 & 40 & 1 & 0 & 1 & 0.1 & 61 \\
  & 24 & 51 & 11 & 4 & 6 & 0.3 & 57 \\
  & 26 & 55 & 65 & 32 & 13 & 0.7 & 77 \\
  & 28 & 60 & 79 & 50 & 22 & 1.9 & 74 \\
  & 31 & 66 & 79 & 56 & 18 & -5.0 & 81 \\
  & 33 & 70 & 84 & 58 & 20 & -5.3 & 80 \\
  & 38 & 81 & 87 & 73 & 15 & -1.2 & 86 \\
  & 47 & 100 & 100 & 70 & 30 & 4.9 & 86 \\
\midrule
\raisebox{-1pt}{\includegraphics[height=8pt]{figures/logos/qwen.png}}\,Qwen~3 4B & 14 & 40 & 0 & 0 & 0 & 0.2 & 46 \\
  & 18 & 51 & 2 & 2 & 1 & 0.3 & 56 \\
  & 19 & 54 & 7 & 2 & 2 & -0.1 & 63 \\
  & 21 & 60 & 14 & 6 & 5 & 0.1 & 67 \\
  & 23 & 66 & 69 & 37 & 28 & 0.2 & 65 \\
  & 24 & 69 & 70 & 52 & 24 & 0.7 & 77 \\
  & 28 & 80 & 83 & 52 & 27 & -0.7 & 75 \\
  & 35 & 100 & 100 & 66 & 33 & 5.1 & 76 \\
\midrule
\raisebox{-1pt}{\includegraphics[height=8pt]{figures/logos/qwen.png}}\,Qwen~3 8B & 14 & 40 & 0 & 0 & 0 & 0.2 & 48 \\
  & 18 & 51 & 4 & 2 & 2 & 0.7 & 58 \\
  & 19 & 54 & 7 & 2 & 3 & -0.1 & 64 \\
  & 21 & 60 & 23 & 8 & 5 & 0.3 & 66 \\
  & 23 & 66 & 78 & 38 & 24 & -0.3 & 67 \\
  & 24 & 69 & 78 & 61 & 16 & -2.8 & 79 \\
  & 28 & 80 & 89 & 66 & 21 & -3.2 & 78 \\
  & 35 & 100 & 100 & 73 & 27 & 8.8 & 79 \\
\midrule
\raisebox{-1pt}{\includegraphics[height=8pt]{figures/logos/llama.png}}\,Llama~3.1 8B & 12 & 39 & 12 & 10 & 3 & -0.2 & 68 \\
  & 16 & 52 & 68 & 42 & 27 & 0.2 & 69 \\
  & 17 & 55 & 70 & 51 & 29 & -0.7 & 75 \\
  & 19 & 61 & 72 & 50 & 27 & -1.3 & 74 \\
  & 20 & 65 & 74 & 51 & 26 & 0.8 & 75 \\
  & 22 & 71 & 83 & 61 & 27 & -0.1 & 78 \\
  & 25 & 81 & 85 & 64 & 28 & -1.9 & 81 \\
  & 31 & 100 & 100 & 72 & 29 & 5.3 & 80 \\
\bottomrule
\end{tabular}}

\label{tab:direction_patch}
\end{table}

The projection recovers part of the model's logit difference, but the remainder also has
a substantial effect (Table~\ref{tab:direction_patch}). We summarise the 29 layers across
five models where the full displacement recovers at least $0.2$ of the original
logit difference. Those layers are selected by an outcome measured on the same pairs, so
the range below describes where the full displacement already works rather than an
independently chosen set of layers. Relative to that full-displacement effect, projection recovery ranges from 30--84\% and remainder recovery from 17--49\%. The two ranges describe different layers and should be read separately. The matched random direction changes
the original logit difference by at most 8.8\% in either direction and has a smaller
absolute effect than the projection at every eligible layer.

The projection retains 46--86\% of $\delta$'s norm across those layers. To account for
displacement size, we divide each pair's recovered logit fraction by its component's
fraction of $\delta$'s norm, then average over pairs for each component. This normalises
the effect rather than matching the displacement in the experiment, and only the random
control is injected at a matched length. The projection's
normalised effect exceeds the remainder's at all 29 layers, by at least 1.5 times at
23 layers but by less than 10\% at two. Thus the direction contributes disproportionately
to recovery without explaining the entire decision.

\subsection{Per-layer residual patching}
\label{app:attribution_method}

This test measures the effect of replacing a tool-request state with a non-tool-request state. At each layer, we replace the last-prompt-token residual from ``What is the weather in Tokyo?'' with that from ``Hello, how are you today?'', then measure the change in the logit of the original run's top token. It contrasts tool calling with small talk, whereas the component patching in Section~\ref{sec:circuit} compares token-aligned requests for different tools. The experiments therefore address different contrasts and need not show their largest effects at the same depth.

Whole-residual replacement also replaces information accumulated at earlier layers, so it cannot say where a computation was first performed. On Gemma~3 4B, the average effect over five requests rises from 3.4 at L15 to 10.7 at L33 without a local maximum; Qwen~3 4B and Llama~3.1 8B also have their largest average effects at the final layer. Individual requests differ: two of Gemma's five requests peak at L15 and L16, while three peak at L33. Every row of Table~\ref{tab:attribution_summary} except the two Gemma~3 4B measurements uses a single request pair. We use the instruction-tuned/base ratio to compare responses to the same patch at the same layer, and do not interpret the reported maximum as the location of a mechanism.

Eight rows lacked stored outputs when the artefact was assembled and were rerun on different hardware. Five rerun effects differ from the reported values by at most 0.05 and a sixth by at most 0.25. The largest differences are 0.4: Gemma~3 270M gives 2.0 rather than 1.6, and Gemma~3 1B gives 7.0 rather than 7.4. The layer with the largest effect shifts by up to two layers. The table retains the original reported values; the supplementary archive contains a stored output for every row. These reruns document, rather than remove, the discrepancy with the original values.

\begin{table}[htbp]
\centering
\caption{Last-prompt-token residual replacement, comparing a tool request with small talk. Importance is the resulting logit change; Peak is the layer with the largest measured change, and Depth is its relative position. This layer does not locate where tool choice is computed (Appendix~\ref{app:attribution_method}). IT/Base compares the models at the reported IT layer; Gemma~3 1B base has its own maximum at L15, rather than the IT model's L25. Dashes: Qwen base models were not run in this comparison. $^\dag$Qwen~1.7B reaches its maximum well before the final layer.}
\label{tab:attribution_summary}

\small
\setlength{\tabcolsep}{7pt}
\begin{tabular}{@{}l c r rrr@{}}
\toprule
\rowcolor{white}
& & & \multicolumn{3}{c}{\textsc{Importance}} \\
\cmidrule(l){4-6}
\rowcolor{white}
& Peak & Depth & IT & Base & IT/Base \\
\midrule
\multicolumn{6}{@{}l}{\raisebox{-1pt}{\includegraphics[height=8pt]{figures/logos/gemma.png}}\,\textbf{Gemma~3}} \\
\rowcolor{tablerow} \quad 270M & L17 & 94\% & 1.6  & 0.4 & 4.0$\times$ \\
\quad 1B   & L25 & 96\% & 7.4  & 0.6 & 12.3$\times$ \\
\rowcolor{tablerow} \quad 4B   & L33 & 97\% & 10.7 & 3.4 & 3.2$\times$ \\
\quad 12B  & L44 & 92\% & 8.1  & 1.6 & 5.1$\times$ \\
\rowcolor{tablerow} \quad 27B  & L61 & 98\% & 14.2 & 1.8 & 7.9$\times$ \\
\midrule
\multicolumn{6}{@{}l}{\raisebox{-1pt}{\includegraphics[height=8pt]{figures/logos/qwen.png}}\,\textbf{Qwen~3 / 2.5}} \\
\rowcolor{tablerow} \quad 0.6B  & L25 & 89\% & 5.9  & \multicolumn{2}{c}{---} \\
\quad 1.7B$^\dag$  & L20 & 71\% & 10.1 & \multicolumn{2}{c}{---} \\
\rowcolor{tablerow} \quad 4B    & L35 & 97\% & 12.1 & \multicolumn{2}{c}{---} \\
\quad 8B    & L35 & 97\% & 10.3 & \multicolumn{2}{c}{---} \\
\rowcolor{tablerow} \quad 14B   & L39 & 98\% & 11.8 & \multicolumn{2}{c}{---} \\
\quad Qwen~2.5 7B & L27 & 96\% & 10.3 & \multicolumn{2}{c}{---} \\
\midrule
\multicolumn{6}{@{}l}{\raisebox{-1pt}{\includegraphics[height=8pt]{figures/logos/llama.png}}\,\textbf{Llama~3.1}} \\
\rowcolor{tablerow} \quad 8B & L31 & 97\% & 7.4  & 1.6 & 4.6$\times$ \\
\bottomrule
\end{tabular}

\end{table}

\section{Component analysis}

\subsection{Component-level activation patching}
\label{app:patching_details}

Activation patching measures how individual components affect tool choice. We run requests
for tools A and B, then replace one component's output in the A run with its output from
the B run. The change in the model's score for A measures that component's effect.

We patch every attention head and MLP layer, averaging over 10 tool pairs. On Gemma~3 4B
and Qwen~3 4B, fewer than 5\% of heads account for a disproportionate share of the total
attention effect. Summed effects increase with depth. The middle third of layers accounts for 33--37\%, less than the final third. This aggregate differs from the layer
of the strongest individual head. Raw attention
totals account for 80--88\%, but heads greatly outnumber MLP layers. Per component, an average MLP has 1.94$\times$ the effect of an average head on Gemma and 4.19$\times$ on Qwen. Table~\ref{tab:per_component_norm}
also compares equally sized sets of the strongest components.

The component-level conclusion differs by model. Gemma's strongest heads, including
L17 H0 and H1, exceed its strongest MLPs (Table~\ref{tab:per_component_norm}).
Qwen has similar top-head and top-MLP effects, despite having $32\times$ more heads than
MLP layers. Qwen's larger raw attention total therefore reflects the much larger number of heads.

\subsection{Position-resolved component patching}
\label{app:position_patching}
Early-layer effects occur at request positions, while the largest later component effects
occur in attention at the last prompt token (Table~\ref{tab:position_patching}). This
experiment reports recovery of the clean-vs-corrupt logit gap, not the normalised
component shares of Section~\ref{sec:circuit}.

We replace each layer's attention or MLP output in a clean run with its output from a
corrupt run. Prompt pairs have equal token length and differ only in the user request. We
patch either the \emph{request} positions, defined as all differing tokens, or the
\emph{final} prompt token. Recovery is the fraction of the original first-token logit
difference between the two tools, reported as mean$\pm$std over 16 pairs per model.

\begin{table}[htbp]
\centering
\caption{Logit-gap recovery after replacing one component with its output from a matched corrupt prompt (16 equal-length pairs per model; mean$\pm$std for the principal cells). Early request-position effects precede mid-layer attention effects at the last prompt token. Late MLP patches have negative recovery, explained below. Rebuilt rows repeat the last-layer MLP patch with a new script on the same 16 pairs. $^{\ddagger}$Qwen L0 request-position patches approximate replacing the request representation itself; these values are upper bounds rather than component-specific effects. Gemma L0 provides the more selective early-layer comparison between MLP and attention.}
\label{tab:position_patching}

\small
\setlength{\tabcolsep}{6pt}
\resizebox{\ifdim\width>\columnwidth\columnwidth\else\width\fi}{!}{\begin{tabular}{ll rr rr}
\toprule
& & \multicolumn{2}{c}{\textsc{Request pos.}} & \multicolumn{2}{c}{\textsc{Final tok.}} \\
\textsc{Model} & \textsc{L} & \textsc{attn} & \textsc{mlp} & \textsc{attn} & \textsc{mlp} \\
\midrule
\raisebox{-1pt}{\includegraphics[height=8pt]{figures/logos/gemma.png}}\,Gemma~3 4B & 0  & 0.01 & \textbf{0.34}\tiny{$\pm$0.26} & 0.00 & 0.00 \\
\rowcolor{tablerow} \raisebox{-1pt}{\includegraphics[height=8pt]{figures/logos/gemma.png}}\,Gemma~3 4B & 17 & 0.01 & 0.00 & 0.11\tiny{$\pm$0.05} & 0.01 \\
\raisebox{-1pt}{\includegraphics[height=8pt]{figures/logos/gemma.png}}\,Gemma~3 4B & 23 & 0.00 & 0.01 & \textbf{0.44}\tiny{$\pm$0.08} & 0.02 \\
\rowcolor{tablerow} \raisebox{-1pt}{\includegraphics[height=8pt]{figures/logos/gemma.png}}\,Gemma~3 4B & 33 & 0.00 & 0.00 & 0.04 & $-$0.24\tiny{$\pm$0.12} \\
\rowcolor{tablerow} \quad rebuilt & 33 & -- & -- & -- & $-$0.24\tiny{$\pm$0.12} \\
\midrule
\raisebox{-1pt}{\includegraphics[height=8pt]{figures/logos/qwen.png}}\,Qwen~3 4B & 0  & 0.99$^{\ddagger}$ & 0.99$^{\ddagger}$ & 0.00 & 0.00 \\
\rowcolor{tablerow} \raisebox{-1pt}{\includegraphics[height=8pt]{figures/logos/qwen.png}}\,Qwen~3 4B & 24 & 0.00 & 0.00 & \textbf{0.41}\tiny{$\pm$0.15} & 0.03 \\
\raisebox{-1pt}{\includegraphics[height=8pt]{figures/logos/qwen.png}}\,Qwen~3 4B & 35 & 0.00 & 0.00 & 0.02 & $-$0.24\tiny{$\pm$0.10} \\
\quad rebuilt & 35 & -- & -- & -- & $-$0.25\tiny{$\pm$0.10} \\
\bottomrule
\end{tabular}}

\end{table}

\textbf{Why the last MLP gives negative recovery.} In both models the last layer's MLP lowers the score of the tool the model is about to name. Taken on its own, the MLP output narrows the gap between the first-token scores of the winning and the losing tool in 27 of the 32 unpatched runs on Gemma~3 4B and in 30 of 32 on Qwen~3 4B. The corrupt run's MLP output therefore lowers the corrupt tool. In the clean run that tool is the loser, so copying the output in widens the gap instead of closing it. Projecting the change in the MLP output onto the difference of the two tools' output directions accounts for the effect. The projection has the sign of the measured change on all 16 pairs of each model, and its median size is 0.98 of the measured change on Gemma and 1.00 on Qwen, so the effect runs through the output direction rather than through the final normalization. We rebuilt this patch from the archived pairs, since the script that produced the table is not in our repository. The rebuilt rows in Table~\ref{tab:position_patching} match the archived ones. Late neurons that lower the likeliest token have been described before \citep{stolfo2024confidence}.

\subsection{Full SAE feature analysis}
\label{app:sae_full}
Sparse autoencoders (SAEs) decompose activations into sparsely active features \citep{lieberum2024gemmascope,rajamanoharan2024jumping}. We count features associated with one tool and features shared across tools. These counts predict steering poorly. The 270M base model has more tool-specific features than its instruction-tuned counterpart, yet steering remains ineffective. The relative count also varies non-monotonically with model size. At the matched 1B layer, instruction tuning increases tool-specific counts and reduces shared counts, while the 4B comparison is nearly unchanged. These observations describe the chosen SAEs.

\begin{figure}[htbp]
\centering
\includegraphics[width=0.85\linewidth]{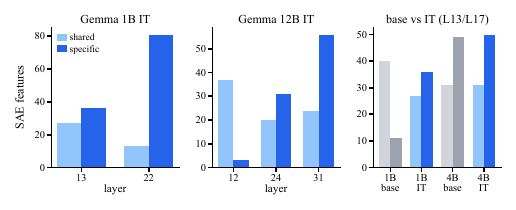}
\caption{SAE feature counts. Light bars: features shared across tools; dark bars: features active for one tool only. \textbf{Left and centre:} on Gemma~3 1B and 12B, tool-specific counts generally rise with depth while shared counts decrease or remain lower. \textbf{Right:} instruction tuning increases the tool-specific count at the matched 1B layer but leaves it nearly unchanged at 4B. The pattern is not uniform across models. Table~\ref{tab:sae_full} gives all per-layer counts.}
\label{fig:sae_sharpening}
\end{figure}

Table~\ref{tab:sae_full} reports all included model and layer counts; Figure~\ref{fig:sae_sharpening} shows selected comparisons. Counts from different SAEs are not directly comparable.

\begin{table}[htbp]
\centering
\caption{SAE feature counts for every Gemma~3 model and layer we ran, from Gemma Scope at 16K width. \textsc{Active} is the number of features with positive activation for a single tool, averaged over the 15 tools. \textsc{Shared} counts the features that activate above 0.1 for every tool and \textsc{Specific} those that activate above 0.1 for exactly one, both taken over the union across tools. The three are measured differently, so \textsc{Specific} is not a subset of \textsc{Active} and exceeds it in nine of the rows. Counts are only comparable within one SAE, so the single Llama row is marked: $^*$Goodfire (\href{https://goodfire.ai}{\nolinkurl{goodfire.ai}}) at 65K. Runs against other SAEs (andyrdt for Qwen~2.5 7B and for Llama, EleutherAI at 32K width for Llama base) are not shown here for the same reason.}
\label{tab:sae_full}

\small
\setlength{\tabcolsep}{2pt}
\rowcolors{2}{tablerow}{white}
\resizebox{\ifdim\width>\columnwidth\columnwidth\else\width\fi}{!}{\begin{tabular}{@{}l c r r r r r@{}}
\toprule
\rowcolor{white}
& & & \multicolumn{3}{c}{\textsc{Feature counts}} & \textsc{Sp/Sh} \\
\cmidrule(lr){4-6}
\rowcolor{white}
Model & Layer & Depth & Active & Shared & Specific & Ratio \\
\midrule
\raisebox{-1pt}{\includegraphics[height=8pt]{figures/logos/gemma.png}}\,Gemma~3 270M-IT & L5 & 28\% & 34 & 30 & 0 & 0.0 \\
\raisebox{-1pt}{\includegraphics[height=8pt]{figures/logos/gemma.png}}\,Gemma~3 270M-IT & L9 & 50\% & 51 & 40 & 9 & 0.2 \\
\raisebox{-1pt}{\includegraphics[height=8pt]{figures/logos/gemma.png}}\,Gemma~3 270M-IT & L12 & 67\% & 104 & 86 & 9 & 0.1 \\
\raisebox{-1pt}{\includegraphics[height=8pt]{figures/logos/gemma.png}}\,Gemma~3 270M-IT & L15 & 83\% & 67 & 51 & 16 & 0.3 \\
\raisebox{-1pt}{\includegraphics[height=8pt]{figures/logos/gemma.png}}\,Gemma~3 270M-base & L5 & 28\% & 45 & 41 & 1 & 0.0 \\
\raisebox{-1pt}{\includegraphics[height=8pt]{figures/logos/gemma.png}}\,Gemma~3 270M-base & L9 & 50\% & 48 & 36 & 9 & 0.3 \\
\raisebox{-1pt}{\includegraphics[height=8pt]{figures/logos/gemma.png}}\,Gemma~3 270M-base & L12 & 67\% & 91 & 49 & 49 & 1.0 \\
\raisebox{-1pt}{\includegraphics[height=8pt]{figures/logos/gemma.png}}\,Gemma~3 270M-base & L15 & 83\% & 81 & 32 & 77 & 2.4 \\
\midrule
\raisebox{-1pt}{\includegraphics[height=8pt]{figures/logos/gemma.png}}\,Gemma~3 1B-IT & L7 & 27\% & 48 & 41 & 3 & 0.1 \\
\raisebox{-1pt}{\includegraphics[height=8pt]{figures/logos/gemma.png}}\,Gemma~3 1B-IT & L13 & 50\% & 56 & 27 & 36 & 1.3 \\
\raisebox{-1pt}{\includegraphics[height=8pt]{figures/logos/gemma.png}}\,Gemma~3 1B-IT & L17 & 65\% & 67 & 12 & 125 & 10.4 \\
\raisebox{-1pt}{\includegraphics[height=8pt]{figures/logos/gemma.png}}\,Gemma~3 1B-IT & L22 & 85\% & 51 & 13 & 81 & 6.2 \\
\raisebox{-1pt}{\includegraphics[height=8pt]{figures/logos/gemma.png}}\,Gemma~3 1B-base & L7 & 27\% & 38 & 33 & 3 & 0.1 \\
\raisebox{-1pt}{\includegraphics[height=8pt]{figures/logos/gemma.png}}\,Gemma~3 1B-base & L13 & 50\% & 55 & 40 & 11 & 0.3 \\
\raisebox{-1pt}{\includegraphics[height=8pt]{figures/logos/gemma.png}}\,Gemma~3 1B-base & L17 & 65\% & 48 & 35 & 19 & 0.5 \\
\raisebox{-1pt}{\includegraphics[height=8pt]{figures/logos/gemma.png}}\,Gemma~3 1B-base & L22 & 85\% & 69 & 33 & 43 & 1.3 \\
\midrule
\raisebox{-1pt}{\includegraphics[height=8pt]{figures/logos/gemma.png}}\,Gemma~3 4B-IT & L9 & 26\% & 41 & 37 & 3 & 0.1 \\
\raisebox{-1pt}{\includegraphics[height=8pt]{figures/logos/gemma.png}}\,Gemma~3 4B-IT & L17 & 50\% & 66 & 31 & 50 & 1.6 \\
\raisebox{-1pt}{\includegraphics[height=8pt]{figures/logos/gemma.png}}\,Gemma~3 4B-IT & L22 & 65\% & 60 & 31 & 36 & 1.2 \\
\raisebox{-1pt}{\includegraphics[height=8pt]{figures/logos/gemma.png}}\,Gemma~3 4B-IT & L29 & 85\% & 59 & 16 & 80 & 5.0 \\
\raisebox{-1pt}{\includegraphics[height=8pt]{figures/logos/gemma.png}}\,Gemma~3 4B-base & L17 & 50\% & 68 & 31 & 49 & 1.6 \\
\raisebox{-1pt}{\includegraphics[height=8pt]{figures/logos/gemma.png}}\,Gemma~3 4B-base & L22 & 65\% & 80 & 35 & 73 & 2.1 \\
\midrule
\raisebox{-1pt}{\includegraphics[height=8pt]{figures/logos/gemma.png}}\,Gemma~3 12B-IT & L12 & 25\% & 44 & 37 & 3 & 0.1 \\
\raisebox{-1pt}{\includegraphics[height=8pt]{figures/logos/gemma.png}}\,Gemma~3 12B-IT & L24 & 50\% & 48 & 20 & 31 & 1.6 \\
\raisebox{-1pt}{\includegraphics[height=8pt]{figures/logos/gemma.png}}\,Gemma~3 12B-IT & L31 & 65\% & 52 & 24 & 56 & 2.3 \\
\raisebox{-1pt}{\includegraphics[height=8pt]{figures/logos/gemma.png}}\,Gemma~3 12B-IT & L41 & 85\% & 49 & 19 & 78 & 4.1 \\
\raisebox{-1pt}{\includegraphics[height=8pt]{figures/logos/gemma.png}}\,Gemma~3 12B-base & L12 & 25\% & 60 & 38 & 10 & 0.3 \\
\raisebox{-1pt}{\includegraphics[height=8pt]{figures/logos/gemma.png}}\,Gemma~3 12B-base & L24 & 50\% & 69 & 39 & 28 & 0.7 \\
\raisebox{-1pt}{\includegraphics[height=8pt]{figures/logos/gemma.png}}\,Gemma~3 12B-base & L31 & 65\% & 77 & 30 & 62 & 2.1 \\
\raisebox{-1pt}{\includegraphics[height=8pt]{figures/logos/gemma.png}}\,Gemma~3 12B-base & L41 & 85\% & 78 & 33 & 78 & 2.4 \\
\midrule
\raisebox{-1pt}{\includegraphics[height=8pt]{figures/logos/gemma.png}}\,Gemma~3 27B-IT & L16 & 26\% & 47 & 43 & 4 & 0.1 \\
\raisebox{-1pt}{\includegraphics[height=8pt]{figures/logos/gemma.png}}\,Gemma~3 27B-IT & L31 & 50\% & 53 & 27 & 45 & 1.7 \\
\raisebox{-1pt}{\includegraphics[height=8pt]{figures/logos/gemma.png}}\,Gemma~3 27B-IT & L40 & 65\% & 46 & 18 & 60 & 3.3 \\
\raisebox{-1pt}{\includegraphics[height=8pt]{figures/logos/gemma.png}}\,Gemma~3 27B-IT & L53 & 85\% & 51 & 18 & 75 & 4.2 \\
\raisebox{-1pt}{\includegraphics[height=8pt]{figures/logos/gemma.png}}\,Gemma~3 27B-base & L16 & 26\% & 93 & 69 & 12 & 0.2 \\
\raisebox{-1pt}{\includegraphics[height=8pt]{figures/logos/gemma.png}}\,Gemma~3 27B-base & L31 & 50\% & 70 & 33 & 62 & 1.9 \\
\raisebox{-1pt}{\includegraphics[height=8pt]{figures/logos/gemma.png}}\,Gemma~3 27B-base & L40 & 65\% & 79 & 27 & 72 & 2.7 \\
\raisebox{-1pt}{\includegraphics[height=8pt]{figures/logos/gemma.png}}\,Gemma~3 27B-base & L53 & 85\% & 77 & 31 & 79 & 2.5 \\
\midrule
\raisebox{-1pt}{\includegraphics[height=8pt]{figures/logos/llama.png}}\,Llama~3.1 8B-IT$^*$ & L19 & 59\% & 85 & 14 & 95 & 6.8 \\
\bottomrule
\end{tabular}}
\rowcolors{1}{}{}

\end{table}

Figure~\ref{fig:sae_sharpening} separates features shared across tools from those active for only one tool, using the within-SAE counts above.

\subsection{Cross-layer divergence and attention}
\label{app:divergence}

\begin{figure}[htbp]
\centering
\begin{minipage}[t]{\linewidth}
\centering
\includegraphics[width=\linewidth]{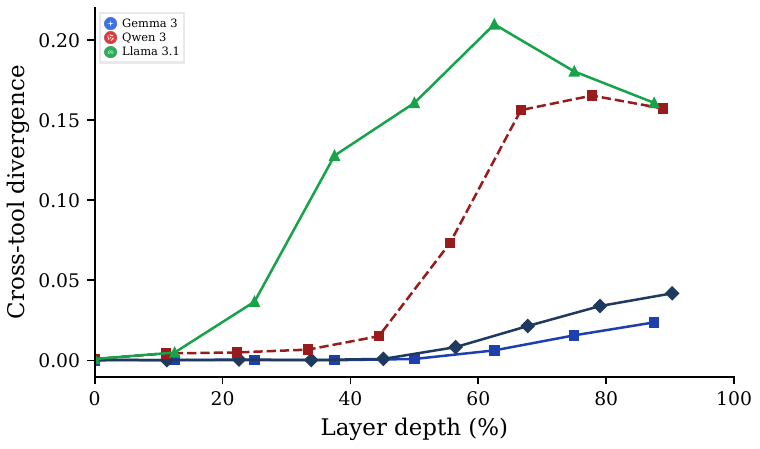}
\caption{Cross-tool divergence across layers for four models: Gemma~3 12B and 27B (blue), Qwen~3 8B (red), and Llama~3.1 8B (green). Both Gemma curves continue rising at the last sampled layer; Qwen and Llama peak earlier, with Llama peaking near 62\% of depth. Absolute levels are not comparable across families.}
\label{fig:divergence}
\end{minipage}

\begin{minipage}[t]{\linewidth}
\centering
\includegraphics[width=0.85\linewidth]{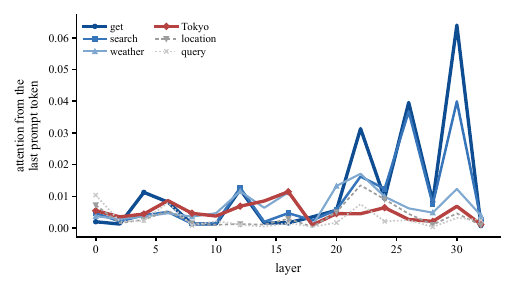}
\caption{Attention from the last prompt token to named prompt tokens on Gemma~3 4B, every second layer. The tool-name tokens spike late and in bursts (\texttt{get} reaches 0.064 at L30, \texttt{search} 0.040), while the request entity \texttt{Tokyo} does not follow them: it peaks early at L16 (0.012) and is back to 0.001 by the last layer. The late rise concerns tool-name tokens rather than the request entity; these are layer-level attention measurements.}
\label{fig:attention}
\end{minipage}
\end{figure}

Figures~\ref{fig:divergence} and \ref{fig:attention} trace how tool representations separate
across layers and how the last prompt token attends to tool names and request entities.
Figure~\ref{fig:combined_pca} shows low-dimensional structure within the synthetic,
$\tau$-bench, and ToolBench tool lists.

\section{Alternative explanations}

\subsection{Within-topic probe details}
\label{app:within_topic_probe}

Tools remain distinguishable when their requests concern the same domain.
The steering analysis measures an intervention; the probes measure information available
to a trained readout.

\textbf{Steering inside the same domain.} Eleven of the fourteen airline tools are
airline-specific; three are generic (\texttt{think}, \texttt{calculate}, and
\texttt{transfer\_to\_human\_agents}). We restrict the 50 sampled pairs to the 25 whose
source and target are both airline-specific. Under the original prefix rule, switch
rates are 88\% on Gemma~3 4B and 76\% on Qwen~3 4B, versus 92\% and 84\% over all pairs.
Full-name rates are 52\% and 44\%, versus 70\% and 60\% over all pairs. The tool list contains two \texttt{get\_}, two \texttt{search\_} and three \texttt{update\_} tools, so shared prefixes can count as successes without identifying the requested API. Steering therefore
persists within this domain, but at lower rates under complete-name scoring. Pair-level
records are available for these two models, not the other nine.

\textbf{$\tau$-bench airline readout.} The 14-tool airline environment includes
distinct booking, cancellation, update, search, and support actions. From real multi-turn
trajectories in the matched-context run, we collect residuals immediately before tool-name
generation. We fit multinomial logistic regression with an 80/20 stratified split,
$\ell_2$ regularization $C{=}1$, and StandardScaler features. Keeping only tools with at
least 10 calls per model leaves $K$ between 5 and 9.

\begin{table}[htbp]
\centering
\caption{Airline probe accuracy for tools called $\geq 10$ times per model. \textsc{vs chance}: top-1 divided by uniform chance. Always guessing the most frequent tool would score at most 27--57\%. Training-label permutations yield 8--25\%; matched-shape Gaussian features yield 18--42\%. The probe exceeds the Gaussian control by 29--57 percentage points on every model. Small samples and high-dimensional features limit the precision of these estimates.}
\label{tab:within_topic_taubench}

\small
\setlength{\tabcolsep}{5pt}
\resizebox{\ifdim\width>\columnwidth\columnwidth\else\width\fi}{!}{\begin{tabular}{l c c c c c}
\toprule
\textsc{Model} & $K$ & \textsc{Chance} & \textsc{Top-1} & \textsc{Top-5} & \textsc{vs chance} \\
\midrule
Gemma~3 4B IT     & 8 & 12.5\% & 61\% & 96\% & 4.9$\times$ \\
\rowcolor{tablerow} Gemma~3 12B IT    & 9 & 11.1\% & 66\% & 97\% & 5.9$\times$ \\
Qwen~3 4B         & 5 & 20.0\% & 71\% & 100\% & 3.5$\times$ \\
\rowcolor{tablerow} Qwen~3 8B         & 6 & 16.7\% & 73\% & 100\% & 4.4$\times$ \\
Qwen~3 14B        & 5 & 20.0\% & 89\% & 100\% & 4.5$\times$ \\
\bottomrule
\end{tabular}}

\end{table}

\textbf{BFCL Movies readout.} The Movies cluster of BFCL \texttt{live\_multiple}
contains three tools: \texttt{Movies\_1\_BuyMovieTickets}, \texttt{Movies\_1\_FindMovies},
and \texttt{Movies\_3\_FindMovies}. With 10--20 real user requests per tool, the same probe
reaches 100\% on Gemma~3 4B / 12B, Qwen~3 4B, and Llama~3.1 8B. However, the test set
contains only 13 examples. Gaussian-feature accuracy is 38--46\% against 33\% uniform
chance, leaving a probe advantage of 54--62 percentage points. We treat this as a small
consistency check; Table~\ref{tab:within_topic_taubench} is the primary within-topic test.

\textbf{Uniform chance and class imbalance.} The \textsc{Chance} column is $1/K$;
\textsc{vs chance} divides top-1 accuracy by that uniform rate. Airline calls are
imbalanced, so always predicting the most frequent tool can exceed $1/K$ without using
activations. In two stored trajectory pools with tool sets matching a probe run,
\texttt{get\_reservation\_details} accounts for 25.3\% of 158 calls and 56.5\% of 207,
versus uniform rates of 11.1\% and 20.0\%. These pools are not the probes' own train/test splits, whose per-class counts were not archived. The per-tool test accuracies were archived, and their denominators bound the per-tool test counts. Under every count assignment consistent with them, the most frequent tool makes up at most 27--57\% of a test split, and the probe's top-1 is at least 32 points above that on every model. The shuffle-label control preserves class proportions while removing their association with activations and is the more relevant null comparison.

\textbf{Cross-permutation control.} To rule out readout that depends on a particular name string, we replay each trajectory with two anonymous tool-to-name mappings. Under
\textbf{P1}, the 14 airline tools in alphabetical order map to \texttt{act\_A}--\texttt{act\_N}.
Under \textbf{P2}, the mapping shifts cyclically by seven, assigning every tool a different
tag. We rename tools in both the prompt schema and dialog history, while retaining
descriptions and parameter schemas verbatim. For each retained sample, we capture the
probe-position residual under both mappings. A shared train/test split supports four
probes: within-P1, within-P2, P1$\to$P2, and P2$\to$P1.

\begin{table}[htbp]
\centering
\caption{Cross-permutation probe accuracy. \textsc{wP1} and \textsc{wP2} are within-permutation probes (train and test on the same permutation); P1$\to$P2 and P2$\to$P1 are transfer probes (train on one permutation, test on the other). Mean transfer accuracy exceeds mean within-permutation accuracy by $0.011$ across the five models, and the gap is at most $0.04$ in either direction on any single model, well within per-model binomial standard error ($\pm 0.06$--$0.08$, $n_{\text{test}}\in[24,59]$).}
\label{tab:within_topic_xperm}

\small
\setlength{\tabcolsep}{7pt}
\begin{tabular}{@{}l c c c c c@{}}
\toprule
\textsc{Model} & $K$ & \textsc{wP1} & \textsc{wP2} & \textsc{P1$\to$P2} & \textsc{P2$\to$P1} \\
\midrule
\raisebox{-1pt}{\includegraphics[height=8pt]{figures/logos/gemma.png}}\,Gemma~3 4B IT     & 8 & 0.51 & 0.55 & 0.57 & 0.57 \\
\rowcolor{tablerow} \raisebox{-1pt}{\includegraphics[height=8pt]{figures/logos/gemma.png}}\,Gemma~3 12B IT    & 9 & 0.73 & 0.71 & 0.73 & 0.76 \\
\raisebox{-1pt}{\includegraphics[height=8pt]{figures/logos/qwen.png}}\,Qwen~3 4B         & 5 & 0.79 & 0.75 & 0.75 & 0.75 \\
\rowcolor{tablerow} \raisebox{-1pt}{\includegraphics[height=8pt]{figures/logos/qwen.png}}\,Qwen~3 8B         & 6 & 0.78 & 0.78 & 0.80 & 0.78 \\
\raisebox{-1pt}{\includegraphics[height=8pt]{figures/logos/qwen.png}}\,Qwen~3 14B        & 5 & 0.86 & 0.89 & 0.86 & 0.89 \\
\bottomrule
\end{tabular}

\end{table}

\textbf{Interpretation.} Within-domain accuracy and transfer between anonymous
name mappings argue against readout based only on a broad topic or a fixed surface tag
(Table~\ref{tab:within_topic_xperm}). They support a tool-identity signal that transfers
between the two tested mappings. These probes remain correlational.

\subsection{PCA geometry of the tool subspace}
\label{app:pca}

\subsubsection{Cumulative variance, one model per family}
\label{sec:subspace}
\begin{figure}[htbp]
\centering
\includegraphics[width=0.85\linewidth]{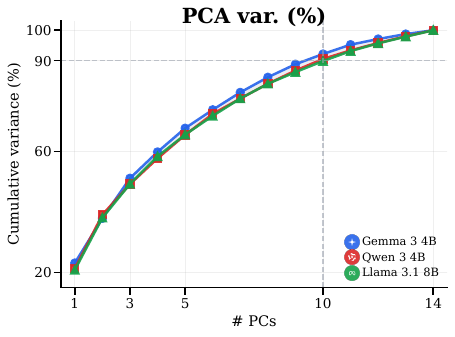}
\caption{Cumulative variance of the 15 centred tool vectors, one model per family. All three reach ${\sim}$90\% at 10 of the 14 available components. Cross-source (39 tools) in Appendix~\ref{app:pca}.}
\label{fig:pca}
\end{figure}

This section describes the geometry of the tool vectors. The geometry is descriptive. The causal evidence is the steering and patching of the main text. For the 15-tool set, we centre the 15 tool vectors, measured at the last prompt token, and apply principal component analysis (PCA; Appendix~\ref{app:pca}; Figure~\ref{fig:pca}).

Non-tool requests under the same prompt also occupy relatively few dimensions, so we use PCA only to describe the activations. Steering and patching test the role of the tool vectors in selection.

\textbf{Matched-prompt control.} Independent Gaussian points do not share the long prompt prefix used to collect the tool vectors. A stronger control replaces tool requests with non-tool topics while keeping the rest of the prompt fixed (Table~\ref{tab:matched_control}, Appendix~\ref{app:matched_prompt}); Appendix~\ref{app:matched_prompt} reports the pairwise cosines. At $K{=}200$, tool vectors occupy more dimensions than control means on some models and fewer on others, with no clear difference on Gemma~3 4B. The spectrum alone therefore does not distinguish tool identity from shared-prompt effects. The matched $K{=}15$ control reaches the same conclusion (Appendix~\ref{app:matched_prompt}). The within-topic probe shows that tool labels remain readable within one topic, and steering and patching test effects on selection (Section~\ref{sec:discussion}).

For the five models with stored full spectra, $k_{90}$, the smallest number of components explaining 90\% of the variance, is 10 or 11 out of a maximum of 14. This measures the spread of the tool vectors, not a count of tool features, since networks can represent overlapping concepts in shared directions \citep{elhage2022superposition}.

The tool vectors occupy fewer dimensions than the Gaussian baseline, but more than non-tool topics under the matched prompt ($k_{90}{=}7$--9 across four models). Comparable spectra in base models show that this geometry is present before instruction tuning.

Replacing tool names with anonymous IDs preserves the low-dimensional spectrum (Appendix~\ref{app:anon_names}), and shuffling the tool list across five permutations leaves the tool vectors similar (Appendix~\ref{app:position_bias}). These controls address name and position effects. The topic-injection analysis (Section~\ref{sec:discussion}) and description ablation (Appendix~\ref{app:desc_ablation}) address information supplied by the descriptions.

We next check the same pattern with more tools, using real ToolBench API definitions \citep{qin2024toolllm}, sampled evenly across 49 domains from a pool of 12,000+ APIs (Table~\ref{tab:pca_toolbench}, Appendix~\ref{app:pca}).

More tools generally require more principal components, but the increase is slower than for independent Gaussian points. This curve describes the spectrum only.

\begin{figure}[htbp]
\centering
\includegraphics[width=\linewidth]{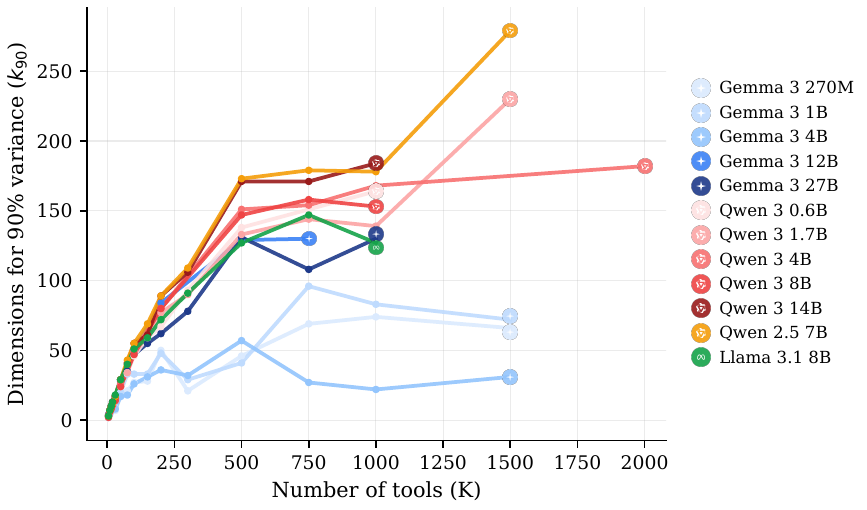}
\caption{PCA dimensionality $k_{90}$ for ToolBench tool lists of 5 to 2{,}000 APIs across 12 instruction-tuned models from three families (stratified across 49 domains, with descriptions). Dimensionality grows more slowly than the number of tools, with substantial differences between models. Gemma~3 4B declines sharply beyond $K{=}500$, where prompts become long. Table~\ref{tab:pca_crossmodel} gives the corresponding model comparisons; architecture and parameter count vary together.}
\label{fig:pca_scaling}
\end{figure}

\textbf{Prompt-length effects.} The curve is non-monotonic, and its shape changes with API sampling (Figure~\ref{fig:pca_scaling}). Descriptions reduce $k_{90}$ with moderate numbers of tools but lengthen the prompt enough for it to fall sharply beyond $K{=}750$ on Gemma~3 4B (Appendix~\ref{app:desc_ablation}). They also improve cross-lingual classification in the separate test of Appendix~\ref{app:crosslingual}.

\subsubsection{PCA method details}
\label{app:pca_method}

For each of $K$ tools, we collect 2--3 example requests and record the last-layer residual activation at the last prompt token, just before the output projection. Averaging over requests gives one tool vector per tool: $\bar{\resid}_i \in \mathbb{R}^{d}$ for $i = 1, \ldots, K$.

\textbf{Step 1: Center.} Subtract the grand mean so the data is zero-centered:
$\mathbf{C}_i = \bar{\resid}_i - \frac{1}{K}\sum_j \bar{\resid}_j$

\textbf{Step 2: SVD.} Stack the centered vectors into a $K \times d$ matrix and compute the singular value decomposition $\mathbf{C} = \mathbf{U} \mathbf{S} \mathbf{V}^\top$, where the diagonal entries $s_1 \geq s_2 \geq \ldots$ are the singular values.

\textbf{Step 3: Variance explained.} The fraction of total variance captured by the first $k$ components is:
$\text{Var}(k) = \frac{\sum_{i=1}^{k} s_i^2}{\sum_{i=1}^{K-1} s_i^2}$

We define $k_{90}$ as the smallest $k$ such that $\text{Var}(k) \geq 0.90$.

\textbf{Interpretation.} The maximum rank of $K$ centred points is $K{-}1$. A smaller $k_{90}$ means that fewer directions explain most of their variance. A Monte Carlo baseline using random points in $\mathbb{R}^{2560}$ has a larger $k_{90}$ than the observed tool vectors (Table~\ref{tab:pca_toolbench}). The matched-prompt controls show how much of that difference is specific to tools.

\subsubsection{Cross-model PCA comparison}
\label{app:pca_compare}

Table~\ref{tab:pca_compare} reports cumulative variance and $k_{90}$ across Gemma and Qwen at $K{=}15$.

\begin{table}[htbp]
\centering
\caption{Cumulative variance explained by the top-$k$ principal components of the 15 tool vectors. Gemma~3 4B and Qwen~3 4B have similar cumulative spectra.}
\label{tab:pca_compare}

\small
\setlength{\tabcolsep}{7pt}
\rowcolors{2}{tablerow}{white}
\resizebox{\ifdim\width>\columnwidth\columnwidth\else\width\fi}{!}{\begin{tabular}{rcc}
\toprule
\textsc{Top-$k$ PCs} & \raisebox{-1pt}{\includegraphics[height=8pt]{figures/logos/gemma.png}}\,\textsc{Gemma~3 4B} & \raisebox{-1pt}{\includegraphics[height=8pt]{figures/logos/qwen.png}}\,\textsc{Qwen~3 4B} \\
\midrule
1 & 27.5\% & 21.0\% \\
3 & 51.3\% & 48.6\% \\
5 & 67.5\% & 64.5\% \\
10 & 92.5\% & 90.2\% \\
14 & 100.0\% & 100.0\% \\
\bottomrule
\end{tabular}}
\rowcolors{1}{}{}

\end{table}

\subsubsection{Cross-source tool subspace}
\label{app:combined_pca}

\begin{figure}[htbp]
\centering
\includegraphics[width=0.62\linewidth]{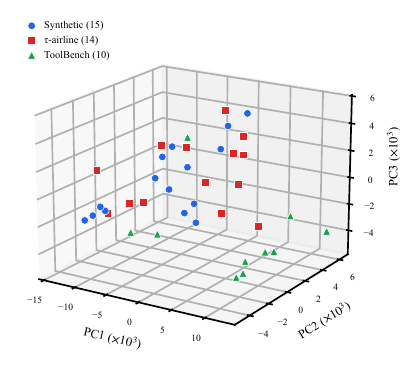}
\caption{First three principal components of the tool vectors for 39 tools on Gemma~3 4B: 15 synthetic, 14 $\tau$-bench airline and 10 ToolBench tools. These components explain 59\% of the variance. ToolBench appears separated on PC3, but that component accounts for only 6\%. In the full space, leave-one-out nearest-neighbour source classification reaches 44\% against a 38\% majority baseline; source-centroid distances are 0.57--1.16 times the within-source spread. These results give limited evidence for separation by source.}
\label{fig:combined_pca}
\end{figure}

Combining the three tool sets tests for one shared subspace.
We take the 15 synthetic, 14 $\tau$-bench airline and 10 ToolBench tool vectors on Gemma~3 4B
and run the same centred PCA (Figure~\ref{fig:combined_pca}). Leave-one-out
nearest-neighbour classification of a mean's source reaches 44\% against a 38\% majority
baseline, and the distances between source centroids are 0.57 to 1.16 times the spread
within a source, so the sets overlap by source.

\subsubsection{PCA scaling with ToolBench APIs}
\label{app:pca_toolbench}

Table~\ref{tab:pca_toolbench} shows the full PCA scaling curve using real API definitions from ToolBench (12,000+ APIs, stratified sampling across 49 domains, with descriptions).

\begin{table}[htbp]
\centering
\caption{PCA scaling on Gemma~3 4B with stratified ToolBench APIs and descriptions. Max is $K{-}1$; Random is $k_{90}$ for independent Gaussian points; Compress is $k_{90}/(K{-}1)$. The $K{=}500$ prompt contains about 10K tokens. A smaller value indicates a more concentrated spectrum.}
\label{tab:pca_toolbench}

\small
\setlength{\tabcolsep}{7pt}
\rowcolors{2}{tablerow}{white}
\begin{tabular}{@{}r rrr r@{}}
\toprule
\textsc{$K$} & \textsc{$k_{90}$} & \textsc{Max} & \textsc{Random} & \textsc{Compress} \\
\midrule
5 & 2 & 4 & 4.0 & 50\% \\
10 & 5 & 9 & 8.6 & 56\% \\
15 & 7 & 14 & 13.0 & 50\% \\
20 & 8 & 19 & 17.0 & 42\% \\
30 & 8 & 29 & 26.0 & 28\% \\
50 & 17 & 49 & 43.0 & 35\% \\
75 & 18 & 74 & 65.0 & 24\% \\
100 & 26 & 99 & 86.0 & 26\% \\
150 & 31 & 149 & 127.0 & 21\% \\
200 & 36 & 199 & 167.1 & 18\% \\
300 & 32 & 299 & 245.5 & 11\% \\
500 & 57 & 499 & 392.0 & 11\% \\
\bottomrule
\end{tabular}
\rowcolors{1}{}{}

\end{table}

\textbf{With vs.\ without descriptions.} Removing descriptions lets Gemma~3 4B process more tools without the sharp decline in $k_{90}$ seen with descriptions. The name-only prompt remains under 12K tokens at $K{=}2000$, where $k_{90}{=}122$; with descriptions, the decline appears beyond $K{=}750$ at roughly 16K tokens. At moderate sizes, descriptions instead reduce the number of components needed. These measurements identify a prompt-length constraint on the spectrum.

\textbf{Cross-model comparison.}
We repeat the PCA scaling on all 12 instruction-tuned models from three families. Table~\ref{tab:pca_crossmodel} shows the results.

\begin{table}[htbp]
\centering
\caption{PCA scaling across all 12 instruction-tuned models. H is the attention-head count from the official configurations; arrows mark declines discussed in the text. The Gemma~3 4B spectrum contracts with many tools, while larger Gemma models retain more dimensions. The two smallest Gemma models vary non-monotonically at lower absolute dimensionality. Beyond the largest tool list shown here, separate runs reach $K{=}2000$ on Qwen~3 4B ($k_{90}{=}182$) and $K{=}1500$ on Qwen~2.5 7B and Qwen~3 1.7B ($k_{90}{=}279$ and $230$).}

\small
\setlength{\tabcolsep}{6pt}
\rowcolors{2}{tablerow}{white}
\begin{tabular}{@{}l r rrrrrr@{}}
\toprule
\textsc{Model} & \textsc{H} & \multicolumn{6}{c}{\textsc{$k_{90}$ at $K=$}} \\
\cmidrule(lr){3-8}
& & 50 & 100 & 200 & 500 & 750 & 1000 \\
\midrule
\raisebox{-1pt}{\includegraphics[height=8pt]{figures/logos/gemma.png}}\,Gemma~3 270M & 4 & 17 & 27 & 50 & 46 & 69 & 74 \\
\raisebox{-1pt}{\includegraphics[height=8pt]{figures/logos/gemma.png}}\,Gemma~3 1B & 4 & 20 & 33 & 48 & 41 & 96 & 83 \\
\raisebox{-1pt}{\includegraphics[height=8pt]{figures/logos/gemma.png}}\,Gemma~3 4B & 8 & 17 & 26 & 36 & 57 & 27$\downarrow$ & 22$\downarrow$ \\
\raisebox{-1pt}{\includegraphics[height=8pt]{figures/logos/gemma.png}}\,Gemma~3 12B & 16 & 28 & 50 & 84 & 129 & 130 & --- \\
\raisebox{-1pt}{\includegraphics[height=8pt]{figures/logos/gemma.png}}\,Gemma~3 27B & 32 & 29 & 47 & 62 & 131 & 108 & 130 \\
\raisebox{-1pt}{\includegraphics[height=8pt]{figures/logos/qwen.png}}\,Qwen~3 0.6B & 16 & 24 & 46 & 67 & 138 & 151 & 164 \\
\raisebox{-1pt}{\includegraphics[height=8pt]{figures/logos/qwen.png}}\,Qwen~3 1.7B & 16 & 25 & 49 & 77 & 133 & 144 & 139 \\
\raisebox{-1pt}{\includegraphics[height=8pt]{figures/logos/qwen.png}}\,Qwen~3 4B & 32 & 26 & 49 & 80 & 151 & 154 & 168 \\
\raisebox{-1pt}{\includegraphics[height=8pt]{figures/logos/qwen.png}}\,Qwen~3 8B & 32 & 24 & 47 & 80 & 147 & 158 & 153 \\
\raisebox{-1pt}{\includegraphics[height=8pt]{figures/logos/qwen.png}}\,Qwen~3 14B & 40 & 29 & 55 & 89 & 171 & 171 & 184 \\
\raisebox{-1pt}{\includegraphics[height=8pt]{figures/logos/qwen.png}}\,Qwen~2.5 7B & 28 & 29 & 55 & 89 & 173 & 179 & 178 \\
\raisebox{-1pt}{\includegraphics[height=8pt]{figures/logos/llama.png}}\,Llama~3.1 8B & 32 & 29 & 51 & 72 & 127 & 147 & 127$\downarrow$ \\
\bottomrule
\end{tabular}
\rowcolors{1}{}{}

\label{tab:pca_crossmodel}
\end{table}

With more tools, the models differ substantially in $k_{90}$ (Table~\ref{tab:pca_crossmodel}). Head count varies together with hidden width, parameter count, architecture and pretraining data, so this comparison cannot identify which of them causes the differences.

With the most tools, larger Gemma models retain more dimensions than Gemma~3 4B, which may reflect context handling or other architectural differences.

\textbf{Sampling sensitivity.} Prioritizing APIs with different name prefixes within each domain changes the curve: at $K{=}500$, this sampling method gives $k_{90}{=}30$, compared with 57 under stratified sampling. The non-monotonic pattern therefore depends on the sampled API set, rather than being a fixed property of the model.

\textbf{Same-domain vs.\ cross-domain.} Finance-only tool lists occupy fewer dimensions than the cross-domain set at $K{=}500$ (Table~\ref{tab:domain_pca}). The Finance spectrum also changes little between $K{=}50$ and $75$. This is consistent with additional tools sharing activation directions already represented in the tool list.

\textbf{Multi-domain comparison.} We run the same analysis on individual domains to see how within-domain similarity affects compression. Table~\ref{tab:domain_pca} shows the results.

\begin{table}[htbp]
\centering
\caption{PCA dimensionality $k_{90}$ by domain and number of tools. Spectra are similar with few tools and diverge as tools are added. Some domains require more components, while others plateau or decline. These are observations about sampled activations; prompt length and API composition both vary with the number of tools. Arrows mark declines from the previous column. $^\dag$Fewer APIs available.}
\label{tab:domain_pca}

\small
\setlength{\tabcolsep}{7pt}
\rowcolors{2}{tablerow}{white}
\begin{tabular}{@{}l rrrrrr@{}}
\toprule
\textsc{Domain} & \textsc{5} & \textsc{15} & \textsc{50} & \textsc{100} & \textsc{200} & \textsc{500} \\
\midrule
Stratified (49) & 2 & 7 & 17 & 26 & 36 & 57 \\
Business & 4 & 10 & 17 & 31 & 33 & 56 \\
Finance & 3 & 10 & 15 & 22 & 25 & 32 \\
Data & 3 & 8 & 16 & 26 & 22 & 39 \\
Sports & 3 & 7 & 14 & 24 & 33 & 11$\downarrow$ \\
Gaming & 4 & 8 & 10 & 16 & 13 & 25 \\
Entertainment & 2 & 6 & 11 & 16 & 11$\downarrow$ & 5$\downarrow$ \\
Weather$^\dag$ & 3 & 7 & 21 & 21 & --- & --- \\
Science$^\dag$ & 3 & 8 & 14 & --- & --- & --- \\
\bottomrule
\end{tabular}
\rowcolors{1}{}{}

\end{table}

\subsection{Matched-prompt controls}
\label{app:matched_prompt}

\subsubsection{Cosine geometry under matched prompt}
\label{app:cosine_baseline}

Table~\ref{tab:cosine_baseline} compares cosine similarities among tool vectors, among topic means, and across the two groups, using the same prompt prefix.

\begin{table}[htbp]
\centering
\caption{Pairwise cosine of 200 averaged states under matched prompt prefix (last layer). \textsc{Tool} / \textsc{Topic} / \textsc{Cross}: cosines among tool vectors, among topic means, and between the two. $\Delta$: tool-pair minus topic-pair cosine; negative means tool vectors are more spread.}
\label{tab:cosine_baseline}

\footnotesize
\setlength{\tabcolsep}{7pt}
\begin{tabular}{@{}l rrrr@{}}
\toprule
& \textsc{Tool} & \textsc{Topic} & \textsc{Cross} & \textsc{$\Delta$} \\
\midrule
\multicolumn{5}{@{}l}{\textit{Instruction-tuned}} \\
\raisebox{-1pt}{\includegraphics[height=8pt]{figures/logos/gemma.png}}\,Gemma~3 4B    & 0.960 & 0.957 & 0.916 & $+$0.003 \\
\rowcolor{tablerow} \raisebox{-1pt}{\includegraphics[height=8pt]{figures/logos/gemma.png}}\,Gemma~3 12B   & 0.927 & 0.963 & 0.906 & $-$0.036 \\
\raisebox{-1pt}{\includegraphics[height=8pt]{figures/logos/llama.png}}\,Llama~3.1 8B  & 0.873 & 0.947 & 0.855 & $-$0.074 \\
\rowcolor{tablerow} \raisebox{-1pt}{\includegraphics[height=8pt]{figures/logos/qwen.png}}\,Qwen~3 4B     & 0.792 & 0.945 & 0.779 & $-$0.153 \\
\midrule
\multicolumn{5}{@{}l}{\textit{Base (pretrained-only)}} \\
\raisebox{-1pt}{\includegraphics[height=8pt]{figures/logos/gemma.png}}\,Gemma~3 4B-pt    & 0.999 & 0.999 & 0.998 & $-$0.000 \\
\rowcolor{tablerow} \raisebox{-1pt}{\includegraphics[height=8pt]{figures/logos/gemma.png}}\,Gemma~3 12B-pt   & 0.997 & 0.997 & 0.996 & $+$0.000 \\
\raisebox{-1pt}{\includegraphics[height=8pt]{figures/logos/llama.png}}\,Llama~3.1 8B-base & 0.922 & 0.943 & 0.892 & $-$0.022 \\
\bottomrule
\end{tabular}

\end{table}

\subsubsection{K=15 matched-prompt control}
\label{app:matched_k15}

We compare requests for each of 15 synthetic tools with requests about 15 non-tool
topics under the same tool list and prompt template (Table~\ref{tab:matched_control_k15}).
The control topics are Roman history, quantum physics, economics, philosophy, literature,
biology, music theory, geography, religion, mathematics, astronomy, anthropology,
psychology, art history, and linguistics, with two requests per topic. We average
last-layer residuals at the final prompt token across the two requests and run PCA over
the 15 averages in each condition.

\begin{table}[htbp]
\centering
\caption{K=15 PCA cumulative variance and $k_{90}$ for tool-trigger requests vs.\ non-tool topic requests under the same 15-tool prompt prefix. \textsc{t-4} / \textsc{t-10}: cumulative variance in the top 4 / top 10 components.}
\label{tab:matched_control_k15}

\small
\setlength{\tabcolsep}{6pt}
\begin{tabular}{@{}l rrr rrr@{}}
\toprule
& \multicolumn{3}{c}{\textsc{Tools}} & \multicolumn{3}{c}{\textsc{Non-tool}} \\
\cmidrule(lr){2-4} \cmidrule(lr){5-7}
& t-4 & t-10 & $k_{90}$ & t-4 & t-10 & $k_{90}$ \\
\midrule
\raisebox{-1pt}{\includegraphics[height=8pt]{figures/logos/gemma.png}}\,Gemma~3 4B IT  & 0.61 & 0.93 & 10 & 0.71 & 0.94 & 9 \\
\raisebox{-1pt}{\includegraphics[height=8pt]{figures/logos/qwen.png}}\,Qwen~3 4B      & 0.57 & 0.91 & 10 & 0.77 & 0.96 & 8 \\
\raisebox{-1pt}{\includegraphics[height=8pt]{figures/logos/llama.png}}\,Llama~3.1 8B   & 0.57 & 0.90 & 11 & 0.68 & 0.93 & 9 \\
\raisebox{-1pt}{\includegraphics[height=8pt]{figures/logos/gemma.png}}\,Gemma~3 12B IT & 0.56 & 0.90 & 10 & 0.84 & 0.96 & 7 \\
\bottomrule
\end{tabular}

\end{table}

Non-tool requests require fewer principal components than tool requests on all four models
(Table~\ref{tab:matched_control_k15}).  Evidence beyond PCA comes from steering
and patching interventions, together with the correlational within-topic probe.

\subsubsection{K=200 matched-prompt control}
\label{app:matched_control_200}

At $K{=}200$, we compare tool vectors for 200 ToolBench APIs with averages for 200 non-tool topics,
holding the 200-tool prompt prefix fixed.

\begin{table}[htbp]
\centering
\caption{Last-layer $k_{90}$ at $K{=}200$. \textsc{Tools}: ToolBench tool vectors. \textsc{Match}: non-tool topic means under the same 200-tool prompt prefix. Negative \textsc{Diff} means tools require more dimensions; positive means fewer. \textsuperscript{$\dagger$}Gemma~3 12B's non-tool baseline is dominated by PC1 (72\% variance). A separate subsampling run using 80\% of the 200 requests, $B{=}200$, gives difference intervals of $[-2, +5]$ (Gemma~3 4B), $[-26, -18]$ (Qwen~3 4B), $[+10, +17]$ (Llama~3.1 8B), and $[-56, -51]$ (Gemma~3 12B). Only Gemma~3 4B's interval crosses zero. Subsampling changes both $k_{90}$ values, so these intervals describe that run's differences rather than the full-sample table values.}

\small
\setlength{\tabcolsep}{7pt}
\begin{tabular}{@{}l rrr l@{}}
\toprule
& \textsc{Tools} & \textsc{Match} & \textsc{Diff} & \textsc{Reading} \\
\midrule
\raisebox{-1pt}{\includegraphics[height=8pt]{figures/logos/gemma.png}}\,Gemma~3 4B   & 36 & 39 & $+$3   & similar        \\
\raisebox{-1pt}{\includegraphics[height=8pt]{figures/logos/qwen.png}}\,Qwen~3 4B    & 80 & 52 & $-$28  & tools spread   \\
\raisebox{-1pt}{\includegraphics[height=8pt]{figures/logos/llama.png}}\,Llama~3.1 8B & 72 & 83 & $+$11  & tools cluster  \\
\raisebox{-1pt}{\includegraphics[height=8pt]{figures/logos/gemma.png}}\,Gemma~3 12B  & 84 & 11 & $-$73  & tools spread\textsuperscript{$\dagger$} \\
\bottomrule
\end{tabular}

\label{tab:matched_control}
\end{table}

Tool requests occupy more dimensions on two models, fewer on one, and a similar number on one (Table~\ref{tab:matched_control}).

\subsection{The state follows the tool list as well as the request}
\label{app:menu_removal}
The state reflects the tool list as well as the request. To show this, we hold the request fixed and remove its requested tool. For each of the 15 tools, we run its
example request with the full tool list and with that tool deleted. We classify both states by
cosine similarity to the tool vectors of the 14 remaining tools. Among requests where the model
names a replacement from the tool list, we compare which state better predicts that choice
(Table~\ref{tab:menu_removal}).

\begin{table}[htbp]
\centering
\caption{Same request, tool list with and without the gold tool. \textsc{Sub}: requests out of 15
where the model names a different listed tool once the gold one is deleted. \textsc{After} and
\textsc{Before}: how many of those the cosine readout names correctly from the state after
the deletion and from the state before it. Chance is 1 in 14.}
\label{tab:menu_removal}
\small
\setlength{\tabcolsep}{5pt}
\rowcolors{2}{tablerow}{white}
\begin{tabular}{@{}l l r r r r@{}}
\toprule
\textsc{Model} & \textsc{Layer} & L & \textsc{Sub} & \textsc{After} & \textsc{Before} \\
\midrule
\raisebox{-1pt}{\includegraphics[height=8pt]{figures/logos/qwen.png}}\,Qwen~3 4B & sep. & 25 & 10 & 5 & 3 \\
\raisebox{-1pt}{\includegraphics[height=8pt]{figures/logos/gemma.png}}\,Gemma~3 4B & sep. & 21 & 12 & 6 & 3 \\
\raisebox{-1pt}{\includegraphics[height=8pt]{figures/logos/llama.png}}\,Llama~3.1 8B & sep. & 18 & 13 & 5 & 2 \\
\midrule
\raisebox{-1pt}{\includegraphics[height=8pt]{figures/logos/qwen.png}}\,Qwen~3 4B & last & 35 & 10 & 6 & 3 \\
\raisebox{-1pt}{\includegraphics[height=8pt]{figures/logos/gemma.png}}\,Gemma~3 4B & last & 33 & 12 & 7 & 3 \\
\raisebox{-1pt}{\includegraphics[height=8pt]{figures/logos/llama.png}}\,Llama~3.1 8B & last & 31 & 13 & 9 & 1 \\
\bottomrule
\end{tabular}

\end{table}

The state after deletion predicts the replacement better at both depths. Among cases
where only one readout is correct, it wins 18 of 21 at the last layer (exact McNemar
$p{=}0.001$) and 10 of 12 at the peak layer ($p{=}0.04$). Thus the hidden state at the last prompt token
responds to the tool list even when the request is fixed. Request semantics may also contribute to the signal.

\section{Error ranking}
\label{app:risk_coverage}
An internal error signal must be compared with the model's own output probabilities, since models can often judge the correctness of their own answers \citep{kadavath2022know}, and sequence probability is a standard reference for detecting wrong generations \citep{malinin2021uncertainty,kuhn2023semantic}.

\subsection{Ranking calls by error risk}
\label{sec:monitor}
A small gap between the two highest cosine scores is associated with tool-selection errors. We use this \emph{cosine gap} to rank calls by error risk (Table~\ref{tab:monitor}). Errors are 2.9 to 15$\times$ more concentrated in the lowest-gap quartile, but each model has only 13 to 61 errors and some quartiles contain none, making these ratios imprecise. Error-ranking performance exceeds chance on all four models. The baseline margin between summed full-tool-name log-probabilities performs similarly, with no significant paired difference on any model. The first-token margin, which costs no more than the gap, ranks errors worse on three of the four (Table~\ref{tab:monitor}). Qwen~3 4B is the weakest case, with a precision--recall area only 1.3$\times$ the error base rate. These results support a model-dependent ranking.

Withholding calls with the smallest cosine gaps generally reduces the error rate among the remaining calls, although the improvement is not monotonic at every threshold (Table~\ref{tab:risk_coverage}). The table reports both the remaining error rate and the fraction of observed errors withheld at four budgets. Qwen~3 4B improves least. The withholding analysis is retrospective, on one benchmark population.

\begin{table}[htbp]
\centering
\caption{Error rate among retained calls and fraction of all errors withheld when excluding the 10/20/30/50\% of calls with the smallest cosine gaps. The population is format-excluded BFCL (Appendix~\ref{app:eval_protocol}); error rates before exclusion are 2.6--13.4\%. These are retrospective rankings, not results from an intervention on executed calls.}
\label{tab:risk_coverage}
\small
\resizebox{\ifdim\width>\columnwidth\columnwidth\else\width\fi}{!}{\begin{tabular}{lcccccccc}
\toprule
& \multicolumn{2}{c}{\textsc{abstain} 10\%} & \multicolumn{2}{c}{20\%} & \multicolumn{2}{c}{30\%} & \multicolumn{2}{c}{50\%} \\
\textsc{Model} & \textsc{kept err.} & \textsc{caught} & \textsc{kept err.} & \textsc{caught} & \textsc{kept err.} & \textsc{caught} & \textsc{kept err.} & \textsc{caught} \\
\midrule
\raisebox{-1pt}{\includegraphics[height=8pt]{figures/logos/gemma.png}}\,Gemma~3 4B   & 3.4\% & 25\% & 3.3\% & 35\% & 1.4\% & 75\% & 0.8\% & 90\% \\
\rowcolor{tablerow} \raisebox{-1pt}{\includegraphics[height=8pt]{figures/logos/gemma.png}}\,Gemma~3 12B  & 1.6\% & 46\% & 1.5\% & 54\% & 1.2\% & 69\% & 0.8\% & 85\% \\
\raisebox{-1pt}{\includegraphics[height=8pt]{figures/logos/qwen.png}}\,Qwen~3 4B    & 12.7\% & 15\% & 13.2\% & 21\% & 11.0\% & 43\% & 9.3\% & 66\% \\
\rowcolor{tablerow} \raisebox{-1pt}{\includegraphics[height=8pt]{figures/logos/llama.png}}\,Llama~3.1 8B & 2.8\% & 25\% & 3.2\% & 25\% & 1.5\% & 69\% & 0.0\% & 100\% \\
\bottomrule
\end{tabular}}

\end{table}

\subsection{What each signal costs}
\label{app:monitor_cost}
The gap and the two margins in Table~\ref{tab:monitor} are read after the same prompt forward pass, so that pass is not charged to any of them. The generated-name probability is not timed: it needs nothing beyond the generation the model runs anyway, but it is only available once the call has been written. They differ in what each one needs afterwards. The cosine gap is $K$ dot products against stored tool vectors. The first-token margin is two entries of the logits the model has already computed for its first output token. The full-name margin scores every candidate's name tokens, and it is the only one of the three that runs the model again.

We time the full-name margin under the faster of two implementations. One reuses the prompt's key--value cache and scores only the candidates' own name tokens, and the other scores all candidates in a single batch. The naive column is the archived implementation, one forward pass per candidate. On both Gemma models the cache cannot be cropped for reuse, so the batched variant is the faster one there. Timings are medians over 60 examples, best of three repeats, on one NVIDIA GB10, with the same tool lists and the same positions as Table~\ref{tab:monitor}.

\begin{table}[htbp]
\centering
\caption{Median wall-clock cost of each monitor signal on top of the shared prompt pass, which every signal needs and none is charged for. \textsc{Full name} is the faster of a cache-reusing and a batched implementation; \textsc{(naive)} is one forward pass per candidate. \textsc{Dict.} is the size of the stored tool vectors in float32, per tool. Tool lists have four candidates and prompts 155 to 177 tokens at the median; 60 examples per model, best of three repeats, all four on one NVIDIA GB10.}
\label{tab:monitor_cost}
\small
\resizebox{\ifdim\width>\columnwidth\columnwidth\else\width\fi}{!}{\begin{tabular}{lcccccc}
\toprule
& \textsc{Prompt pass} & \multicolumn{4}{c}{\textsc{Additional cost} (ms)} & \textsc{Dict.} \\
\cmidrule(lr){3-6}
\textsc{Model} & (ms) & \textsc{Cosine gap} & \textsc{First token} & \textsc{Full name} & \textsc{(naive)} & (KiB/tool) \\
\midrule
\raisebox{-1pt}{\includegraphics[height=8pt]{figures/logos/gemma.png}}\,Gemma~3 4B & 58 & 0.045 & 0.083 & 150 & 255 & 10 \\
\rowcolor{tablerow} \raisebox{-1pt}{\includegraphics[height=8pt]{figures/logos/gemma.png}}\,Gemma~3 12B & 158 & 0.045 & 0.083 & 446 & 697 & 15 \\
\raisebox{-1pt}{\includegraphics[height=8pt]{figures/logos/qwen.png}}\,Qwen~3 4B & 55 & 0.045 & 0.060 & 138 & 238 & 10 \\
\rowcolor{tablerow} \raisebox{-1pt}{\includegraphics[height=8pt]{figures/logos/llama.png}}\,Llama~3.1 8B & 92 & 0.046 & 0.060 & 233 & 390 & 16 \\
\bottomrule
\end{tabular}}

\end{table}

\section{Extended limitations}
\label{app:limitations_extended}

\textbf{Unsuccessful interventions.}

\emph{Argument steering} fails on the tested Gemma~3 4B setup. Redirecting ``Tokyo'' to ``Paris'' while keeping the tool fixed leaves every argument value unchanged in 30 trials. Unlike tool selection from a fixed tool list, argument values depend on the request. We intervene only at the last prompt token; intervention at an argument's own token position remains untested. \emph{Format steering} (JSON vs.\ plain text) also fails on this model (0/10). Applying a Gemma vector directly to Qwen gives 0\% switching, so these vectors do not transfer directly across families. Transfer after aligning their activation spaces remains untested.

\subsection{Multi-turn deployment}
\label{sec:multiturn}
The readout-based override does not consistently improve multi-turn task success. On $\tau$-bench airline, a cosine readout selects the nearest tool vector at each tool call and replaces the model's choice when they disagree, unless a schema-compatibility check blocks the replacement. No vector is added. We compare this override with fresh unsteered runs on the same twenty tasks. At the last-layer residual used for single-turn steering, task success changes by $-$30 to $+$5 percentage points across five 4B--14B models (Gemma~3 4B/12B, Qwen~3 4B/8B/14B). The two largest changes are decreases (Gemma~12B and Qwen~8B, both at the last layer). A four-layer sweep on Gemma~3 4B reaches $+$10 at its best layer. This selected maximum is not comparable to the fixed-layer range. With $N{=}20$ tasks, no difference is significant (Fisher exact $p \geq 0.10$), so these runs establish neither reliable benefit nor harm.

A within-topic probe reads the upcoming tool above its noise control (Section~\ref{sec:discussion}). A readout-based override in a full agent loop, and vector injection inside that loop, remain untested. Replacing a call could disrupt the model's plan, or longer context could weaken the readout.

Steering can change the tool name and argument structure without producing a valid call. Outputs sometimes invent argument values, such as a restaurant name absent from the request (Section~\ref{sec:schema}; Table~\ref{tab:schema_ar}). Outside the multi-turn runs above, we evaluate tool choice and JSON structure, not end-to-end task completion. Whether an API would accept these calls and complete the task is untested.

At the last layer, the steering direction and an output direction both change the next token's scores, so last-layer steering alone cannot tell the two directions apart. Figure~\ref{fig:depth_3fam} compares the two directions across depth on held-out requests (Appendix~\ref{app:output_row}), and Table~\ref{tab:midstack_heldout} gives held-out checks at selected layers. The layers that separate the two directions are less effective places to steer than the last layer in our tests (Appendix~\ref{app:output_row}).

Steering is unreliable in the smallest tested models. Gemma~3 270M shows no effective steering, and sub-1B results vary across tool lists (Table~\ref{tab:generalization}). A minimum model size, and the training changes that make steering effective, remain open.

Steering fails entirely on near-synonymous tools (0\% on 10 similar Sports APIs, Gemma~3 4B; Appendix~\ref{app:toolbench_domains}). These are the tools whose vectors are nearly parallel (Appendix~\ref{app:robustness}).

The example requests used to build the tool vectors are written to clearly match one tool. We also test paraphrased, indirect, entity-varied, ambiguous and out-of-distribution example requests (Appendix~\ref{app:robustness}). Ambiguity reduces steering more than readout accuracy. These comparisons keep the tool list fixed, and the list-change test keeps the example requests fixed.

Replacing argument keys with positional labels such as \texttt{p1} and \texttt{p2}, while holding the tool name fixed, reduces conformity to the displayed schema (Section~\ref{sec:schema}). This suggests that familiar key names help argument generation.

Long prompts limit the observed geometry. On Gemma~3 4B, tool lists with descriptions reach ${\sim}$16K tokens at $K{=}750$, where $k_{90}$, the number of PCA components needed for 90\% of the variance, falls sharply. Name-only tool lists retain a larger $k_{90}$ at $K{=}1500$ (Appendix~\ref{app:pca}). This comparison concerns the activation spectrum only.

Only dense (non-mixture-of-experts) Gemma, Qwen, and Llama families are tested; other architectures may differ.

\end{document}